\documentclass[twoside,11pt]{article}

\usepackage{jmlr2e}

\usepackage[cmex10]{amsmath}
\usepackage{bm}
\usepackage{algpseudocode}
\usepackage{subfigure}
\usepackage{color}
\newcommand{\mb}[1]{\mbox{\boldmath $#1$}}
\newcommand{\valpha}[0]{\mb{\alpha}}

\newcommand{\vK}[0]{\mb{K}}
\newcommand{\vQ}[0]{\mb{Q}}
\newcommand{\vM}[0]{\mb{M}}
\newcommand{\vx}[0]{\mb{x}}
\newcommand{\vy}[0]{\mb{y}}

\newcommand{\pd}[2]{{\frac{\partial #1}{\partial #2}}}
\newcommand{\argmax}{\mathop{\rm argmax}}
\newcommand{\argmin}{\mathop{\rm argmin}}
\newcommand{\eq}[1]{(\ref{#1})}

\newcommand{\lw}[1]{\smash{\lower2.ex\hbox{#1}}}

\newcommand{\veps}{\varepsilon}

\newcommand{\RR}{\mathbb{R}}

\newcommand{\cD}{{\cal D}}

\newcommand{\cF}{{\cal F}}

\newcommand{\cK}{{\cal K}}

\newcommand{\cM}{{\cal M}}
\newcommand{\cN}{{\cal N}}
\newcommand{\cO}{{\cal O}}
\newcommand{\cP}{{\cal P}}

\newcommand{\cR}{{\cal R}}

\newcommand{\cV}{{\cal V}}
\newcommand{\cX}{{\cal X}}

\jmlrheading{1}{2000}{1-48}{4/00}{10/00}{Masayuki Karasuyama, Naoyuki
Harada, Masashi Sugiyama and Ichiro Takeuchi}

\ShortHeadings{Multi-parametric Solution-path Algorithm for
Instance-weighted SVMs}{Karasuyama, Harada,
Sugiyama and Takeuchi}
\firstpageno{1}

\begin{document}

\title{Multi-parametric Solution-path Algorithm for Instance-weighted Support Vector Machines}

\author{\name Masayuki Karasuyama \email krsym@goat.ics.nitech.ac.jp \\
%       \addr Department of Engineering\\
%       Nagoya Institute of Technology\\
%       Nagoya, Aichi 466-8555, Japan
%       \AND
       \name Naoyuki Harada \email harada@goat.ics.nitech.ac.jp \\
       \addr Department of Engineering\\
       Nagoya Institute of Technology\\
       Nagoya, Aichi 466-8555, Japan
       \AND
       \name Masashi Sugiyama \email sugi@cs.titech.ac.jp \\
       \addr Department of Computer Science\\
       Tokyo Institute of Technology\\
       Tokyo 152-8552 Japan
       \AND
       \name Ichiro Takeuchi \email takeuchi.ichiro@nitech.ac.jp \\
       \addr Department of Engineering\\
       Nagoya Institute of Technology\\
       Nagoya, Aichi 466-8555, Japan
       }

\editor{}

\maketitle

% \begin{abstract}%   <- trailing '%' for backward compatibility of .sty file
% An \emph{instance-weighted} variant of the support vector machine
% (SVM) has attracted considerable attention recently since they are
% useful in various machine learning tasks such as 
% non-stationary data analysis, heteroscedastic data modeling, transfer
% learning, learning to rank, and transduction. An important challenge in
% these scenarios is to overcome the computational bottleneck---instance
% weights often change dynamically or adaptively, and thus the
% weighted SVM solutions must be repeatedly computed. In this paper, we
% develop an algorithm that can efficiently and exactly update the
% weighted SVM solutions for arbitrary change of instance
% weights. Technically, this contribution can be regarded as an extension
% of the conventional \emph{solution-path} algorithm for a single
% regularization parameter to multiple instance-weight parameters.
% \red{However, this extension gives rise to a significant technical difficulty
% that \emph{breakpoints} (at which the solution path turns)
% have to be identified in high-dimensional space.}
% To facilitate this, we introduce a notion of \emph{critical region}:
% a polyhedron in which the current \emph{affine} solution remains to be optimal.
% Then we find breakpoints at intersections of the solution path and boundaries of polyhedrons.
% Through extensive experiments on various practical applications, we
% demonstrate the usefulness of the proposed algorithm.
% \end{abstract}

% \newpage
\begin{abstract}%   <- trailing '%' for backward compatibility of .sty file
An \emph{instance-weighted} variant of the support vector machine
(SVM) has attracted considerable attention recently since they are
useful in various machine learning tasks such as 
non-stationary data analysis, heteroscedastic data modeling, transfer
learning, learning to rank, and transduction. An important challenge in
these scenarios is to overcome the computational bottleneck---instance
weights often change dynamically or adaptively, and thus the
weighted SVM solutions must be repeatedly computed. In this paper, we
develop an algorithm that can efficiently and exactly update the
weighted SVM solutions for arbitrary change of instance
weights. Technically, this contribution can be regarded as an extension
of the conventional \emph{solution-path} algorithm for a single
regularization parameter to multiple instance-weight parameters.
However, this extension gives rise to a significant problem that
 \emph{breakpoints} (at which the solution path turns) have to be identified in
 high-dimensional space. To facilitate this, we introduce a parametric
 representation of instance weights. We
 also provide a geometric interpretation in weight space using a notion of
 \emph{critical region}: a polyhedron in which the current affine solution
 remains to be optimal.
 Then we find breakpoints at intersections of the solution path and boundaries of polyhedrons.
 Through extensive experiments on various practical applications, we
 demonstrate the usefulness of the proposed algorithm.
\end{abstract}


%==================================================
% Sec::Introduction
%==================================================

\section{Introduction}
The most fundamental principle of machine learning
would be the \emph{empirical risk minimization},
i.e., the sum of empirical losses over training instances is minimized:
\begin{align*}
  \min \sum_{i} L_{i},
\end{align*}
where $L_{i}$ denotes the empirical loss for the $i$-th training instance.
This empirical risk minimization approach was proved to produce \emph{consistent} estimators
\citep{book:Vapnik:1995}.
On the other hand,
one may also consider an \emph{instance-weighted} variant of empirical risk minimization:
\begin{align*}
  \min \sum_{i}C_{i}L_{i},
\end{align*}
where $C_{i}$ denotes the weight for the $i$-th training instance.
This weighted variant plays an important role in
various machine learning tasks:

\begin{itemize}
%\item\textbf{Cost-sensitive learning and imbalanced classification:}
%When the cost of misclassification from positive to negative is
%significantly different from that from negative to positive
%(e.g., rain without an umbrella or no rain with an umbrella),
%weighting the loss function according to the cost ratio
%helps improve the performance \cite{IJCAI:Elkan:2001}.
%The same idea can also be applied to imbalanced classification scenarios,
%where the numbers of positive and negative training samples are significantly different
%\cite{mach:lin+lee+wahba:2002}.

\item\textbf{Non-stationary data analysis:}
When training instances are provided in a sequential manner under changing environment,
smaller weights are often assigned to older instances for imposing some `forgetting' effect
\citep{nn:Murata+Kawanabe+Ziehe:2002,Cao03}.

\item\textbf{Heteroscedastic data modeling:}
A supervised learning setup
where the noise level in output values depends on input points
is said to be \emph{heteroscedastic}.
In heteroscedastic data modeling,
larger weights are often assigned to instances with smaller noise variance
\citep{kersting:icml07}.
The traditional \emph{Gauss-Markov theorem}
\citep{book:Albert:1972} forms the basis of this idea.

\item\textbf{Covariate shift adaptation, transfer learning, and multi-task learning:}
A supervised learning situation 
where training and test \emph{inputs} follow different distributions
is called covariate shift.
Under covariate shift,
using the \emph{importance} (the ratio of the test and training input densities)
as instance weights assures the consistency of estimators
\citep{JSPI:Shimodaira:2000}.
Similar importance-weighting ideas can be applied also to
transfer learning
(where data in one domain is transferred to another domain) \citep{ACL:Jiang+Zhai:2007}
and multi-task learning 
(where multiple learning problems are solved simultaneously
by sharing training instances) \citep{ICML:Bickel+etal:2008}.

\item\textbf{Learning to rank and ordinal regression:}
The goal of ranking (a.k.a.~ordinal regression) is 
to give an ordered list of items based on their relevance
\citep{Herbrich00,Liu09}.
In practical ranking tasks such as information retrieval,
users are often not interested in the entire ranking list,
but only in the top few items.
In order to improve the prediction accuracy in the top of the list,
larger weights are often assigned to higher-ranked items \citep{Xu06}.

\item\textbf{Transduction and semi-supervised learning:}
Transduction is a supervised learning setup where
the goal is not to learn the entire input-output mapping,
but only to estimate the output values for pre-specified unlabeled input points
\citep{book:Vapnik:1995}.
A popular approach to transduction is to 
label the unlabeled samples using the current estimator,
and then modify the estimator using the `self-labeled' samples \citep{Joachims99,raina:icml07}.
In this procedure, smaller weights are usually assigned to the self-labeled samples
than the originally-labeled samples due to their high uncertainty.

\end{itemize}

A common challenge in the research of instance-weighted learning has been to
overcome the computational issue.
In many of these tasks, 
instance weights often change dynamically or adaptively,
and thus the instance-weighted solutions must be repeatedly computed.
For example, 
in on-line learning, 
every time when a new instance is observed, 
all the instance weights must be updated 
in such a way that newer instances have larger weights 
and older instances have smaller weights. 
Model selection in instance-weighted learning also poses a considerable computational burden. 
In many of the above scenarios, 
we only have qualitative knowledge about instance weights. 
For example, 
in the aforementioned ranking problem, 
we only know that higher-ranked items should have larger weights than lower-ranked items, 
but it is often difficult to know 
how large or small these weights should be.
The problem of selecting the optimal weighting patterns is an instance
of model selection, and many instance-weighted solutions with various weighting
patterns must be computed in the model selection phase.
The goal of this paper is to alleviate the computational bottleneck of
instance-weighted learning.

In this paper, we focus on the \emph{support vector machine} (SVM) \citep{COLT:Boser+etal:1992,mach:Cortes+Vapnik:1995},
which is a popular classification algorithm minimizing a regularized empirical risk:
\begin{align*}
  \min R +C\sum_{i}L_{i},
\end{align*}
where $R$ is a regularization term and
$C\ge0$ controls the trade-off between the regularization effect and the empirical risk minimization.
We consider an instance-weighted variant of SVM,
which we refer to as the \emph{weighted SVM} (WSVM)
\citep{mach:lin+lee+wahba:2002,Lin02,Yang07}:
\begin{align*}
  \min R +\sum_{i}C_{i}L_{i}.
\end{align*}

For ordinary SVM, the \emph{solution path algorithm} was proposed
\citep{Hastie04},
which allows efficient computation of SVM solutions
for all $C$ by utilizing the piecewise-linear structure of the solutions
w.r.t.~$C$.
This technique is known as \emph{parametric programming} in the
optimization community
\citep{UWaterloo:Best:1982,Ritter84,Allgower93,BenBre97},
and has been applied to various machine learning tasks recently
\citep{nips02-AA48,Zhu03,Bach06,Gunter07,Rosset07,Lee07,Sjostrand07,Wang08,Arreola08, Takeuchi09,JMLR:Kanamori+etal:2009};
the \emph{incremental-decremental SVM algorithm}, which efficiently
follows the piecewise-linear solution path
when some training instances are added or removed from the training set,
is also based on the same parametric programming technique
\citep{Cauwenberghs01, Laskov06, Karasuyama09b}.

%\citep{Bach06} analyzed the ROC curve of the two class cost sensitive learning  
%by tracing the path of a cost asymmetry parameter for the false negative and false
%positive.

The solution path algorithms described above have been
developed for problems with a \emph{single} hyper-parameter. 
Recently, attention has been paid to studying solution-path tracking in
two-dimensional hyper-parameter space. 
For example, \citet{Wang08} 
developed a path-following algorithm for 
regularization parameter $C$ and an insensitive zone thickness
$\veps$ in \emph{support vector regression}
\citep{Vapnik96,Mattera99,Muller99}.
\citet{Rosset09} studied a path-following algorithm for 
regularization parameter $\lambda$ and quantile parameter $\tau$ in
\emph{kernel quantile regression}~\citep{Takeuchi06}.
However, these works are highly specialized to specific problem structure
of bivariate path-following, and it is not straightforward to extend them to
more than two hyper-parameters.
Thus, the existing approaches may not be applicable to
path-following of WSVM, which contains $n$-dimensional
instance-weight parameters $\bm{c} = [C_1, \ldots, C_n]^\top$,
where $n$ is the number of training instances.

% On the other hand, WSVM can be regarded as containing multiple `regularization' parameters.
% Based on this interpretation, 
In order to go beyond the limitation of the existing approaches,
we derive a general solution path algorithm for efficiently
computing the solution path of \emph{multiple} instance-weight parameters  $\bm{c}$ in WSVM.
This extension involves a significant
problem that
 \emph{breakpoints} (at which the solution path turns) have to be identified in
 high-dimensional space. To facilitate this, we introduce a parametric
 representation of instance weights. We
 also provide a geometric interpretation in weight space using a notion of
 \emph{critical region}
from the studies of \emph{multi-parametric programming} \citep{Gal72, Pistikopoulos07}.
%: a polyhedron in which the current affine solution remains to be optimal.
%  technical difficulty
% that \emph{breakpoints} at which the solution path turns
% have to be identified in general high-dimensional space.}
% To facilitate this, we introduce a notion of \emph{critical region}
%from the studies of \emph{multi-parametric programming} \citep{Gal72, Pistikopoulos07}.
A critical region is a polyhedron in which the current \emph{affine} solution
remains to be optimal (see \figurename~\ref{fig:exmp_path}).
This enables us to find breakpoints at intersections
of the solution path and the boundaries of polyhedrons.

% As we see later in experiments (Section~\ref{sec:applications}), 
% such situation is very common in many applications of instance-weighted learning problem.

This paper is organized as follows.
Section~\ref{sec:wsvm} reviews the definition of WSVM 
and its optimality conditions.
Then we derive the path-following algorithm for WSVM in Section~\ref{sec:wsvm_path}.
Section~\ref{sec:applications} is devoted to 
experimentally illustrating advantages of our algorithm
on a toy problem, on-line time-series analysis and covariate shift adaptation.
Extensions to regression, ranking, and transduction scenarios are discussed
in Section~\ref{sec:wsvm_extension}.
Finally, we conclude in Section~\ref{sec:conclusion}.

%%% Local Variables: 
%%% mode: latex
%%% TeX-master: "paper"
%%% End: 

%\clearpage

%==================================================
% Sec::Weighted SVM 
%==================================================

\section{Problem Formulation} \label{sec:wsvm}
In this section, we review the definition of 
the \emph{weighted support vector machine} (WSVM)
and its optimality conditions.
For the moment, we focus on binary classification scenarios.
Later in Section~\ref{sec:wsvm_extension},
we extend our discussion to more general scenarios
such as regression, ranking, and transduction.

\subsection{WSVM} 
Let us consider a binary classification problem.
Denote $n$ training instances as
$\{ (\bm{x}_i, y_i) \}_{i=1}^n$,
where $\bm{x}_i \in \cX \subseteq \mathbb{R}^{p}$ 
is the input and
$y_i \in \{-1, +1\}$ is the output label.

SVM \citep{COLT:Boser+etal:1992,mach:Cortes+Vapnik:1995}
is a learning algorithm of
a linear decision boundary
\begin{align*}
  f(\bm{x}) = \mb{w}^\top \Phi(\bm{x}) + b
\end{align*}
in a feature space ${\cal F}$,
where 
$\Phi: {\cal X} \rightarrow {\cal F}$
is a map from the input space $\cX$
to the feature space $\cF$,
$\mb{w} \in {\cal F}$ is a coefficient vector, 
$b \in \mathbb{R}$ is a bias term,
and ${}^\top$ denotes the transpose. 
The parameters $\mb{w}$ and $b$ are learned as
\begin{align}
 \min_{\mb{w},b} \frac{1}{2} \| \mb{w} \|_2^2 
 + C \sum_{i=1}^n  [1 - y_i f(\bm{x}_i)]_+,
  \label{eq:reg_risk}
\end{align}
where 
$\frac{1}{2} \| \mb{w} \|_2^2$ is the regularization term,
$\|\cdot\|$ denotes the Euclidean norm,
$C$ is the trade-off parameter,
and
\begin{align*}
  [z]_+=\max\{0,z\}.
\end{align*}
$[1 - y_i f(\bm{x}_i)]_+$ is the so-called \emph{hinge-loss}
for the $i$-th training instance.

WSVM is an extension of the ordinary SVM
so that each training instance possesses its own weight
\citep{mach:lin+lee+wahba:2002,Lin02,Yang07}:
\begin{align}
\min_{\mb{w},b}~~
 \frac{1}{2} \| \mb{w} \|_2^2 
  + \sum_{i=1}^n  C_{i} [1 - y_i f(\bm{x}_i)]_+,
 \label{eq:weighted_reg_risk}
\end{align}
where $C_{i}$ is the weight for the $i$-th training instance.
WSVM includes the ordinary SVM as a special case
when $C_i=C$ for $i=1,\ldots,n$.
The primal optimization problem 
(\ref{eq:weighted_reg_risk}) is expressed as 
the following quadratic program:
\begin{eqnarray}
  \begin{split}
\min_{\mb{w}, b, \{\xi_i\}_{i=1}^n} ~~& \frac{1}{2} \| \mb{w} \|_2^2 
 + \sum_{i=1}^n C_i \xi_i, \\
{\rm s.t. }        ~~& y_i f(\bm{x}_i) \geq 1 - \xi_i, 
		 \  \xi_i \geq 0, \ i = 1, \ldots, n.
               \end{split}
               \label{eq:svm_primal}
\end{eqnarray}

The goal of this paper is to derive an algorithm
that can efficiently compute the 
sequence of WSVM solutions 
for arbitrary weighting patterns of 
$\mb{c} = [C_1, \ldots, C_n]^\top$. 

\subsection{Optimization in WSVM} 
Here we review basic optimization issues of WSVM
which are used in the following section.

Introducing Lagrange multipliers $\alpha_i \geq 0$ and $\rho_i \geq 0$,
we can write the \emph{Lagrangian} of \eqref{eq:svm_primal} as
\begin{align}
L =  \frac{1}{2} \| \mb{w} \|^2 + \sum_{i=1}^{n} C_i \xi_i 
   - \sum_{i=1}^n \alpha_i \{ y_i f(\bm{x}_i) - 1 + \xi_i \} 
      - \sum_{i=1}^n \rho_i \xi_i. \label{eq:lagrange}
\end{align}
Setting the derivatives of the above Lagrangian w.r.t.~the primal variables $\mb{w}$, $b$, and
$\xi_i$ to zero, we obtain
\begin{eqnarray*}
\pd{L}{\mb{w}} = \mb{0} & \Leftrightarrow & \mb{w} 
 = \sum_{i=1}^n \alpha_i y_i \Phi(\bm{x}_i),  \\
 \pd{L}{b} = 0 & \Leftrightarrow & \sum_{i=1}^n \alpha_i y_i = 0,  \\
 \pd{L}{\xi_i} = 0 & \Leftrightarrow & \alpha_i = C_i - \rho_i, \ i = 1, \ldots, n,
\end{eqnarray*}
where $\mb{0}$ denotes the vector with all zeros.
Substituting these equations into (\ref{eq:lagrange}), we arrive at the
following dual problem:
\begin{eqnarray}
  \begin{split}
 \max_{ \{ \alpha_i \}_{i=1}^n } ~~&
  - \frac{1}{2} \sum^n_{i=1} \sum^n_{j=1} \alpha_i \alpha_j Q_{ij}
  + \sum_{i = 1}^n \alpha_i\\
 {\rm s.t. } ~~&
  \sum_{i = 1}^n y_i \alpha_i = 0, \ 
  0 \leq \alpha_i \leq C_i,
\end{split}
 \label{eq:svm_dual}
\end{eqnarray}
where
\begin{align*}
  Q_{ij}=y_i y_j K(\bm{x}_i,\bm{x}_j),
\end{align*}
and
$K(\bm{x}_i, \bm{x}_j) = \Phi(\bm{x}_i)^T \Phi(\bm{x}_j)$ is
a \emph{reproducing kernel} \citep{AMS:Aronszajn:1950}.
%The final solution, i.e., 
The discriminant function $f:{\cal X} \rightarrow {\mathbb R}$
is represented in the following form:
\begin{eqnarray*}
 f(\bm{x}) = \sum_{i=1}^{n} \alpha_i y_i K(\bm{x}, \bm{x}_i) + b.
\end{eqnarray*}

The optimality conditions
of the dual problem \eqref{eq:svm_dual},
called the \emph{Karush-Kuhn-Tucker (KKT) conditions}
\citep{book:Boyd+Vandenberghe:2004},
are summarized as follows:
\begin{subequations}
\begin{eqnarray}
  y_i f(\bm{x}_i) \geq 1, & {\rm if} &  \alpha_i = 0, 
  \label{eq:yf_gt_1} \\
  y_i f(\bm{x}_i) = 1, & {\rm if} &  0 < \alpha_i < C_i, 
  \label{eq:yf_eq_1} \\
  y_i f(\bm{x}_i) \leq 1, & {\rm if} &  \alpha_i = C_i, 
  \label{eq:yf_lt_1} \\
  \sum_{i = 1}^n y_i \alpha_i = 0.
  \label{eq:eq_const}
\end{eqnarray}
\label{eq:kkt}
\end{subequations}

\begin{figure}
 \centering
 \includegraphics[width=3in]{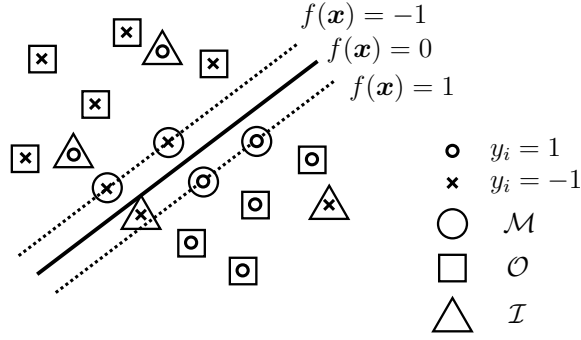}
 \caption{The partitioning of the data points in SVM.}
 \label{fig:svm_sv}
\end{figure}

We define the following three index sets for later use:
\begin{subequations}
 \begin{align}
  {\cal O} &= \{ i \mid  \alpha_i = 0\},
   \label{eq:outside_set} \\
  {\cal M} &= \{ i \mid  0 < \alpha_i < C_i \}, 
  \label{eq:margin_set} \\
  {\cal I} &= \{ i \mid  \alpha_i = C_i \},
  \label{eq:inside_set} 
 \end{align}
\end{subequations}
where $\cal O$, $\cal M$, and $\cal I$ stand for
`Outside the margin' ($y_i f(\bm{x}_i) \geq 1$),
`on the Margin' ($y_i f(\bm{x}_i) = 1$),
and
`Inside the margin' ($y_i f(\bm{x}_i) \leq 1$), respectively
(see \figurename~\ref{fig:svm_sv}).

In what follows, the subscript by an index set such as 
$\mb{v}_{\cal I}$ for a vector $\mb{v} \in {\Bbb R}^n$
indicates a
sub-vector of $\mb{v}$ whose elements are indexed by 
${\cal I}$.
For example, for $\mb{v}=(a,b,c)^\top$ and ${\cal I}=\{1,3\}$, $\mb{v}_{\cal I}=(a,c)^\top$.
 Similarly, the subscript by two index sets such as 
$\vM_{{\cal M},{\cal O}}$ for a matrix $\vM \in {\Bbb R}^{n \times n}$
denotes a sub-matrix whose rows and columns are indexed by ${\cal M}$
and ${\cal O}$, respectively.
The principal sub-matrix such as 
$\vM_{{\cal M},{\cal M}}$
is abbreviated as $\vM_{\cal M}$.

%%% Local Variables: 
%%% mode: latex
%%% TeX-master: "paper"
%%% End: 

%\clearpage

%==================================================
% Sec::Path Following for WSVM
%==================================================
\section{Solution-Path Algorithm for WSVM} \label{sec:wsvm_path}
The path-following algorithm for the ordinary SVM \citep{Hastie04}  
computes the entire solution path for the single regularization parameter
$C$.
In this section, we develop
a path-following algorithm for the vector of weights 
$\mb{c} = [C_1, \ldots, C_n]^\top$. 
Our proposed algorithm keeps track of the optimal $\alpha_i$ and $b$ 
when the weight vector $\mb{c}$ is changed.

\subsection{Analytic Expression of WSVM Solutions}
Let \begin{align*}
\valpha =
\begin{bmatrix}
  \alpha_1\\ \vdots\\ \alpha_n
\end{bmatrix},\ 
\vy =
\begin{bmatrix}
  y_1\\ \vdots\\ y_n
\end{bmatrix},\ 
\mbox{ and }
\vQ=
\begin{bmatrix}
  Q_{11}&\cdots&Q_{1n}\\
  \vdots&\ddots&\vdots\\
  Q_{n1}&\cdots&Q_{nn}\\
\end{bmatrix}.
\end{align*}
Then, using the index sets (\ref{eq:margin_set}) and (\ref{eq:inside_set}), 
we can expand one of the KKT conditions, (\ref{eq:yf_eq_1}), as 
% \begin{eqnarray}
% y_i f(\vx_i) = \sum_{j \in {\cal M}} Q_{ij} \alpha_j 
%  + \sum_{j \in {\cal I}} Q_{ij} C_j + y_i b = 1, && i \in {\cal M}.
%  \label{eq:kkt_linear_system1'}
% \end{eqnarray}
% \eqref{eq:kkt_linear_system1'} can be compactly expressed as
\begin{align}
 \vQ_{\cal M} \valpha_{\cal M} + 
 \vQ_{\cal M, \cal I} \mb{c}_{\cal I} + \vy_{\cal M} b = \mb{1},
 \label{eq:kkt_linear_system1}
\end{align}
where $\mb{1}$ denotes the vector with all ones.
Similarly, another KKT condition \eqref{eq:eq_const} is expressed as
\begin{align}
 \vy_{\cal M}^\top \valpha_{\cal M} 
  + \vy_{\cal I}^\top \mb{c}_{\cal I} = 0.
 \label{eq:kkt_linear_system2}
\end{align}

Let
\begin{eqnarray*}
 \vM =
  \begin{bmatrix}
   0 & \vy_{\cal M}^\top \\
   \vy_{\cal M} & \vQ_{\cal M} 
  \end{bmatrix}.
\end{eqnarray*}
Then \eqref{eq:kkt_linear_system1} and \eqref{eq:kkt_linear_system2}
can be compactly expressed as
the following system of $|{\cal M}| + 1$ linear equations,
where $|{\cal M}|$ denotes the number of elements in the set $\cal M$:
\begin{align}
 \vM 
  \begin{bmatrix}
   b \\
   \valpha_{\cal M}
  \end{bmatrix}
 + 
  \begin{bmatrix}
   \vy_{\cal I}^\top \\
   \vQ_{{\cal M},{\cal I}} 
  \end{bmatrix} 
 \mb{c}_{\cal I}
 =
  \begin{bmatrix}
   0 \\
   \mb{1}
  \end{bmatrix}.
 \label{eq:kkt_linear_system}
\end{align}
Solving (\ref{eq:kkt_linear_system}) w.r.t.~$b$ and $\valpha_{\cal M}$,
we obtain
\begin{align}
  \begin{bmatrix}
   b \\
   \valpha_{\cal M}
  \end{bmatrix}
  =
 - \vM^{-1}
  \begin{bmatrix}
   \vy_{\cal I}^\top \\
   \vQ_{{\cal M},{\cal I}} 
  \end{bmatrix} 
 \mb{c}_{\cal I}
 + \vM^{-1}
  \begin{bmatrix}
   0 \\
   \mb{1}
  \end{bmatrix}, \label{eq:affine_solution}
\end{align}
where we implicitly assumed that $\vM$ is invertible\footnote{
The invertibility of the matrix $\vM$
is assured if and only if
the submatrix $\vQ_{\cal M}$ is positive definite in the subspace 
$\{\mb{z} \in \mathbb{R}^{|{\cal M}|} \mid \vy_{\cal M}^\top \mb{z} = 0\}$.
We assume this technical condition here.
A notable exceptional case is that $\cal M$ is empty---we
will discuss how to cope with this case
in detail in Section~\ref{subsec:empty-margin}.
}.
Since $b$ and $\valpha_{\cal M}$ are \emph{affine} w.r.t.~$\mb{c}_{\cal I}$,
we can calculate the change of $b$ and $\valpha_{\cal M}$ by (\ref{eq:affine_solution})
as long as the weight vector $\mb{c}$ is changed continuously.
By the definition of $\cal I$ and $\cal O$,
the remaining parameters $\valpha_{\cal I}$ and $\valpha_{\cal O}$ are merely given by
\begin{align}
 \valpha_{\cal I} &= \mb{c}_{\cal I}, \label{eq:alpha_inside}\\
 \valpha_{\cal O} &= \mb{0}. \label{eq:alpha_outside}
\end{align}

A change of the index sets ${\cal M}$, ${\cal O}$, and ${\cal I}$ is called an \emph{event}.
As long as no event occurs,
the WSVM solutions for all $\mb{c}$
can be computed by (\ref{eq:affine_solution})--(\ref{eq:alpha_outside})
since all the KKT conditions (\ref{eq:yf_gt_1})--(\ref{eq:eq_const})
are still satisfied.
However, when an event occurs, we need to check the violation of the KKT conditions.
Below, we address the issue of event detection when $\mb{c}$ is changed.

\subsection{Event Detection}
\begin{figure}[t]
 \centering
 \includegraphics[width=2.5in]{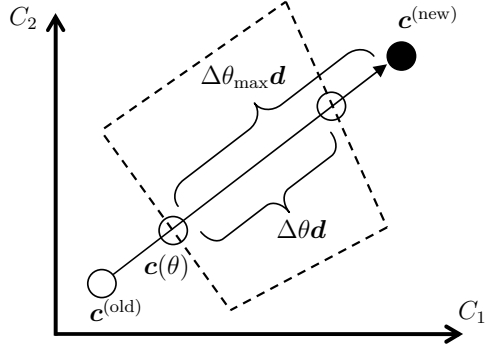}
 \caption{
 The schematic illustration of path-following in the space of 
 $\mb{c} \in \mathbb{R}^2$,
 where the WSVM solution is updated from $\mb{c}^{({\rm old})}$ to $\mb{c}^{({\rm new})}$.
 Suppose we are currently at $\mb{c}(\theta)$.
 The vector $\mb{d}$ represents the update direction 
 $\mb{c}^{({\rm new})} - \mb{c}^{({\rm old})}$,
 and the polygonal region enclosed by dashed lines indicates the current critical region. 
 Although $\mb{c}(\theta) + \Delta \theta_{\max} \mb{d}$ seems to
 directly lead the solution to $\mb{c}^{({\rm new})}$,
 the maximum possible update from $\mb{c}(\theta)$ is $\Delta \theta \mb{d}$;
 otherwise the KKT conditions are violated.
 To go beyond the border of the critical region,
 we need to update the index sets ${\cal M}$, ${\cal I}$, and ${\cal O}$
 to fulfill the KKT conditions.
 }
 \label{fig:exmp_path}
\end{figure}

Suppose we want to change the weight vector 
from $\mb{c}^{({\rm old})}$ to $\mb{c}^{({\rm new})}$
(see \figurename~\ref{fig:exmp_path}).
This can be achieved by moving the weight vector $\mb{c}^{({\rm old})}$
toward the direction of $\mb{c}^{({\rm new})} - \mb{c}^{({\rm old})}$.

Let us write the line segment between 
$\mb{c}^{({\rm old})}$ and
$\mb{c}^{({\rm new})}$ in the following parametric form
\begin{eqnarray*}
 \mb{c}(\theta) = \mb{c}^{({\rm old})} 
  + \theta \left( \mb{c}^{({\rm new})} - \mb{c}^{({\rm old})} \right),
  \; 
  \theta \in [0, 1],
\end{eqnarray*} 
where $\theta$ is a parameter.
This parametrization allows us to derive a path-following
algorithm between arbitrary 
$\mb{c}^{({\rm old})}$ and
$\mb{c}^{({\rm new})}$ by considering the change of the solutions when
$\theta$ is moved from $0$ to $1$.
Suppose we are currently at $\mb{c}(\theta)$ on the path, 
and the current solution is $(b, \valpha)$.
Let
\begin{eqnarray}
 \mb{\Delta} \mb{c} = 
  \Delta \theta \left( \mb{c}^{({\rm new})} - \mb{c}^{({\rm old})} \right),
  \:
  \Delta \theta \ge 0,
  \label{eq:delta_c}
\end{eqnarray}
where
the operator $\Delta$ represents the amount of change of each
variable from the current value.
If $\Delta \theta$ is increased from $0$, we may encounter a point at which
some of the KKT conditions (\ref{eq:yf_gt_1})--(\ref{eq:yf_lt_1}) do not hold.
This can be checked by investigating the following conditions.
\begin{equation}
\left\{
  \begin{array}{r@{\ }lr@{\ }l}
  y_i f(\vx_i) + y_i \Delta f(\vx_i) &\geq 1,& i &\in {\cal O}, \\
  \alpha_i + \Delta \alpha_i &> 0,& i &\in {\cal M}, \\
  \alpha_i + \Delta \alpha_i -(C_i + \Delta C_i,)&<0,&  i &\in {\cal M}, \\
  y_i f(\vx_i) + y_i \Delta f(\vx_i) &\leq 1,& i &\in {\cal I}.
  \end{array}
\right.
 \label{eq:event_ineqs}
\end{equation}
% where the operator $\Delta$ represents the amount of change of each
% variable from the current value.
The set of inequalities \eqref{eq:event_ineqs} defines a convex polyhedron,
called a \emph{critical region} in the multi-parametric programming literature 
\citep{Pistikopoulos07}.
The event points lie on the border of critical regions,
as illustrated in \figurename~\ref{fig:exmp_path}.

We detect an event point 
by checking the conditions \eqref{eq:event_ineqs} along the solution path
as follows.
Using (\ref{eq:affine_solution}), we can express the changes of $b$ and 
$\valpha_{\cal M}$ as
\begin{eqnarray}
 \left[ 
  \begin{array}{c}
   \Delta b \\
   \mb{\Delta} \valpha_{\cal M}
  \end{array}
 \right] =
 \Delta \theta \mb{\phi}, \label{eq:delta_b_alpha}
\end{eqnarray}
where 
\begin{eqnarray}
 \mb{\phi} = - \vM^{-1}
 \left[ 
  \begin{array}{c}
   \vy_{\cal I}^\top \\
   \vQ_{{\cal M},{\cal I}} 
  \end{array} 
 \right]
 (\mb{c}_{\cal I}^{({\rm new})} - \mb{c}_{\cal I}^{({\rm old})}).
 \label{eq:phi}
\end{eqnarray}
Furthermore, $y_i \Delta f(\vx_i)$ is expressed as
\begin{align}
 y_i \Delta f(\vx_i) &= 
 \left[ 
  \begin{array}{cc}
   y_i & \vQ_{i,{\cal M}}
  \end{array}
 \right]
 \left[ 
  \begin{array}{c}
   \Delta b \\
   \mb{\Delta} \valpha_{\cal M}
  \end{array}
 \right]
 + 
 \vQ_{i, {\cal I}} \mb{\Delta} \mb{c}_{\cal I} \nonumber \\ 
 &= \Delta \theta \psi_i, \label{eq:y_delta_f}
\end{align}
where
\begin{eqnarray}
 \psi_i = 
 \left[ 
  \begin{array}{cc}
   y_i & \vQ_{i,{\cal M}}
  \end{array}
 \right] \mb{\phi} +
 \vQ_{i, {\cal I}}
(\mb{c}_{\cal I}^{({\rm new})} - \mb{c}_{\cal I}^{({\rm old})}).
\label{eq:psi}
\end{eqnarray}
Let us denote the elements of the index set ${\cal M}$ as
\begin{align*}
  {\cal M} = \{m_1, \ldots, m_{|{\cal M}|}\}.
\end{align*}
Substituting (\ref{eq:delta_b_alpha}) and (\ref{eq:y_delta_f}) into 
the inequalities (\ref{eq:event_ineqs}), we can obtain the maximum step-length
with no event occurrence as
% \begin{eqnarray*}
%  \eta = 
%   \min_{i \in \{1, \ldots, |{\cal M}|\}, j \in {\cal I} \cup {\cal O}}
%   \left\{
%    \left[- \frac{\alpha_{m_i}}{{\phi}_{i + 1}}\right]_+,
%    \left[\frac{C_{m_i} - \alpha_{m_i}}{{\phi}_{i + 1} - d_{m_i}}\right]_+,
%    \left[\frac{1 - y_j f(\vx_j)}{\psi_j}\right]_+
%   \right\},
% \end{eqnarray*}
\begin{eqnarray}
 \Delta \theta = 
  \min_{i \in \{1, \ldots, |{\cal M}|\}, j \in {\cal I} \cup {\cal O}}
  \left\{
   - \frac{\alpha_{m_i}}{{\phi}_{i + 1}},
   \frac{C_{m_i} - \alpha_{m_i}}{{\phi}_{i + 1} - d_{m_i}},
   \frac{1 - y_j f(\vx_j)}{\psi_j}
  \right\}_{+},
  \label{eq:min_eta}
\end{eqnarray}
where ${\phi}_{i }$ denotes the $i$-th element of $\mb{\phi}$
and $d_i = C_i^{({\rm new})} - C_i^{({\rm old})}$.
We used $\min_{i} \{z_i\}_{+}$ as a simplified notation of 
$\min_{i} \{z_i \mid z_i \geq 0\}$.
Based on this largest possible $\Delta \theta$, we can compute $\valpha$ and $b$
along the solution path by \eqref{eq:delta_b_alpha}.

At the border of the critical region,
we need to update the index sets
${\cal M}$, ${\cal O}$, and ${\cal I}$.
For example, if $\alpha_i$ ($i \in {\cal M}$) reaches $0$, 
we need to move the element $i$ from ${\cal M}$ to ${\cal O}$.
Then the above path-following procedure is carried out again
for the next critical region specified by the
updated index sets ${\cal M}$, ${\cal O}$, and ${\cal I}$,
and this procedure is repeated until $\mb{c}$ reaches $\mb{c}^{({\rm new})}$.

\subsection{Empty Margin}\label{subsec:empty-margin}
In the above derivation,
we have implicitly assumed that the index set ${\cal M}$ is not empty---when 
$\cal M$ is empty, we can not use (\ref{eq:delta_b_alpha})
because $\vM^{-1}$ does not exist.

When $\cal M$ is empty,
the KKT conditions (\ref{eq:kkt}) can be re-written as
\begin{subequations}
\begin{eqnarray}
 \sum_{j \in {\cal I}} Q_{ij} C_j + y_i b \geq 1, && i \in {\cal O}, 
  \label{eq:empty_margin_yf_gt_1} \\
 \sum_{j \in {\cal I}} Q_{ij} C_j + y_i b \leq 1, && i \in {\cal I}, 
  \label{eq:empty_margin_yf_lt_1} \\ 
 \sum_{i \in {\cal I}} y_i C_i = 0. &&
  \label{eq:empty_margin_eq_const} 
\end{eqnarray}
\end{subequations}
Although we can not determine the value of $b$ uniquely
only from the above conditions,
(\ref{eq:empty_margin_yf_gt_1}) and (\ref{eq:empty_margin_yf_lt_1})
specify the range of optimal $b$:
\begin{align}
 \max_{i \in {\cal L}} y_i g_i 
  \leq 
  b 
  \leq 
  \min_{i \in {\cal U}} y_i g_i,
  \label{eq:bias_interval}
\end{align}
where
\begin{align*}
 g_i &= 1 - \sum_{j \in {\cal I}} Q_{ij} C_j,\\
 {\cal L} &= \{ i \mid i \in {\cal O}, y_i = 1 \} \cup 
  \{ i \mid i \in {\cal I}, y_i = -1 \},
 \\
 {\cal U} &= \{ i \mid i \in {\cal O}, y_i = -1 \} \cup 
  \{ i \mid i \in {\cal I}, y_i = 1 \}.
\end{align*}

Let 
\begin{align*}
  \delta \equiv \sum_{i \in {\cal I}} y_i d_i,
\end{align*}
where 
\begin{align*}
  d_i = C_i^{({\rm new})} - C_i^{({\rm old})}.
\end{align*}
When $\delta = 0$, the step size $\Delta \theta$ can be increased
as long as the inequality (\ref{eq:bias_interval}) is satisfied.
Violation of (\ref{eq:bias_interval}) can be checked
by monitoring the upper and lower bounds of the bias $b$
(which are piecewise-linear w.r.t.~$\Delta \theta$) when $\Delta \theta$ is increased
\begin{align}
 \begin{array}{r@{\;}l}
  u(\Delta \theta) =& \max_{i \in {\cal U}} y_i (g_i + \Delta g_i (\Delta \theta)), \\
  \ell(\Delta \theta) =& \min_{i \in {\cal L}} y_i (g_i + \Delta g_i (\Delta \theta)),
 \end{array}
 \label{eq:empty_maring_delta_eq_0}
\end{align}
where
\begin{align*}
 \Delta g_i (\Delta \theta) = - \Delta \theta \sum_{j \in {\cal I}} Q_{ij} d_j.
\end{align*}

On the other hand, when $\delta \neq 0$,
$\Delta \theta$ can not be increased without violating the equality condition 
(\ref{eq:empty_margin_eq_const}).
In this case, 
an instance with index
\begin{align*}
   i_{\rm low} = \argmax_{i \in {\cal L}} y_i g_i 
\end{align*}
or
\begin{align*}
 i_{\rm up} = \argmin_{i \in {\cal U}} y_i g_i 
\end{align*}
actually enters the index set $\cal M$.
% In order to keep (\ref{eq:empty_margin_eq_const}) satisfied,
% we need a new margin data point $m_1$.
% From (\ref{eq:bias_interval}), the data point $m_1$ is either
% \begin{eqnarray*}
%  i_{\rm low} = \argmax_{i \in {\cal L}} y_i g_i \ \mbox{ or } \
%  i_{\rm up} = \argmin_{i \in {\cal U}} y_i g_i.
% \end{eqnarray*}
If the instance (we denote its index by $m$) comes from the index set ${\cal O}$,
the following equation must be satisfied
for keeping (\ref{eq:empty_margin_eq_const}) satisfied:
\begin{align*}
% \eta \delta + \Delta \alpha_{m} y_{m} &= 0.
   \Delta \theta \delta  = -\Delta \alpha_{m} y_{m}.
\end{align*}
Since $\Delta \theta > 0$ and $\Delta \alpha_{m} > 0$, 
we have 
\begin{align*}
  {\rm sign}(\delta)={\rm sign}(- y_{m}).
\end{align*}
On the other hand, if the instance comes from the index set ${\cal I}$,
\begin{align*}
%   \eta (\delta - y_{m} d_{m}) + \Delta \alpha_{m} y_{m} &= 0, \\
 \Delta \theta \delta &= y_{m} (\Delta C_{m} - \Delta \alpha_{m})
\end{align*}
must be satisfied.
Since $\Delta \theta >0$ and $\Delta C_{m} - \Delta \alpha_{m} > 0$, 
we have 
\begin{align*}
  {\rm sign}(\delta) = {\rm sign}(y_{m}).
\end{align*}
Considering these conditions, we arrive at the following updating rules
for $b$ and ${\cal M}$:
\begin{eqnarray}
 \begin{array}{rcl}
  \delta > 0 &\Rightarrow& b 
   = y_{i_{\rm up}} g_{i_{\rm up}}, \ {\cal M} = \{ i_{\rm up} \},
   \\ %
   \delta < 0 &\Rightarrow& b 
   = y_{i_{\rm low}} g_{i_{\rm low}}, \ {\cal M} = \{ i_{\rm low} \}.
 \end{array}
 \label{eq:empty_margin_delta_neq_0}
\end{eqnarray}
Note that we also need to remove $i_{\rm up}$ and $i_{\rm low}$
from $\cal O$ and $\cal I$, respectively.

\subsection{Computational Complexity} \label{subsec:complexity}
The entire pseudo-code of the proposed WSVM path-following algorithm
is described in \figurename~\ref{fig:pseudo-code}.

\begin{figure}[t]
  \centering
% ------------------------------------------------
%  Pseudo-code
% ------------------------------------------------
%\begin{algorithm}
%\hrule \kern1ex
% \vbox{\kern1ex 
% \hrule height 1pt depth 0pt
% \kern1ex
% \noindent
% {\bf Algorithm 1} Multi-parametric solution-path algorithm for
% WSVM
% \hrule}
%
%\algblock[Name]{Argument}{EndArg}
\hrule \kern1ex
\algblockdefx[Arguments]{Arg}{EndArg}%
{\textbf{arguments:}}{\textbf{end arguments}}
\begin{algorithmic}[1]
%\State \textbf{arguments:}
\Arg
\State Optimal parameters $\valpha$ and $b$ for $\mb{c}^{\rm (old)}$
\State Sets ${\cal M}$, ${\cal O}$, ${\cal I}$, and Cholesky factor 
 $\mb{L}$ of $\vQ_{\cal M}$
\State New weight vector $\mb{c}^{\rm (new)}$
\EndArg
\Statex
\Function{WSVM-Path}{$\valpha, b, \mb{c}^{\rm (old)}, {\cal M}, {\cal O}, {\cal I}, \mb{L}, \mb{c}^{\rm (new)}$}
        \State $\theta \leftarrow 0, \ \mb{c} \leftarrow \mb{c}^{\rm (old)}$
        \While{ $\theta \neq 1$ }
        \If{ ${\cal M}$ is empty }
		\State $\Delta \theta \leftarrow$ \textsc{EmptyMargin}
	\Else
		\State Calculate $\mb{\phi}$ by (\ref{eq:phi})
                       using Cholesky factor $\mb{L}$ 
		\State Calculate $\mb{\psi}$ by (\ref{eq:psi}) 
		\State Calculate $\Delta \theta$ by (\ref{eq:min_eta})
	\EndIf
        \State If $\theta + \Delta \theta > 1$, then
                $\Delta \theta \leftarrow 1 - \theta$
	\State Update $\valpha$, $b$, and $\mb{c}$ by step length $\Delta \theta$
	\State $\theta \leftarrow \theta + \Delta \theta$
	\State Update ${\cal M}$, ${\cal O}$, and ${\cal I}$ 
               depending on the event type 
        \State Update $\mb{L}$ (Cholesky factor rank-one update)
\EndWhile
\EndFunction
\Statex
\Function{EmptyMargin}{}
	\If{ $\delta(\valpha) \neq 0$ }
                \State Set bias term $b$ by (\ref{eq:empty_margin_delta_neq_0})
                \State $\Delta \theta \leftarrow 0$
	\Else
		\State Trace $u(\Delta \theta)$ and $\ell(\Delta \theta)$ in 
                (\ref{eq:empty_maring_delta_eq_0}) 
                until $u(\Delta \theta) = \ell(\Delta \theta)$
	\EndIf
        \State \Return $\Delta \theta$
\EndFunction
\end{algorithmic}
%\end{algorithm}
\kern1ex\hrule
\vspace{1em}
  \caption{Pseudo-code of the proposed WSVM path-following algorithm.}
  \label{fig:pseudo-code}
\end{figure}

The computational complexity at 
each iteration of our path-following algorithm is the same as that for the ordinary
SVM (i.e., the single-$C$ formulation) \citep{Hastie04}.
Thus, our algorithm inherits a superior computational property of the
original path-following algorithm.

The update of the linear system (\ref{eq:phi}) 
from the previous one at each event point can be carried out efficiently
with $O(|{\cal M}|^2)$ computational cost
based on the \emph{Cholesky decomposition rank-one update} \citep{Golub96}
or the \emph{block-matrix inversion formula} \citep{Schott05}.
Thus, the computational cost required for identifying the next event point 
is $O(n |{\cal M}|)$.

It is difficult to state the number of iterations needed for complete 
path-following because
the number of events depends on the sensitivity of the model and
the data set.
Several empirical results suggest that the number of events
linearly increases w.r.t.~the data set size \citep{Hastie04,Gunter07,Wang08};
our experimental analysis given in Section~\ref{sec:applications}
also showed the same tendency.
This implies that path-following is computationally highly efficient---indeed,
in Section~\ref{sec:applications},
we will experimentally demonstrate
that the proposed path-following algorithm is faster than
an alternative approach in one or two orders of magnitude.

%%% Local Variables: 
%%% mode: latex
%%% TeX-master: "paper"
%%% End: 

%\clearpage

%==================================================
% Sec::Experiments
%==================================================

\section{Experiments} \label{sec:applications}

In this section,
we illustrate the empirical performance
of the proposed WSVM path-following algorithm 
in a toy example and two real-world applications. 
We compared the computational cost
of the proposed path-following algorithm
with the \emph{sequential minimal optimization} (SMO) 
algorithm \citep{Platt99}
when the instance weights of WSVM are changed in various ways.
In particular, 
we investigated the CPU time of updating solutions from some 
$\mb{c}^{\rm (old)}$ to $\mb{c}^{\rm (new)}$.

In the path-following algorithm,
we assume that the optimal parameter 
$\bm{\alpha}$
as well as the Cholesky factor 
$\bm{L}$ 
of $\bm{Q}_{\cM}$ 
for 
$\bm{c}^{\rm (old)}$
has already been obtained.
In the SMO algorithm, 
we used the old optimal parameter 
$\bm{\alpha}$ 
as the initial starting point (i.e., the `hot' start) 
after making them \emph{feasible}
using the \emph{alpha-seeding strategy}~\citep{DeCoste00}.
We set the tolerance parameter in the termination criterion of SMO to $10^{-3}$.
Our implementation of the SMO algorithm is based on \emph{LIBSVM}
\citep{Chang04}.
To circumvent possible numerical instability,
we added small positive constant $10^{-6}$ to the diagonals
of the matrix $\vQ$.
In all the experiments, we used the Gaussian kernel 
\begin{align}
  K(\bm{x}, \bm{x}') = \exp\left(- \frac{\gamma}{p} \| \bm{x} - \bm{x}'\|^2\right),
  \label{Gaussian-kernel}
\end{align}
where $\gamma$ is a hyper-parameter and $p$ is the dimensionality of $\bm{x}$.

%==================================================
% Toy Example
%==================================================
\subsection{Illustrative Example} \label{subsec:toy_example}
First, 
we illustrate the behavior of the proposed path-following algorithm
using an artificial data set.
Consider a binary classification problem 
with the training set 
$\{(\bm{x}_i, y_i)\}_{i = 1}^n$, 
where 
$\bm{x}_i \in \RR^2$ 
and 
$y_i \in \{-1, +1\}$.
Let us define the sets of indices of positive and negative instances
as
$\cK_{-1} = \{ i | y_i = - 1\}$
and 
$\cK_{+1} = \{ i | y_i = + 1\}$,
respectively.
We assume that the loss function is defined as 
\begin{align}
  \label{eq:toy.loss}
  \sum_{i} v_i I(y_i f(\vx_i) \leq 0),
\end{align}
where 
$v_i \in \{1, 2\}$
is the cost of misclassifying the instance 
$(\bm{x}_i, y_i)$, 
and 
$I(\cdot)$ 
is the indicator function.
Let
$\cD_{1} = \{i | v_i = 1 \}$
and 
$\cD_{2} = \{i | v_i = 2 \}$,
i.e., $\cD_{2}$ is the set of instance indices
which have stronger influence on the overall test error
than $\cD_{1}$.

To be consistent with the above error metric,
it would be natural to assign a smaller weight 
$C_1$ 
for 
$i \in \cD_1$ 
and 
a larger weight 
$C_2$ 
for 
$i \in \cD_2$
when training SVM. 
However, naively setting 
$C_2 = 2 C_1$
is not generally optimal 
because 
the hinge loss is used in SVM training, 
while the 0-1 loss is used in performance evaluation (see \eqref{eq:toy.loss}).
In the following experiments, 
we fixed the Gaussian kernel width to $\gamma = 1$
and the instance weight for $\cD_{2}$ to $C_2 = 10$,
and we changed the instance weight $C_1$ for $\cD_{1}$
from $0$ to $10$.
Thus, the change of the weights is represented as 
\begin{align*}
 \begin{bmatrix}
  \mb{c}^{({\rm old})}_{{\cal D}_1} \\
  \mb{c}^{({\rm old})}_{{\cal D}_2}
 \end{bmatrix}
 =
 \begin{bmatrix}
  \mb{0} \\
  \mb{10} \\ 
 \end{bmatrix}
 \ {\rm and} \
 \begin{bmatrix}
  \mb{c}^{({\rm new})}_{{\cal D}_1} \\
  \mb{c}^{({\rm new})}_{{\cal D}_2}
 \end{bmatrix}
 =
 \begin{bmatrix}
  \mb{10} \\ 
  \mb{10}
 \end{bmatrix}.
%\label{eq:toy_weights}
\end{align*}

\begin{figure}[t]
\centering
 \begin{tabular}{ccc}
  \includegraphics[width=0.3\textwidth]{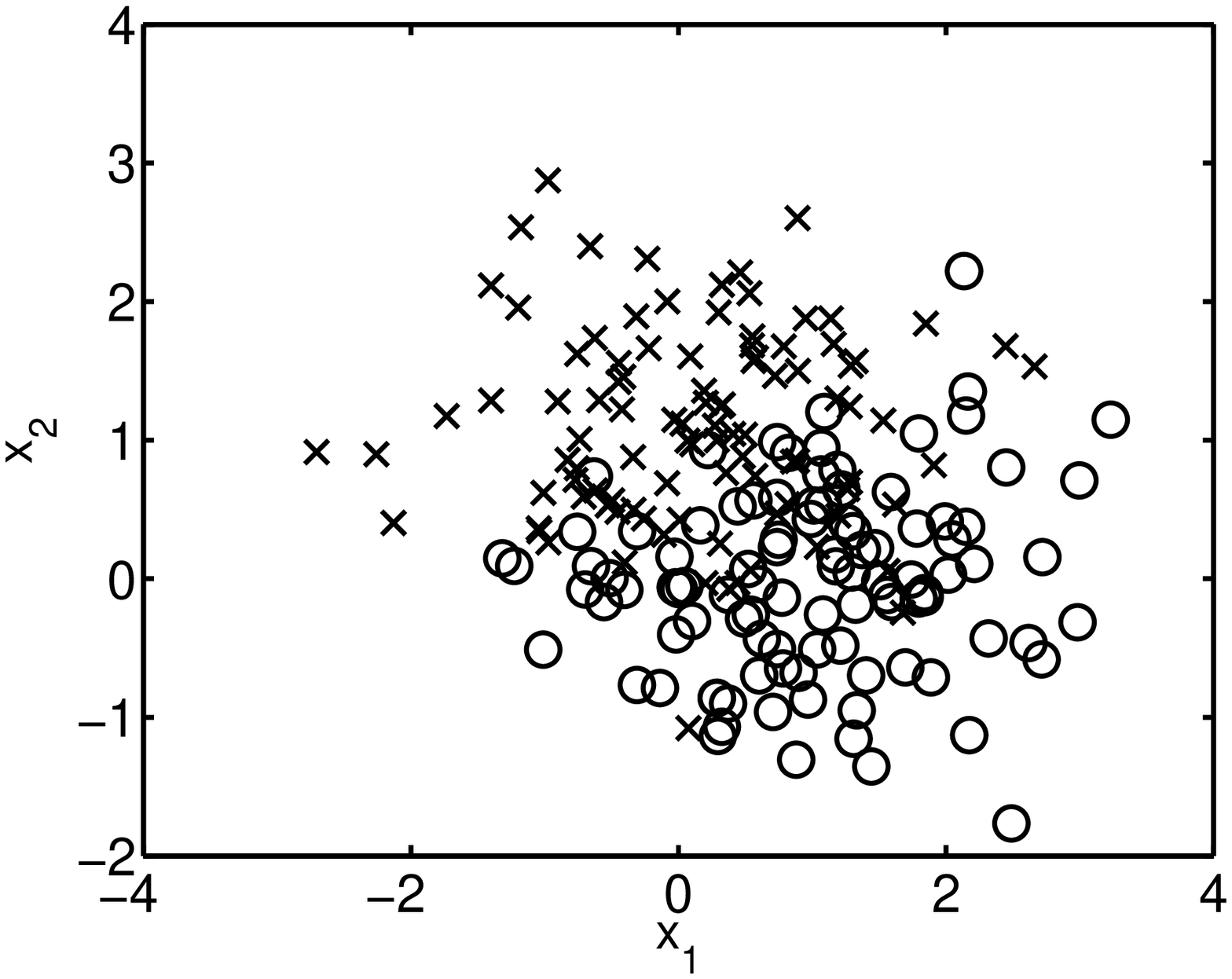} &
  \includegraphics[width=0.3\textwidth]{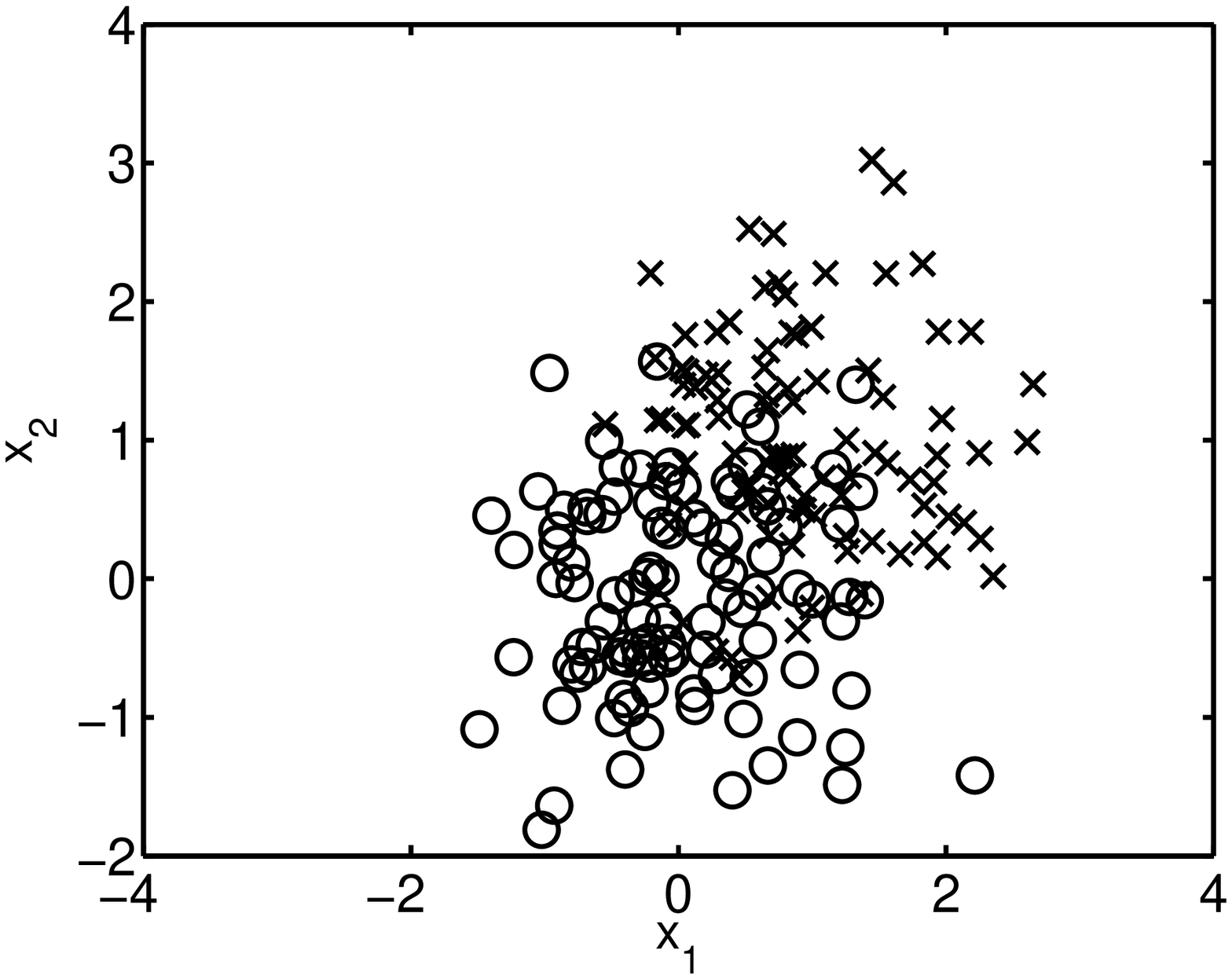} &
  \includegraphics[width=0.3\textwidth]{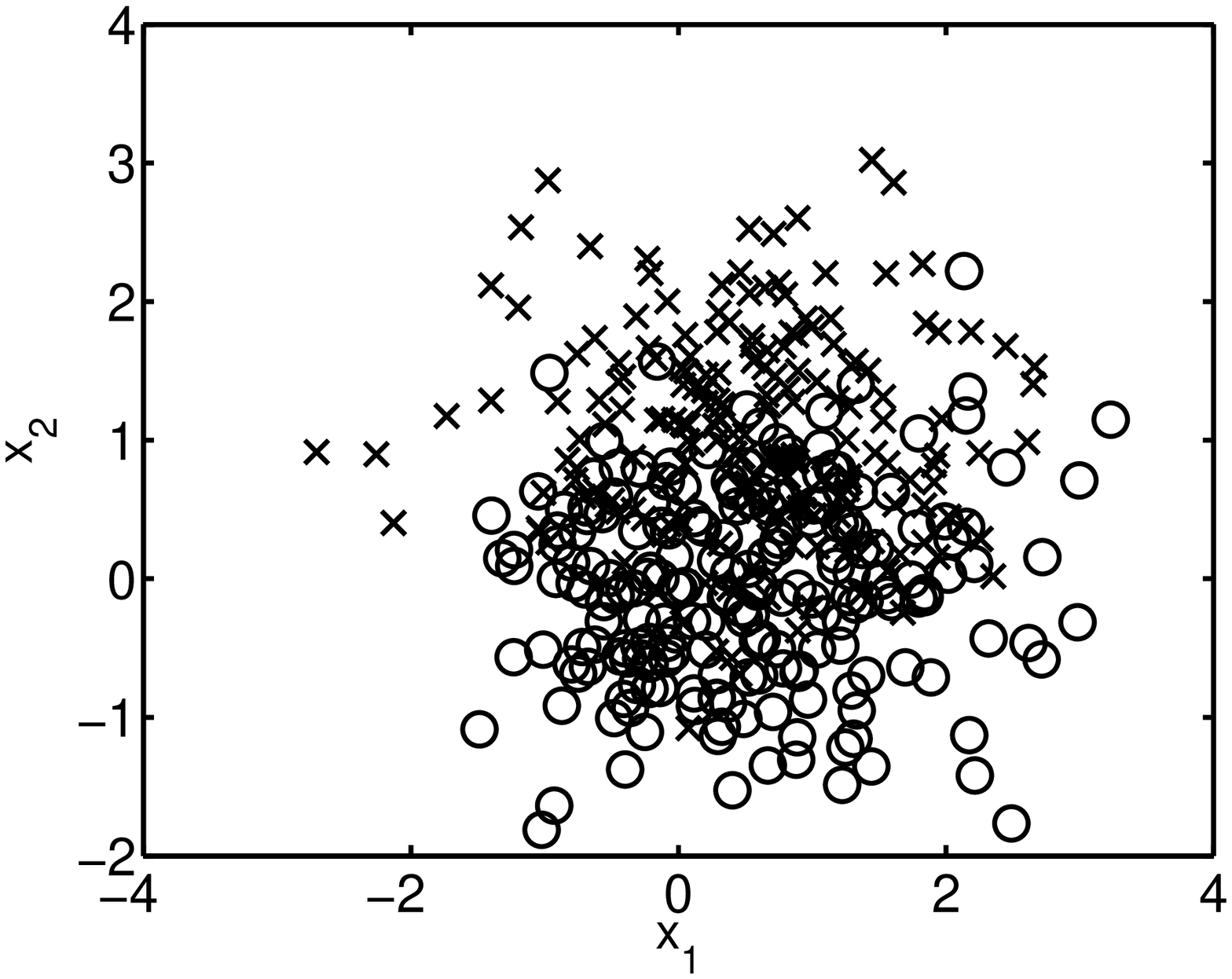} \\
  \footnotesize{Data points in $\cD_1$} &
  \footnotesize{Data points in $\cD_2$} &
  \footnotesize{Data points in $\cD_1 \cup \cD_2$} 
 \end{tabular}
 \caption{
 Artificial data set 
 generated by the distribution \eqref{artifical-data-generative}. 
 The crosses and circles indicate the data points in 
 $\cK_{-1}$ (negative class)
 and 
 $\cK_{+1}$ (positive class),
 respectively.
 The left plot shows the data points 
 in $\cD_1$ (misclassification cost is $1$), 
 the middle plot shows the data points 
 in $\cD_2$ (misclassification cost is $2$), 
 and the right plots show their union.
 }
 \label{fig:toy_data}
% \end{figure}
% %
% %--------------------------------------------------
% % Alpha
% %--------------------------------------------------
% \begin{figure}[t]
%\centering
\vspace*{3mm}
 \begin{tabular}{cc}
  \includegraphics[width=2in]{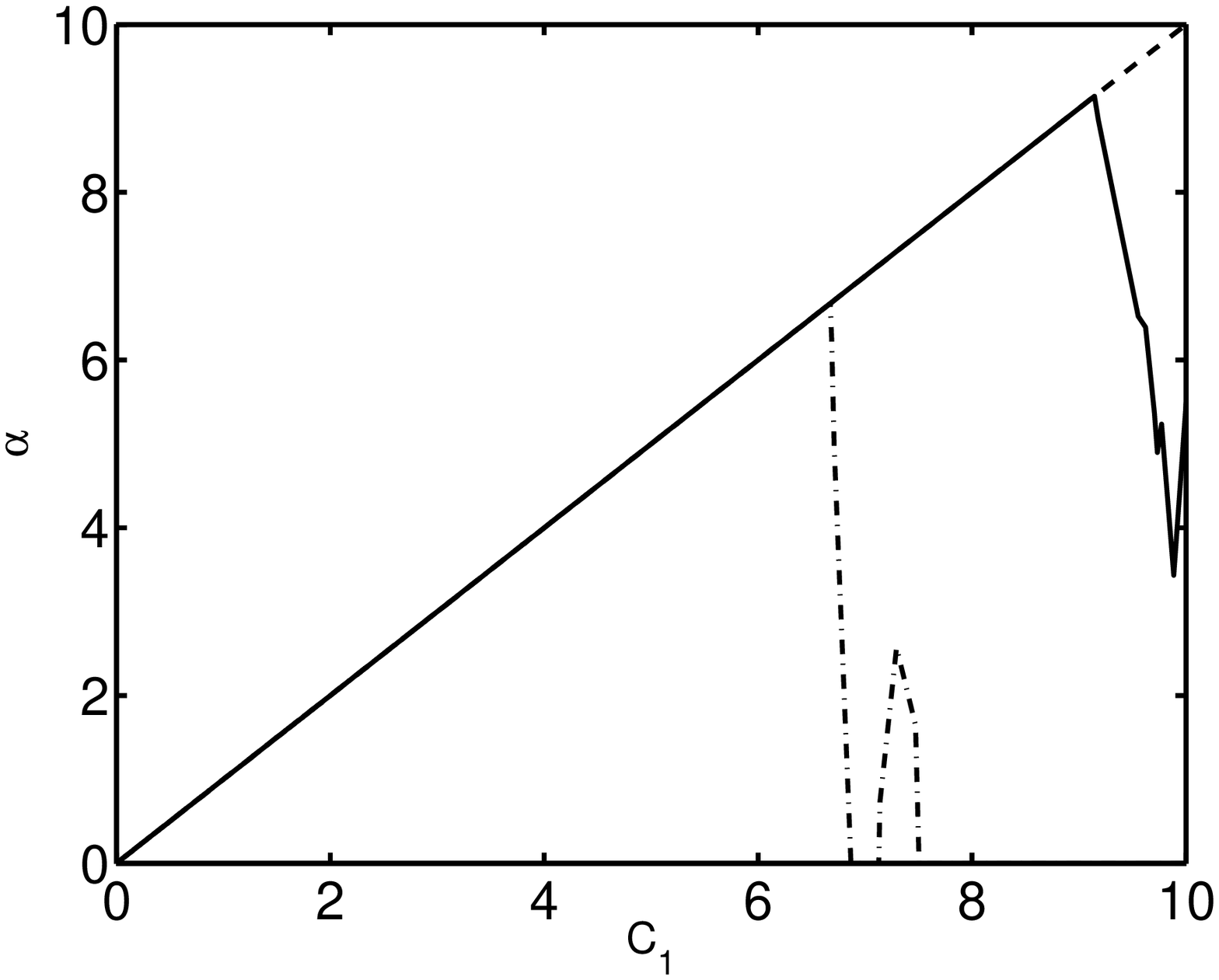} &
  \includegraphics[width=2in]{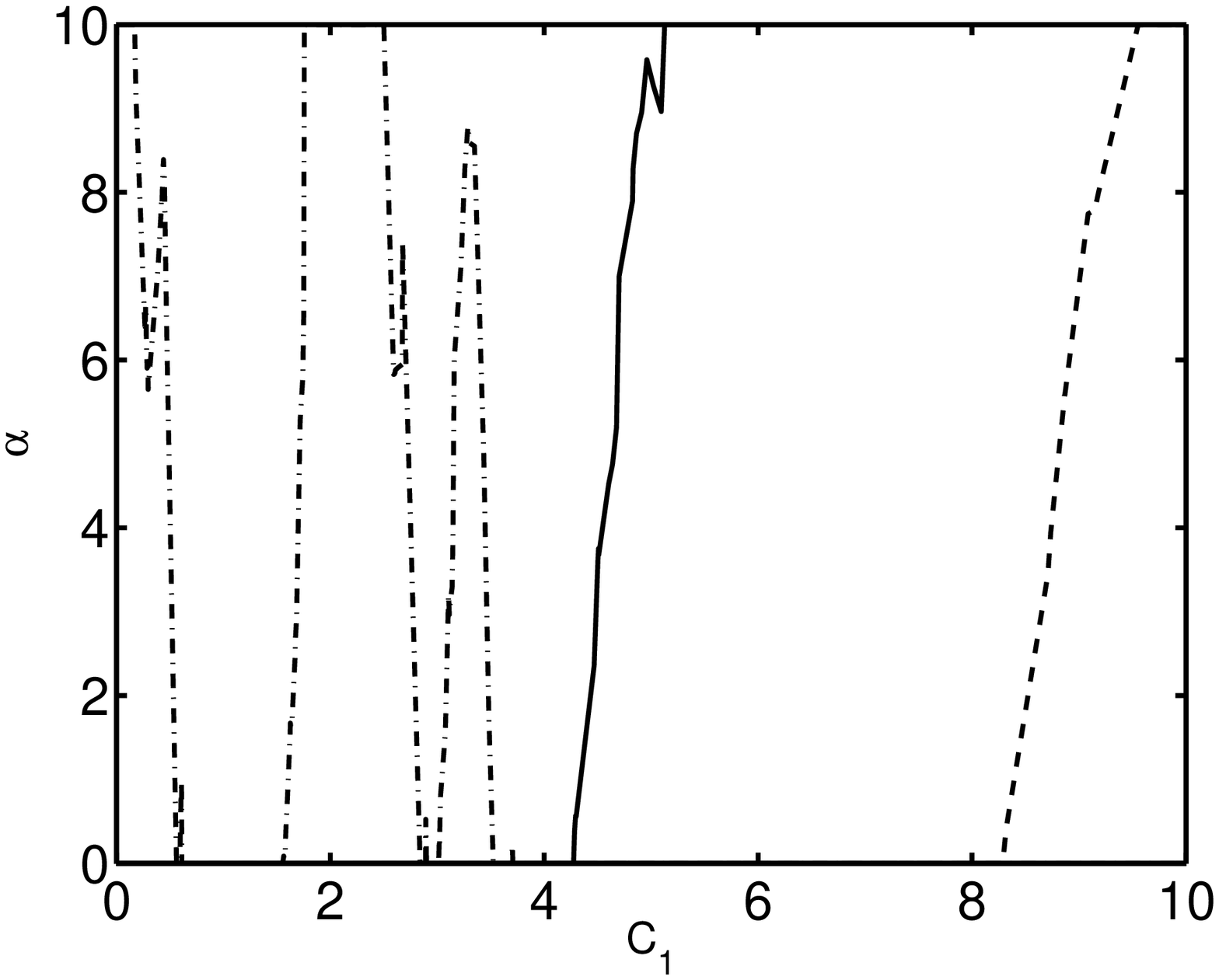} \\
  \footnotesize{$\alpha_i$ for $i \in \cD_1$} &
  \footnotesize{$\alpha_i$ for $i \in \cD_2$} \\
 \end{tabular}
 \caption{
 Examples of piecewise-linear paths of 
 $\alpha_i$ 
 for the artificial data set.
 The weights are changed from 
 $C_i = 0$ 
 to 
 $10$ for $i \in \cD_1$ (for $i \in \cD_2$, $C_i = 10$ is unchanged).
%  and
%  $C_i = 10, i \in \cD_2$
%  to 
%  $C_i = 10, i \in \cD_1 \cup \cD_2$.
 %
 The left and right plots show the paths of three representative
 parameters $\alpha_i$ 
 for 
 $i \in \cD_1$, 
 and 
 for 
 $i \in \cD_2$, 
 respectively.
}
 \label{fig:toy_path}
\end{figure}

The two-dimensional input 
$\{\bm{x}_i\}_{i = 1}^n$ 
were generated from the following distribution:
\begin{align}
\vx_i \sim
 \begin{cases}
 \cN\left(
 \begin{bmatrix}
   1\\ 0
 \end{bmatrix},
   \begin{bmatrix}
     1 & 0 \\ 0 & 0.5 
   \end{bmatrix}
 \right)
 &\mbox{if }i \in \cK_{+1} \cap \cD_1, \\[5mm]
 \cN\left(
 \begin{bmatrix}
   0 \\ 0
 \end{bmatrix},
 \begin{bmatrix}
   0.5 & 0 \\ 0 & 0.5
 \end{bmatrix}
 \right)
&\mbox{if } i \in \cK_{+1} \cap \cD_2, \\[5mm]
 \cN\left(
 \begin{bmatrix}
  0 \\ 1
 \end{bmatrix},
 \begin{bmatrix}
 1 & 0 \\ 0 & 0.5
 \end{bmatrix}
 \right)
&\mbox{if } i \in \cK_{-1} \cap \cD_1,\\[5mm]
 \cN\left(
 \begin{bmatrix}
   1 \\ 1
 \end{bmatrix},
 \begin{bmatrix}
   0.5 & 0 \\ 0 & 0.5
 \end{bmatrix}
 \right)
 &\mbox{if }
 i \in \cK_{-1} \cap \cD_2. 
 \end{cases}
\label{artifical-data-generative}
\end{align}
\figurename~\ref{fig:toy_data} shows the generated instances
for $n=400$,
in which instances in the above four cases have the equal size $n/4$.
Before feeding the generated instances into algorithms,
we normalized the inputs in $[0,1]^2$.

\figurename~\ref{fig:toy_path} shows piecewise-linear paths of some of the solutions
$\alpha_i$ for $C_1 \in [0, 10]$ when $n=400$.
The left graph includes the solution paths of three representative
parameters $\alpha_i$ for $i \in \cD_1$.
All three parameters increase as $C_1$ grows from zero,
and one of the parameters (denoted by the dash-dotted line)
suddenly drops down to zero at around $C_1=7$.
Another parameter (denoted by the solid line) also sharply drops down at around $C_1=9$,
and the last one (denoted by the dashed line) remains equal to $C_1$ until $C_1$ reaches $10$.
The right graph includes the solution paths of three representative
parameters $\alpha_i$ for $i \in \cD_2$,
showing that their behavior is substantially different from that for $\cD_1$.
One of the parameters (denoted by the dash-dotted line) fluctuates
significantly, while the other two parameters
(denoted by the solid and dashed lines) are more stable and 
tend to increase as $C_1$ grows.

%--------------------------------------------------
% Toy data
%--------------------------------------------------
\begin{figure}[p]
\centering
 \includegraphics[width=3in]{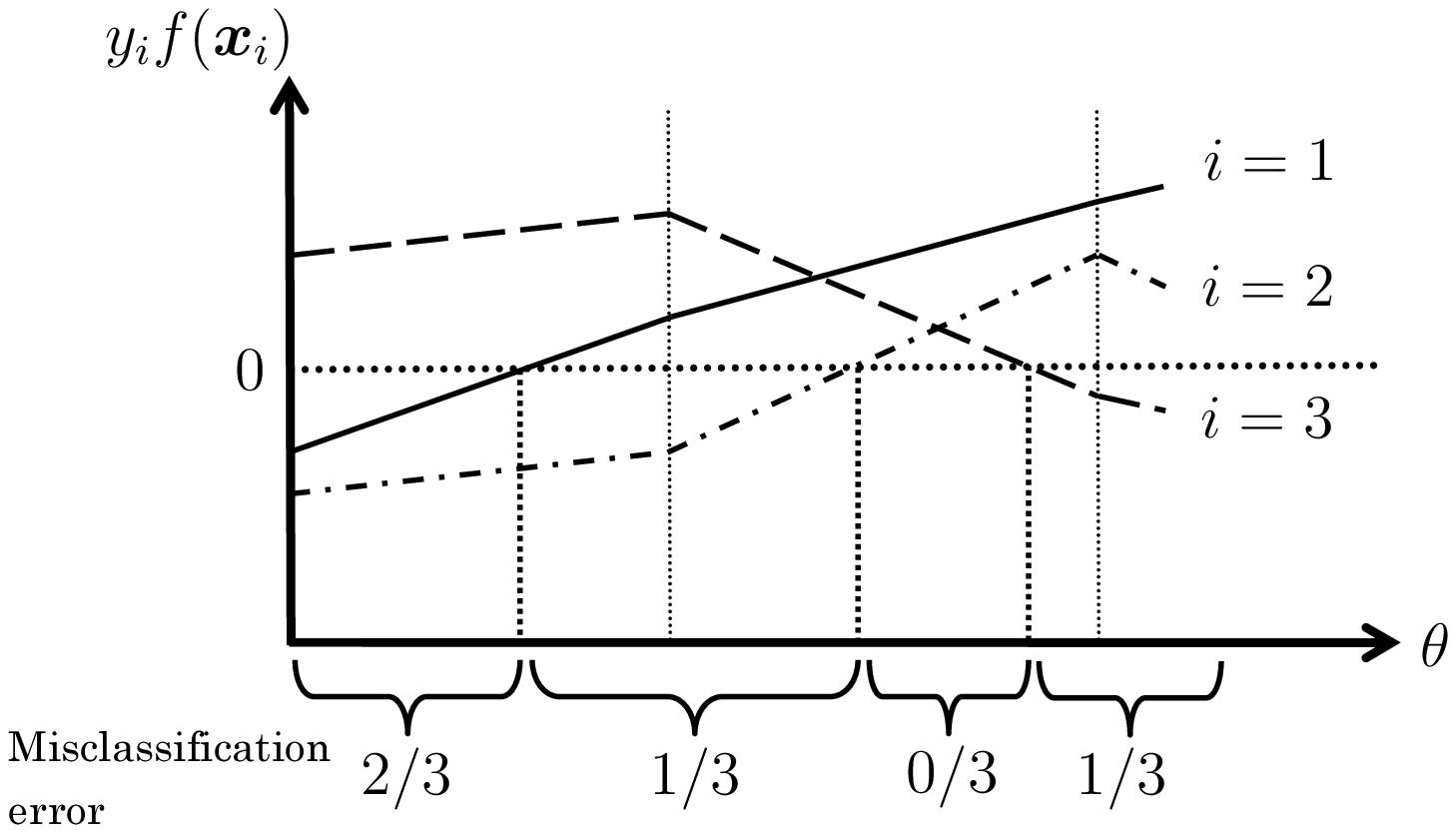}
 \caption{A schematic illustration of validation-error path. In
 this plot,
the path of misclassification error rate 
$\frac{1}{3} \sum_{i = 1}^n I(y_i f(\vx_i)) \le 0)$ 
for the 3 validation instances
$(x_1, y_1)$, $(x_2, y_2)$, and $(x_3, y_3)$
are depicted. The horizontal
axis indicates the parameter $\theta$ and the vertical axis
denotes $y_i f(\vx_i)$, $i = 1, 2, 3$. The path of the
validation error has piecewise-constant form because the 0-1 loss
changes only when $f(\vx_i) = 0$. The breakpoints of the
piecewise-constant validation-error path can be exactly detected
by exploiting the piecewise linearity of $f(\vx_i)$. }
% The schematic illustration of the validation error path.
%  In this plot, there are $3$ lines of $yf(\vx)$ each of which is a
%  piecewise linear function.
%  The fractions under the horizontal line are the classification error rate.
%  Each of them can be considered as the averaged $0$-$1$ loss at each segment.
%  We see that each error corresponds to the number of lines
%  which are the under the dashed line $yf(\vx) = 0$.
%  The classification error changes when one of $yf(\vx)$'s crosses at 
%  $yf(\vx) = 0$. 
 \label{fig:cerror_path}
% \end{figure}
% \begin{figure}[t]
%   \centering
\vspace*{3mm}
 \includegraphics[width=3.5in]{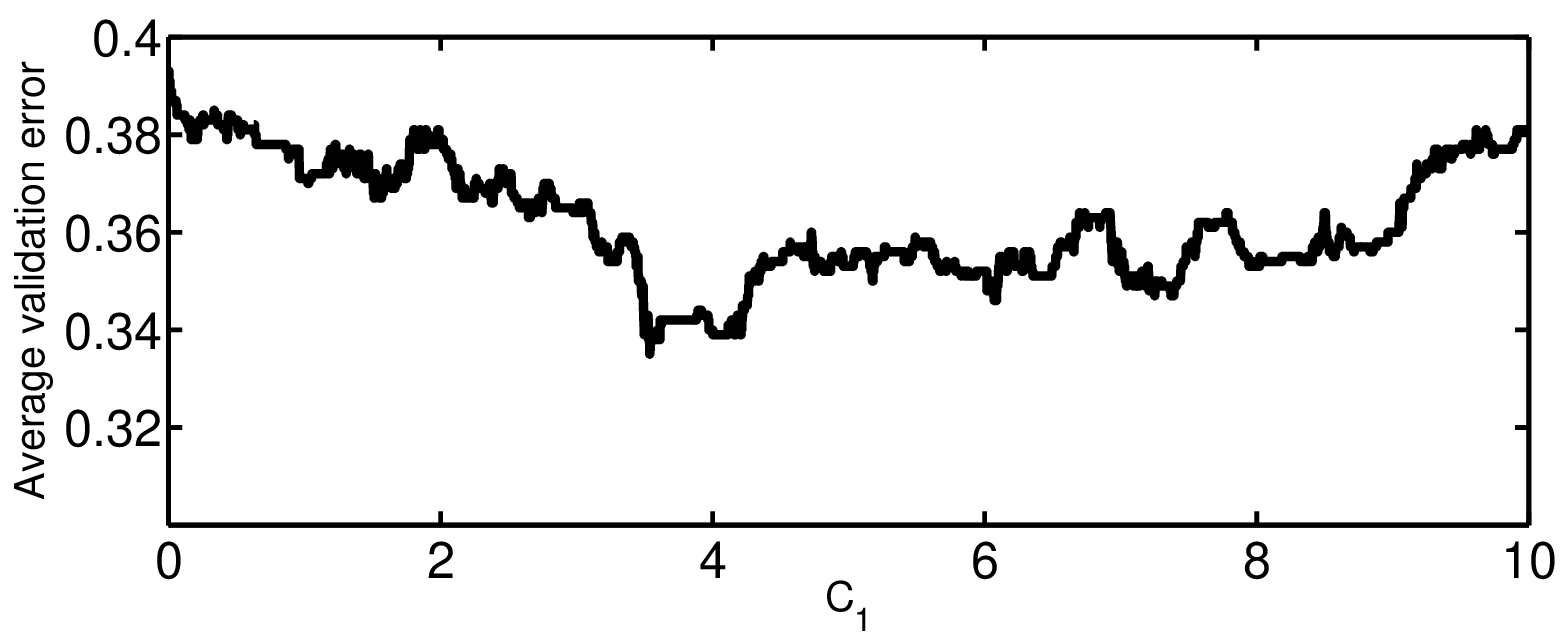}
\vspace*{-3mm}
 \caption{
 An example of the 
 validation-error path for $1000$ validation instances of the
 artificial data set.
 The number of training instances is $400$ and the Gaussian kernel 
 with $\gamma = 1$ is used.}
 \label{fig:toy_test}
\end{figure}

An important advantage of the path following algorithm is that
the path of the validation error can be traced as well (see
\figurename~\ref{fig:cerror_path}). First,
note that the path of the validation error (26) has piecewise-constant
form because the 0-1 loss changes only when the sign of
$f(\vx)$ changes. In our path-following algorithm, the path of 
$f(\vx)$
also has piecewise-linear form because $f(\vx)$ is linear in their
parameters $\valpha$ and $b$. Exploiting the piecewise linearity
of $f(\vx)$, we can exactly detect the point at which the sign of
$f(\vx)$ changes. These points correspond to the breakpoints of
the piecewise-constant validation-error path. 
\figurename~\ref{fig:cerror_path}
illustrates the relationship between the piecewise-linear path of
$f(\vx)$ and the piecewise-constant validation-error path. 
\figurename~\ref{fig:toy_test}
shows an example of piecewise-constant validation-error path when
$C_1$ is increased from 0 to 10, indicating that the
lowest validation error was achieved at around $C_1 = 4$.
% Next, we consider a scenario where,
% for model selection purposes,
% we are given additional validation data
% from the same Gaussian mixture distribution \eqref{artifical-data-generative}
% as the training data.
% \red{
% The validation error was measured by (\ref{eq:toy.loss}).
% Then, the error is a piecewise constant function of $f(\vx)$,
% because it changes only when $f(\vx)$ flips its sign.
% From (\ref{eq:y_delta_f}) and (\ref{eq:psi}), we see that $f(\vx)$
% itself changes piecewise linearly in the path.
% Using these properties we can find the point which increases or
% decreases the validation error exactly by detecting the point $f(\vx) = 0$.
% \figurename~\ref{fig:cerror_path} explains 
% how the validation error behaves
%  using the simple $0$-$1$ loss case.}
% \figurename~\ref{fig:toy_test} depicts the error 
% \eqref{eq:toy.loss} for the validation data
% when $C_1$ is increased from $0$ to $10$;
% this was exactly and efficiently computed by our path-following algorithm.
% The graph shows that the lowest validation error was achieved at around $C_1=4$.

% In each experiment, 
Finally, we investigated the computation time
when the solution path from
$C_1 = 0$ to $10$ was computed.
For comparison, we also investigated the computation time
of the SMO algorithm when the solutions at every breakpoint were computed.
We considered the following four cases: the number of training instances was
$n=400$, $800$, $1200$, and $1600$.
For each $n$, we generated $10$ data sets and the
average and standard deviation % of computation time 
over $10$ runs
are reported.
Table~\ref{tab:toy_cpu} describes the results,
showing that our path-following algorithm is faster than the SMO algorithm
in one or two orders of magnitude; the difference between the two methods
becomes more significant as the training data size $n$ grows.

The table also includes the number of events and
the average number of elements in the margin set ${\cal M}$
(see Eq.\eqref{eq:margin_set}).
This shows that the number of events increases almost linearly 
in % with respect to
the sample size $n$,
which well agrees with the empirical results reported in
\citet{Hastie04}, \citet{Gunter07}, and \citet{Wang08}.
The average number of elements in the set ${\cal M}$
increases very mildly as the sample size $n$ grows.

\begin{table}[p]
\centering
 % --------------------------------------------------
 % Toy CPU Time
 % --------------------------------------------------
 \caption{The experimental results of the artificial data set.
 The average and the standard deviation (in brackets) over $10$ runs are
 reported.
 } \label{tab:toy_cpu}
 \vspace*{2mm}
  \begin{tabular}[htb]{c|cccc}
   \hline
   $n$    & \multicolumn{2}{c}{ CPU time (sec.)}   & \#events & mean $|{\cal M}|$  \\ 
          & path  & SMO & & \\
   \hline 
400  & 0.03 (0.00) & ~0.39 (0.01) & ~326.70 (~7.17) & 3.07 (0.03) \\
800  & 0.08 (0.00) & ~2.84 (0.12) & ~635.30 (17.47) & 3.27 (0.02) \\
1200 & 0.19 (0.00) & 10.63 (0.38) & ~997.60 (26.85) & 3.38 (0.05) \\
1600 & 0.35 (0.01) & 28.11 (0.77) & 1424.00 (31.27) & 3.50 (0.02) \\
% 800  & 0.23 (0.01) & 9.03 (0.41) & 1301.80 (52.46) & 43.77 (0.47) \\
% 1200 & 0.56 (0.01) & 31.56 (1.03) & 2285.50 (61.36) & 49.81 (0.95) \\
% 1600 & 1.04 (0.03) & 77.41 (2.90) & 3213.80 (106.32) & 53.07 (0.87) \\
   \hline
  \end{tabular}
\end{table}

\subsection{Online Time-series Learning} \label{subsec:online_learning}
\begin{figure}[t]
 \centering
 \includegraphics[width=2.5in]{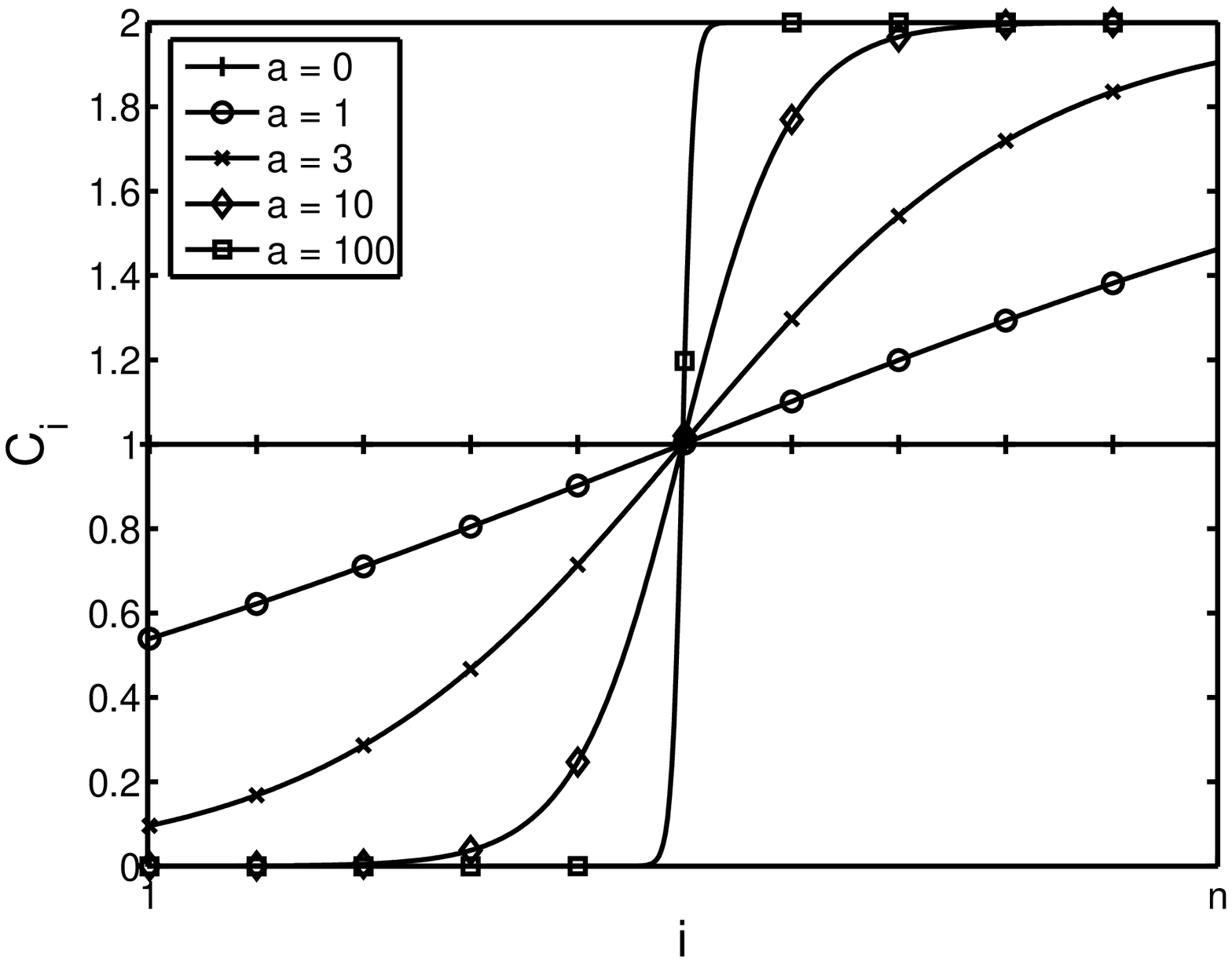}
 \caption{
 The weight functions for financial time-series forecasting~\citep{Cao03}.
 The horizontal axis is the index of training instances which is
 sorted according to
 time (the most recent instance is $i = n$).
 If we set $a = 0$, all the instances are weighted equally.}
 \label{fig:sigmoid_weight}
% \end{figure}
% \begin{figure}[!t]
%  \centering
\vspace*{5mm}
 \includegraphics[width=4in]{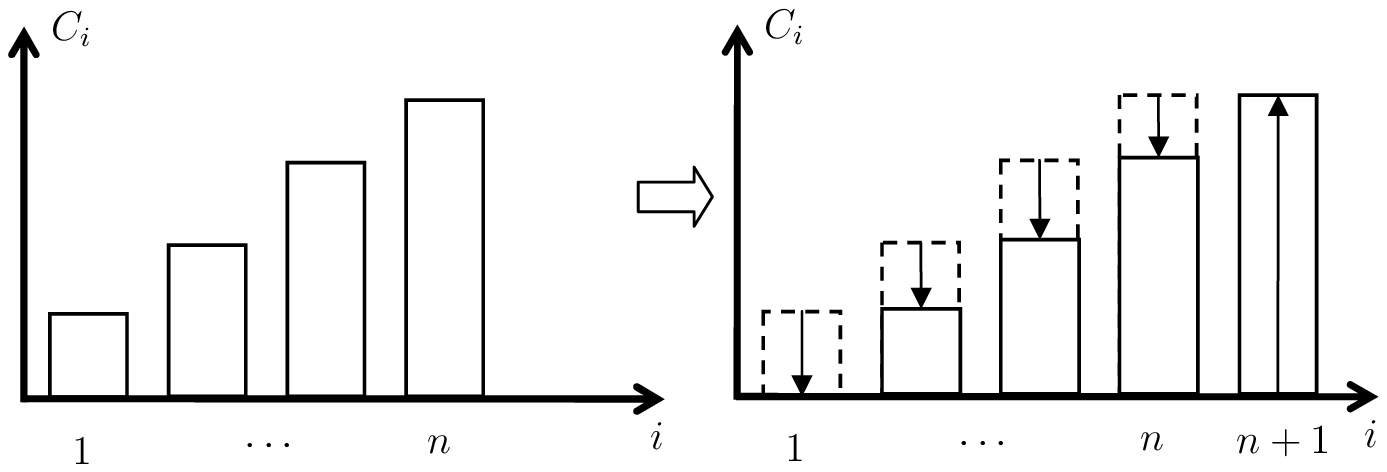}
 \caption{
 A schematic illustration of the change of weights in time-series learning.
 The left plot shows 
 the fact that larger weights are assigned to more recent instances. 
 The right plot describes a situation where we receive a new instance
 ($i = n + 1$). 
 In this situation, the oldest instance ($i = 1$)
 is deleted by setting its weight to zero,
 the weight of the new instance is set to be the largest, 
 and the weights of the rest of the instances are decreased accordingly.
 %that recent data points are imposed larger weights.
 %In the right plot, since a new data point $i = n + 1$ is arrived, we
 %update the weights of all 
 %data points.
 }
 \label{fig:online_weights}
\end{figure}

In online time-series learning, 
larger 
(resp.~smaller)
weights should be assigned to 
newer 
(resp.~older)
instances.
%
%\citep{Cao03} proposed to give more weights on the more recent training
%data points to incorporate the nonstationarity of the financial time
%series data. 
%
For example, 
in \citet{Cao03}, 
the following weight function is used:
\begin{eqnarray}
 C_i = C_0 \frac{2}{1 + \exp (a - 2a \times \frac{i}{n})},
  \label{eq:weight_function}
\end{eqnarray}
where $C_0$ and $a$ are hyper-parameters and 
the instances are assumed to be sorted along the time axis
(the most recent instance is $i = n$).
\figurename~\ref{fig:sigmoid_weight} shows the profile of the weight function
(\ref{eq:weight_function}) when $C_0 = 1$.
In online learning, we need to update parameters
when new observations arrive, 
and all the weights must be updated accordingly
(see \figurename~\ref{fig:online_weights}).

% --------------------------------------------------
% Sigmoid Weight
% --------------------------------------------------

We investigated the computational cost of updating parameters
when several new observations arrive.
The experimental data are obtained from the \emph{NASDAQ composite index}
between January 2, 2001 and December 31, 2009.
As \citet{Cao03} and \citet{Chen06}, 
we transformed the original closing prices using the Relative
Difference in Percentage (RDP) of the price and 
the exponential moving average (EMA).

Extracted features are listed in \tablename~\ref{tab:features_for_financial}
\citep[see][for more details]{Cao03}.
Our task is to predict the sign of RDP+5 using 
EMA15 and four lagged-RDP values 
(RDP-5, RDP-10, RDP-15, and RDP-20).
RDP values which exceed $\pm 2$ standard deviations are replaced
with the closest marginal values.
We have an initial set of training instances with size $n = 2515$.
The inputs were normalized in $[0,1]^p$, where $p$ is the dimensionality
of the input $\vx$.
We used the Gaussian kernel \eqref{Gaussian-kernel} with 
$\gamma \in \{10, 1, 0.1\}$,
and the weight parameter $a$ in (\ref{eq:weight_function}) was set to $3$.
We first trained WSVM using the initial set of instances.
Then we added $5$ instances to the previously trained WSVM and removed
the oldest $5$ instances by decreasing their weights to $0$.
This does not change the size of the training data set,
but the entire weights need to be updated
as illustrated in \figurename~\ref{fig:online_weights}.
We iterated this process $5$ times and compared the total computational costs
of the path-following algorithm and the SMO algorithm.
For fair comparison, we cleared the cache of kernel values at each update
before running the algorithms.
\begin{table}[t]
 \centering
  % --------------------------------------------------
  % Table: Features for Financial Forecasting
  % --------------------------------------------------
  \caption{Features for financial forecasting 
  ($p(i)$ is the closing price of the $i$th day and
  ${\rm EMA}_k(i)$ is the
  $k$-day exponential moving average of the $i$th day.}
  \label{tab:features_for_financial}
  \vspace*{2mm}
  \begin{tabular}{l|l}
   \hline
   Feature & Formula \\
   \hline 
   EMA15 & $p(i) - {\rm EMA}_{15}(i)$ \\
   RDP-5 & $(p(i) - p(i - 5))/p(i - 5) * 100$ \\
   RDP-10 & $(p(i) - p(i - 10))/p(i - 10) * 100$ \\
   RDP-15 & $(p(i) - p(i - 15))/p(i - 15) * 100$ \\
   RDP-20 & $(p(i) - p(i - 20))/p(i - 20) * 100$ \\
   RDP+5 & $(\overline{p(i + 5)} - \overline{p(i)})/\overline{p(i)} * 100$ \\
   & $\overline{p(i)} = {\rm EMA}_3(i)$ \\
   \hline
  \end{tabular}
\end{table}
%
% --------------------------------------------------
% CPU time of Online Learning
% --------------------------------------------------
\begin{figure}[t]
 \centering
 \subfigure[$\gamma = 10$]{
 \includegraphics[width=0.3\linewidth]{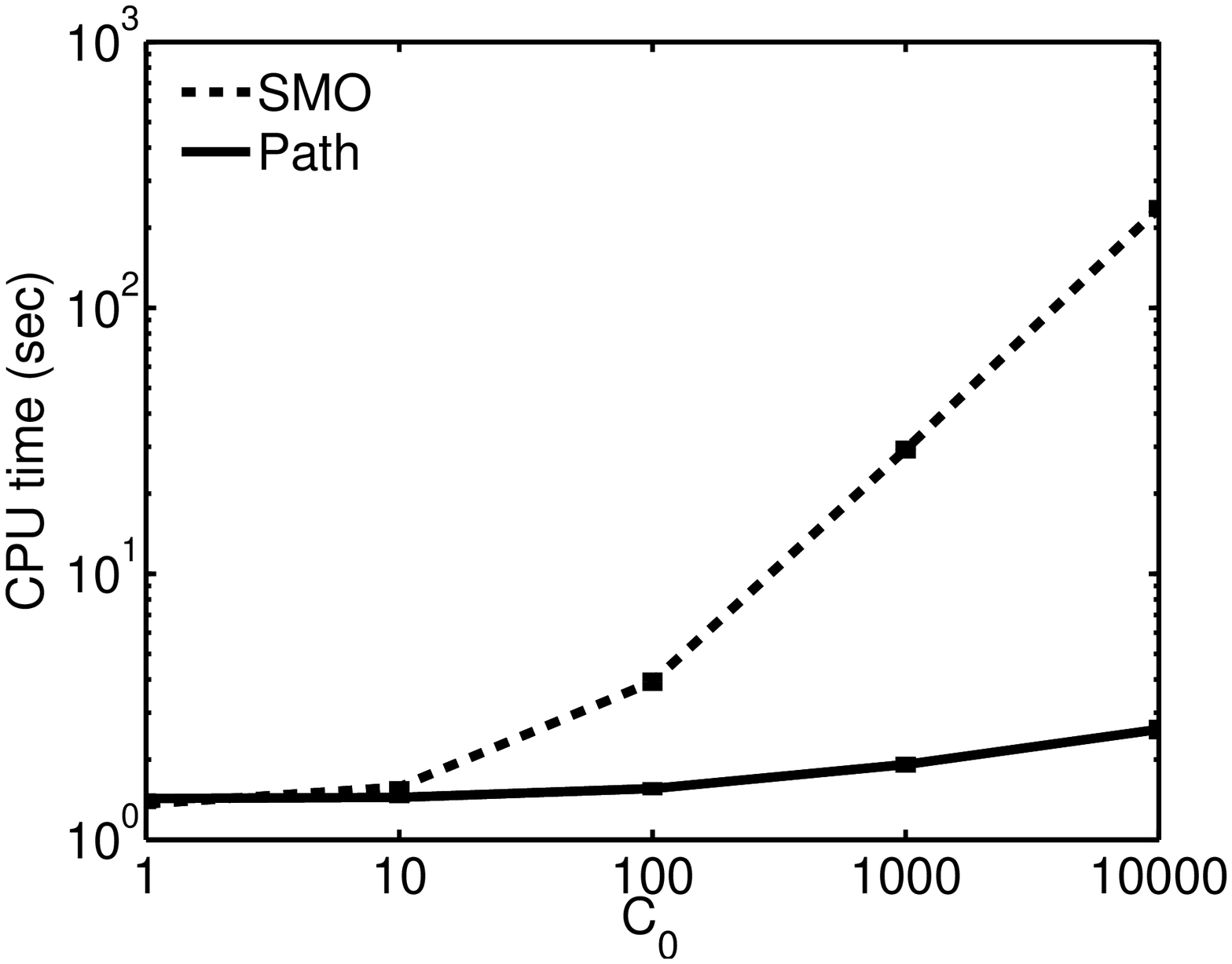}}
 \subfigure[$\gamma = 1$]{
 \includegraphics[width=0.3\linewidth]{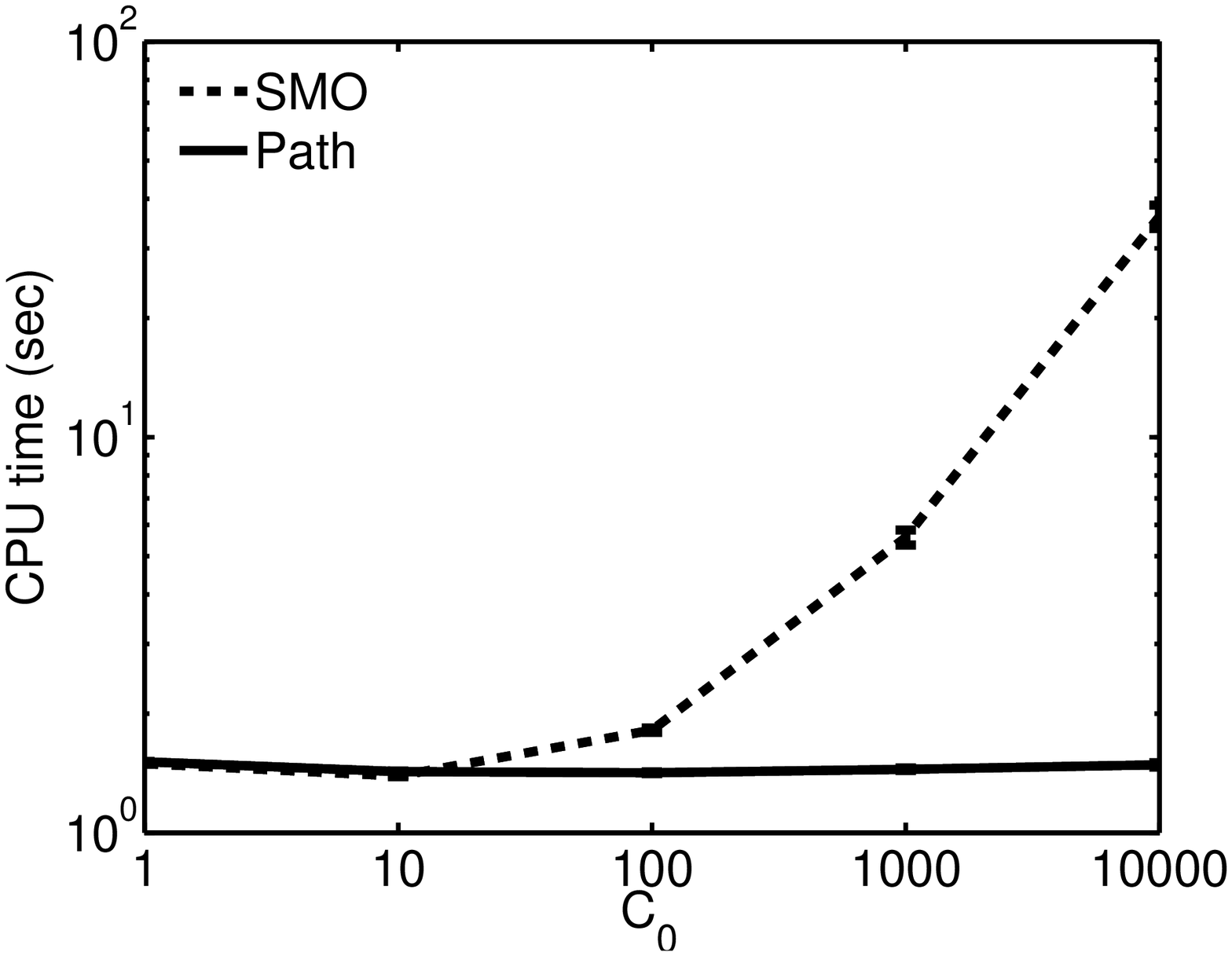}}
 \subfigure[$\gamma = 0.1$]{
 \includegraphics[width=0.3\linewidth]{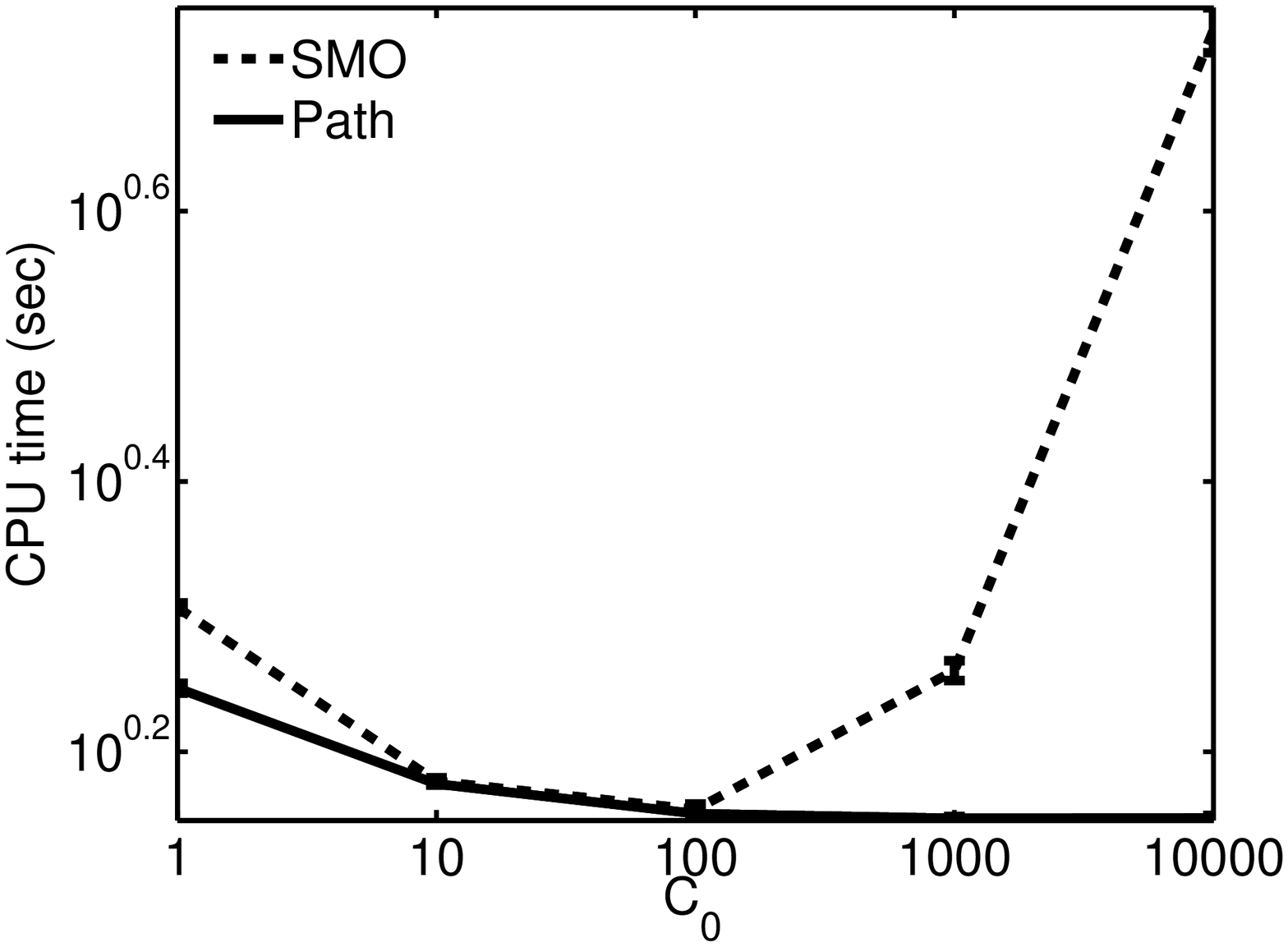}}
 \caption{CPU time comparison for online time-series learning using
NASDAQ composite index.}
 \label{fig:online_cpu}
\end{figure}

\figurename~\ref{fig:online_cpu} shows the average CPU time over $10$ runs
for $C_0 \in \{1, 10, 10^2,10^3, 10^4\}$,
showing that the path-following algorithm
is much faster than the SMO algorithm especially for large $C_0$.

%==================================================
% Covariate Shift Adaptation
%==================================================

\subsection{Model Selection in Covariate Shift Adaptation}

Covariate shift is a situation in supervised learning
where the input distributions change between the training and test
phases but the conditional distribution of outputs given inputs
remains unchanged \citep{JSPI:Shimodaira:2000}.
Under covariate shift, standard SVM and SVR are biased,
and the bias caused by covariate shift can be asymptotically canceled
by weighting the loss function according to the \emph{importance}
(i.e., the ratio of training and test input densities).
% \citepp{JSPI:Shimodaira:2000,ICML:Zadrozny:2004,StatDeci:Sugiyama+Mueller:2005,JMLR:Sugiyama+etal:2007,NIPS2006_915}.

Here, we apply importance-weighted SVMs to \emph{brain-computer interfaces} (BCIs)
\citep{book:Dornhege+etal:2007}.
A BCI is a system which allows for a direct communication from man to machine
via brain signals.
Strong non-stationarity effects have been often observed
in brain signals between training and test sessions, 
which could be modeled as covariate shift \citep{JMLR:Sugiyama+etal:2007}.
We used the BCI datasets provided by the Berlin BCI group \citep{Burde06},
containing 24 binary classification tasks.
The input features are 4-dimensional preprocessed 
\emph{electroencephalogram} (EEG) signals,
and the output labels correspond to the `left' and `right' commands.
The size of training datasets is around $500$ to $1000$,
and the size of test datasets is around $200$ to $300$.

Although the importance-weighted SVM tends to have lower bias,
it in turns has larger estimation variance than the ordinary SVM \citep{JSPI:Shimodaira:2000}.
Thus, in practice, it is desirable to slightly `flatten' the instance
weights so that the trade-off between bias and variance is optimally controlled.
Here, we changed the instance weights from 
the uniform values to the importance values using the proposed path-following algorithm,
i.e., the instance weights were changed from $C_i^{\rm (old)} = C_0$ to
$ C_i^{\rm (new)} = C_0 \frac{p_{test}(\vx_i)}{p_{train}(\vx_i)}$,
$i = 1, \ldots, n$.
The importance values $\frac{p_{test}(\vx_i)}{p_{train}(\vx_i)}$ were estimated by
the method proposed in \citet{JMLR:Kanamori+etal:2009},
which directly estimates the density ratio
without going through density estimation of $p_{test}(\vx)$ and $p_{train}(\vx)$.

For comparison, we ran the SMO algorithm at 
(i) each breakpoint of the solution path, 
and (ii) $100$ weight vectors taken uniformly in $[C_i^{\rm (old)}, C_i^{\rm (new)}]$.
We used the Gaussian kernel and the inputs were normalized in $[0, 1]^p$,
where $p$ is the dimensionality of $\vx$.

\begin{figure}[!t]
 \centering
 \subfigure[$\gamma = 10$]
 {\includegraphics[width=0.3\linewidth]{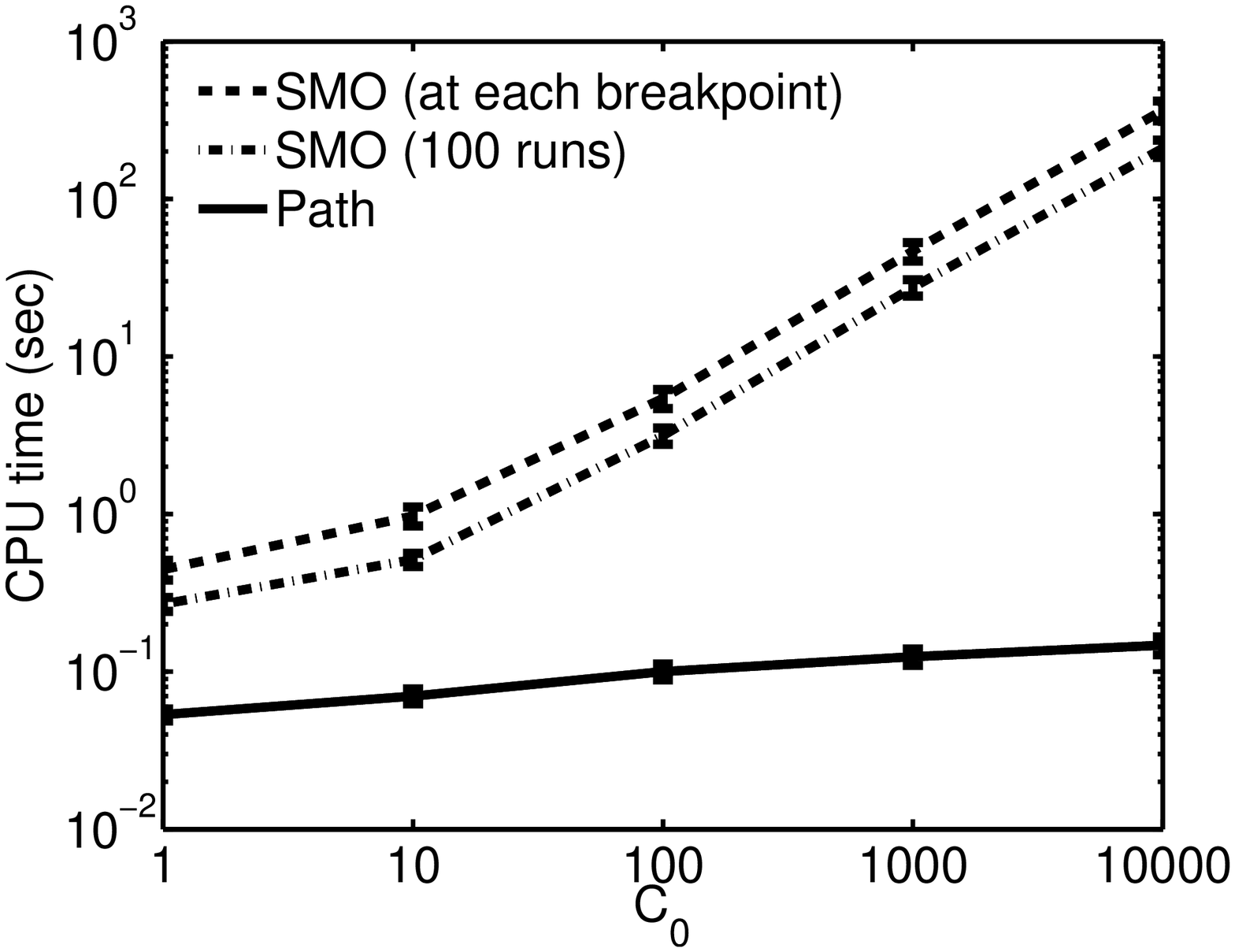}}
 \subfigure[$\gamma = 1$]
 {\includegraphics[width=0.3\linewidth]{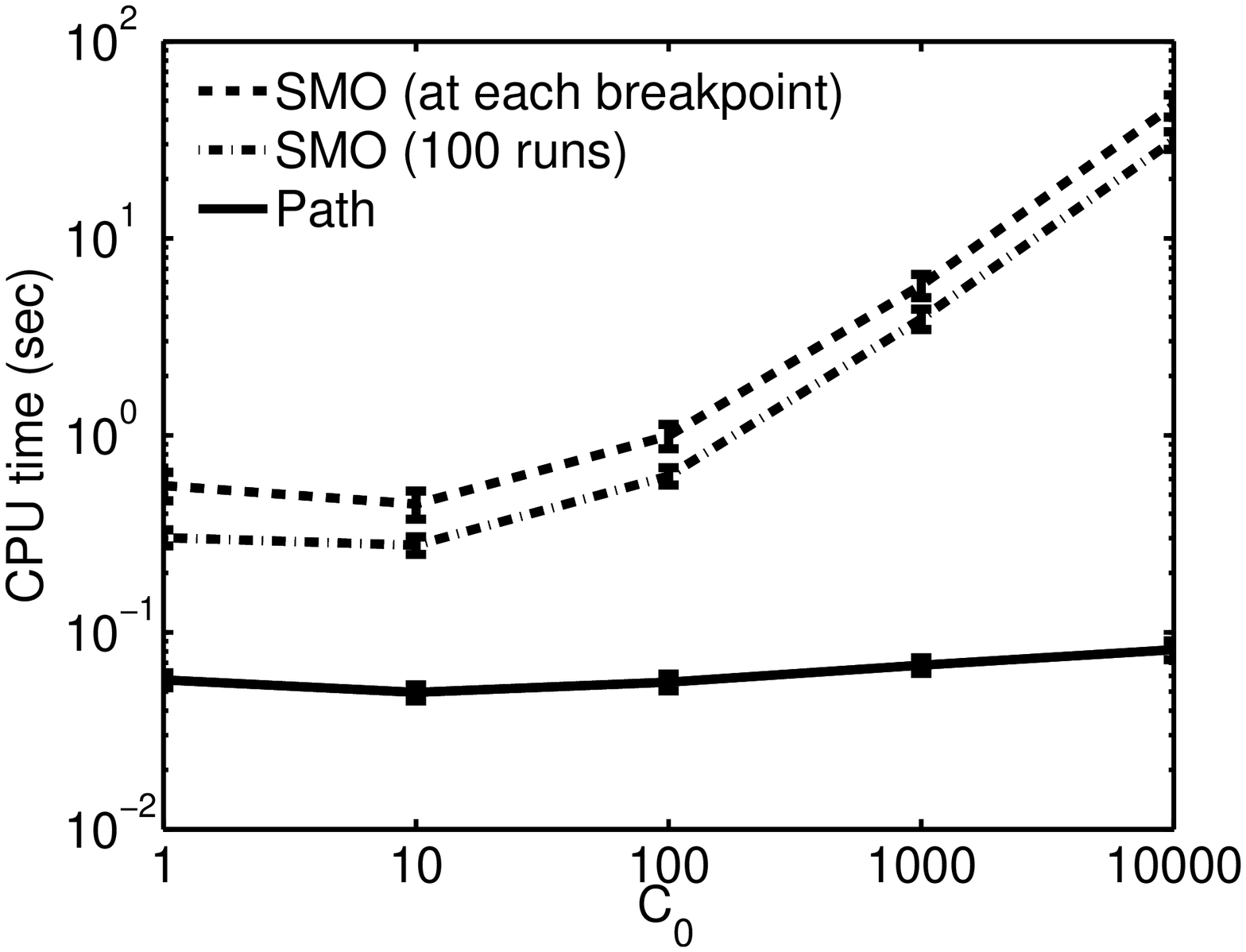}}
 \subfigure[$\gamma = 0.1$]
 {\includegraphics[width=0.3\linewidth]{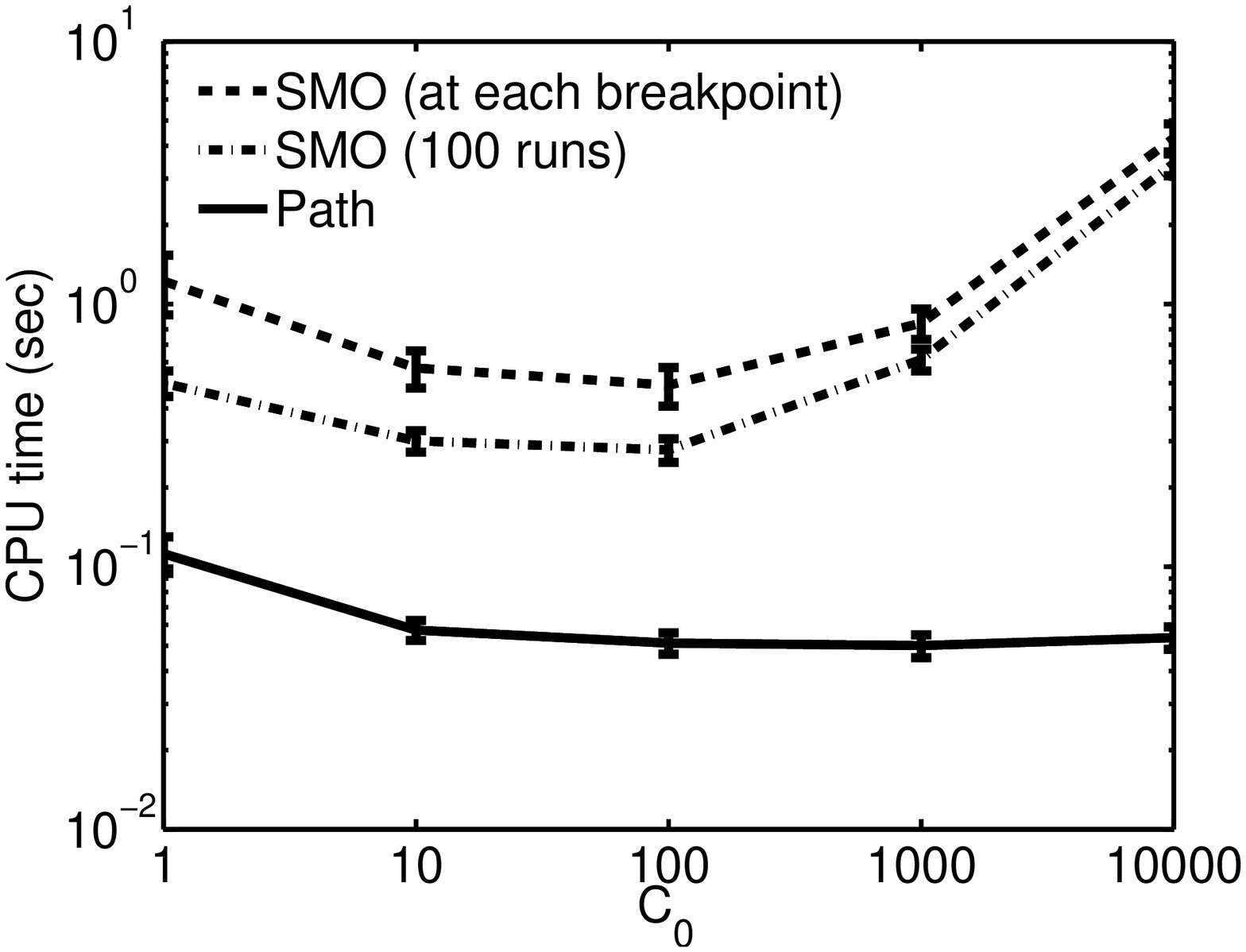}}
 % \subfigure[$\gamma = 0.01$]
 % {\includegraphics[scale=0.35]{./figs/covshift_cpu_g0.01.eps}}
 \caption{CPU time comparison for covariate shift adaptation
   using BCI data.}
 \label{fig:cov_shift_time}
\end{figure}

\figurename~\ref{fig:cov_shift_time} shows the average CPU time and its standard deviation.
We examined several settings of hyper-parameters 
$\gamma \in \{10, 1, \ldots, 10^{-2}\}$ and
$C_0 \in \{1, 10, 10^2,\ldots, 10^4\}$.
The horizontal axis of each plot represents $C_0$.
The graphs show that our path-following algorithm is
faster than the SMO algorithm in all cases.
While the SMO algorithm tended to take longer time for large $C_0$,
the CPU time of the path-following algorithm did not increase with respect to $C_0$.

%%% Local Variables: 
%%% mode: latex
%%% TeX-master: "paper"
%%% End: 

%\clearpage

%==================================================
% Sec::Beyond Classification
%==================================================
\section{Beyond Classification}
\label{sec:wsvm_extension}
So far, we focused on classification scenarios.
Here we show that the proposed path-following algorithm
can be extended to various scenarios
including regression, ranking, and transduction.
% The experimental setup described at the beginning of Section
% \ref{sec:applications} was also employed in this section.

% --------------------------------------------------
\subsection{Regression}
The \emph{support vector regression} (SVR) is a variant of
SVM for regression problems \citep{Vapnik96,Mattera99,Muller99}.

\subsubsection{Formulation}
The primal optimization problem for the weighted SVR (WSVR) is defined by
\begin{eqnarray*}
 \min_{\mb{w}, b, \{\xi_i, \xi_i^*\}_{i=1}^n} && \frac{1}{2} \| \mb{w} \|_2^2 
 + \sum_{i=1}^n C_i (\xi_i + \xi_i^*), \\
 {\rm s.t.}  
  && y_i - f(\vx_i) \leq \varepsilon + \xi_i, \\
  && f(\vx_i) - y_i \leq \varepsilon + \xi_i^*, \\
  && \xi_i, \xi_i^* \geq 0, \ i = 1, \ldots, n,
\end{eqnarray*}
where $\epsilon > 0$ is an insensitive-zone thickness.
Note that, as in the classification case,
WSVR is reduced to the original SVR when $C_i=C$ for $i=1,\ldots,n$.
Thus, WSVR includes SVR as a special case.

The corresponding dual problem is given by
\begin{eqnarray*}
\max_{\{ \alpha_i \}_{i=1}^n} &&
- \frac{1}{2} \sum^n_{i=1} \sum^n_{j=1} \alpha_i \alpha_j K(\vx_i,\vx_j)
 - \varepsilon \sum^n_{i=1} | \alpha_i |  
 + \sum^{n}_{i=1}y_i \alpha_i \label{eq:SVRDual} \\
{\rm s.t.} &&
 \sum^n_{i=1} \alpha_i = 0, \ -C_i \leq \alpha_i  \leq C_i ,\  i = 1, \ldots, n.
\end{eqnarray*}
The final solution, i.e., the regression function $f:{\cal X} \rightarrow {\mathbb R}$,
is in the following form:
\begin{eqnarray*}
 f(\vx) = \sum_{i=1}^{n} \alpha_i K(\vx, \vx_i) + b.
\end{eqnarray*}

\begin{figure}[t]
 \centering
 \includegraphics[width=3in]{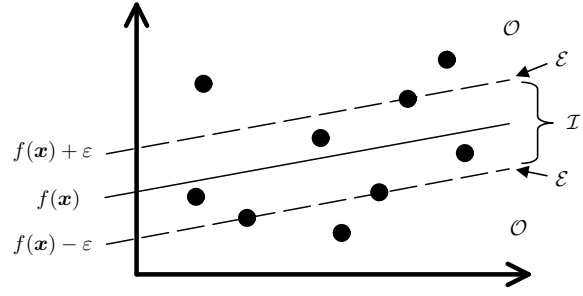}
 \caption{Partitioning of data points in SVR.}
 \label{fig:svr_sv}
\end{figure}

The KKT conditions for the above dual problem are given as
\begin{subequations}
\begin{eqnarray}
  | y_i - f(\vx_i) | \leq \varepsilon, & {\rm if} &  \alpha_i = 0, 
   \label{eq:y_minus_f_leq_eps} \\
  | y_i - f(\vx_i) | = \varepsilon, & {\rm if} &  0 < |\alpha_i| < C_i, 
   \label{eq:y_minus_f_eq_eps} \\
  | y_i - f(\vx_i) | \geq \varepsilon, & {\rm if} &  |\alpha_i| = C_i, 
   \label{eq:y_minus_f_geq_eps} \\
  \sum_{i = 1}^n \alpha_i = 0.
   \label{eq:eq_const_svr}
\end{eqnarray}
\label{eq:kkt_svr}
\end{subequations}
Then the training instances can be partitioned into
the following three index sets (see \figurename~\ref{fig:svr_sv}):
\begin{align*}
 {\cal O} &= 
  \{ i : | y_i - f(\vx_i) | \geq \varepsilon, |\alpha_i| = C_i \}, \\
 {\cal E} &= 
  \{ i : | y_i - f(\vx_i) | = \varepsilon, 0 < |\alpha_i| < C_i \}, \\
 {\cal I} &= 
  \{ i : | y_i - f(\vx_i) | \leq \varepsilon, \alpha_i = 0 \}. 
\end{align*}

%\red
Let 
\begin{eqnarray*}
 \vK^{\cal E} =
 \begin{bmatrix}
   0 & \mb{1}^\top \\
   \mb{1} & \vK_{\cal E} 
 \end{bmatrix}
  \mbox{\ \ and \ \ } 
 \mb{s} = 
 \begin{bmatrix}
  {\rm sign}(y_1 - f(\vx_1))\\ \vdots\\ {\rm sign}(y_n - f(\vx_n))
 \end{bmatrix}.
\end{eqnarray*}
Then, from (\ref{eq:kkt_svr}), we obtain
\begin{align*}
   \begin{bmatrix}
   b \\
   \valpha_{\cal E}
  \end{bmatrix}
  &=
 - (\vK^{\cal E})^{-1}
  \begin{bmatrix}
   \mb{1}_{\cal O}^\top \\
   \vK_{{\cal E},{\cal O}} 
  \end{bmatrix} 
 \mb{c}_{\cal O}
 + 
 (\vK^{\cal E})^{-1}
  \begin{bmatrix}
   0 \\
   \vy_{\cal E} - \varepsilon \mb{s}_{\cal E}
  \end{bmatrix} , %  \label{eq:affine_solution_svr} 
\\
 \valpha_{\cal O} &= {\rm diag}(\mb{s}_{\cal O}) \mb{c}_{\cal O},\\
 \valpha_{\cal O} &= \mb{0}.
\end{align*}
where ${\rm diag}(\mb{s}_\cO)$ indicates the diagonal matrix with
its diagonal part $\mb{s}_\cO$.
These functions are \emph{affine} w.r.t.~$\mb{c}$,
so we can easily detect an event point by monitoring the inequalities in 
(\ref{eq:kkt_svr}).
We can follow the solution path of SVR by
using essentially the same technique as SVM classification 
(and thus the details are omitted).

%==================================================
% Experiments on Regression
%==================================================
\subsubsection{Experiments on Regression}

As an application of WSVR,
we consider a heteroscedastic regression problem,
where output noise variance depends on input points.
In heteroscedastic data modeling, 
larger (resp.~smaller) weights are usually assigned to instances with
smaller (resp.~larger) variances.
Since the point-wise variances are often unknown in practice,
they should also be estimated from data.
A standard approach is to 
alternately estimate the weight vector $\mb{c}$ based on the current WSVR solution
and update the WSVR solutions based on the new weight vector $\mb{c}$
\citep{kersting:icml07}.

We set the weights as
%
% --------------------------------------------------
% Heteroscedastic Weight
% --------------------------------------------------
\begin{eqnarray}
 C_i = C_0 
  % \frac{\sqrt{\frac{1}{n}\sum_{j=1}^n (y_j - f(\vx_j))^2}}
  \frac{\widehat{\sigma}}{| e_i |},
  % {|y_i - f(\vx_i)|},
  \label{eq:hetero_weights}
\end{eqnarray}
where 
$e_i = y_i - \widehat{f}(\vx_i)$ 
is the residual of the instance 
$(\bm{x}_i, y_i)$ 
from the current fit 
$\widehat{f}(\bm{x}_i)$, 
and 
$\widehat{\sigma}$ 
is an estimate of the common standard deviation of the noise computed as 
$\widehat{\sigma} = \sqrt{\frac{1}{n}\sum_{i = 1}^n e_i^2}$.
%
% --------------------------------------------------
% Noise Estimator
% --------------------------------------------------
%\begin{eqnarray*}
% \widehat{\sigma} = \sqrt{\frac{1}{n}\sum_{i = 1}^n e_i^2}.
%\end{eqnarray*}
%
%In this paper, 
We employed the following procedure for 
the heteroscedastic data modeling:
\begin{description}
 \item[Step1:] Training WSVR with uniform weights
	    (i.e., $C_i = C_0, i = 1, \ldots, n$.).
 \item[Step2:] Update weights %$C_i$ 
	    by 
	    %$C_i = C_0 \frac{\widehat{\sigma}}{| e_i |}$
	    (\ref{eq:hetero_weights})
	    and update the solution of WSVR accordingly.
	    Repeat this step until
	    $\frac{1}{n} \sum_{i = 1}^n
	    | (e^{\rm (old)}_i - e_i ) /  e^{\rm (old)}_i |
	    \leq 10^{-3}$ holds,
	    where $e^{\rm (old)}$ 
	    is the previous training error.
\end{description}
%
% --------------------------------------------------
% CPU time of heteroscedastic 
% --------------------------------------------------
\begin{figure}[t]
 \centering
 \subfigure[$\gamma = 10$]{
 \includegraphics[width=0.3\linewidth]{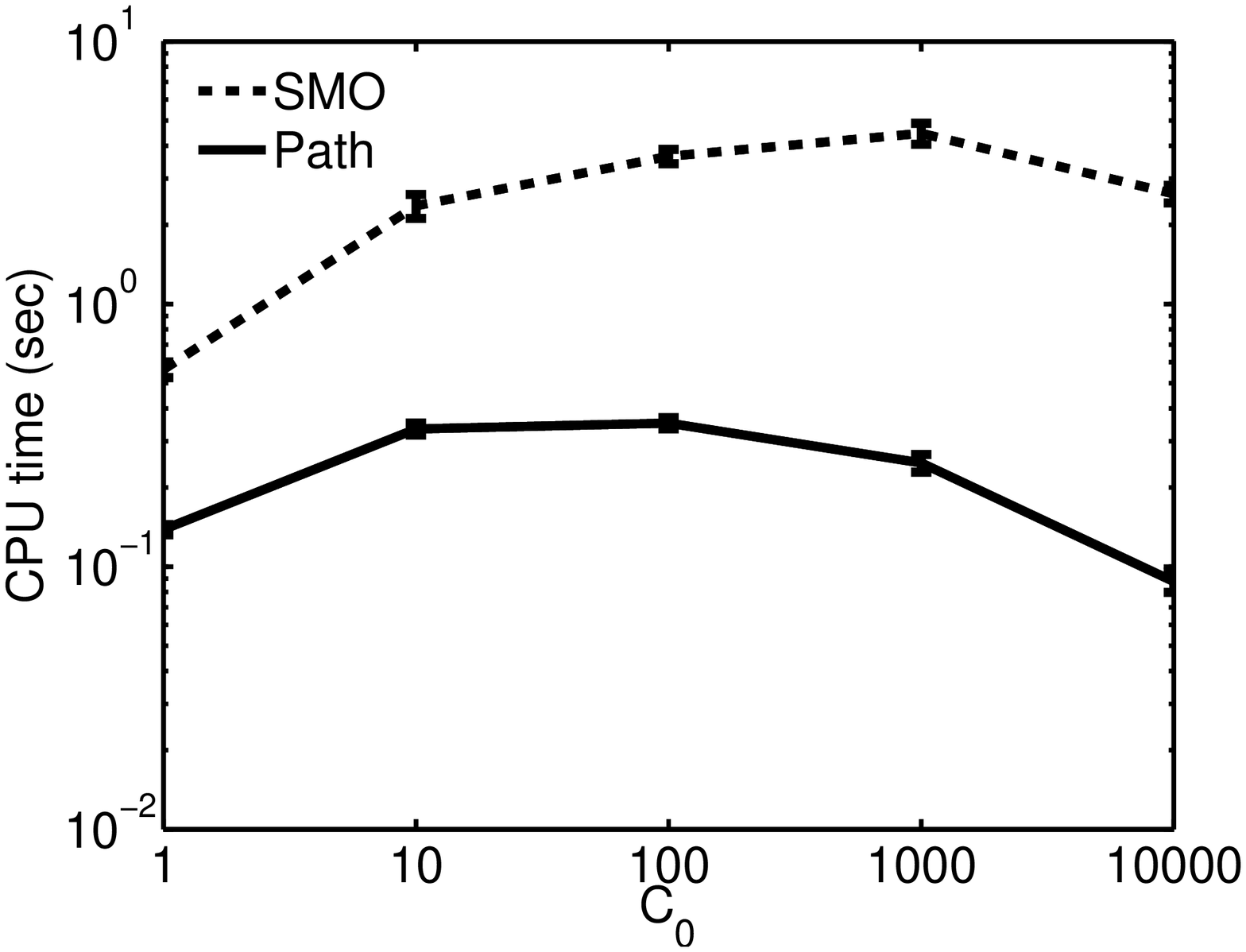}}
 \subfigure[$\gamma = 1$]{
 \includegraphics[width=0.3\linewidth]{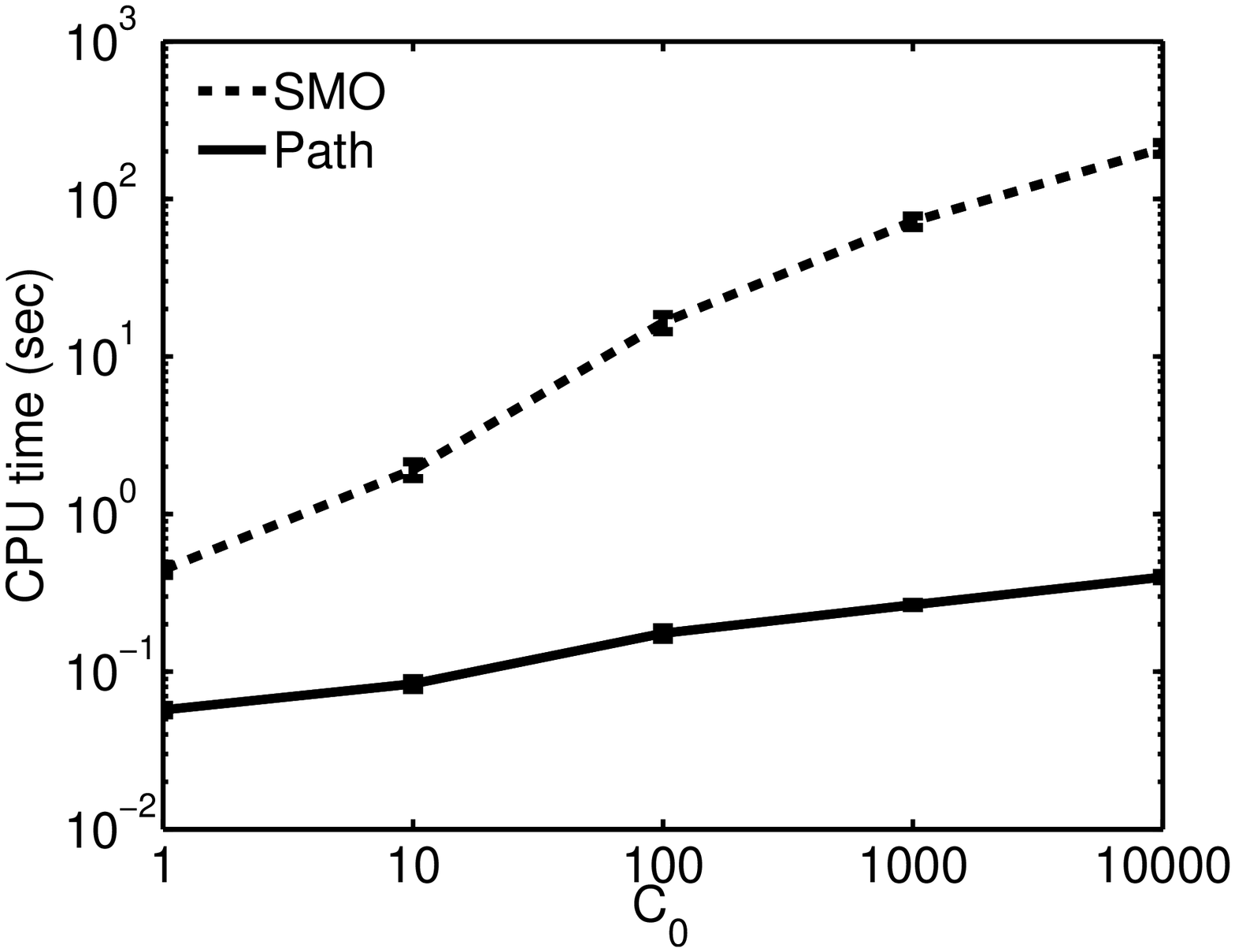}}
 \subfigure[$\gamma = 0.1$]{
 \includegraphics[width=0.3\linewidth]{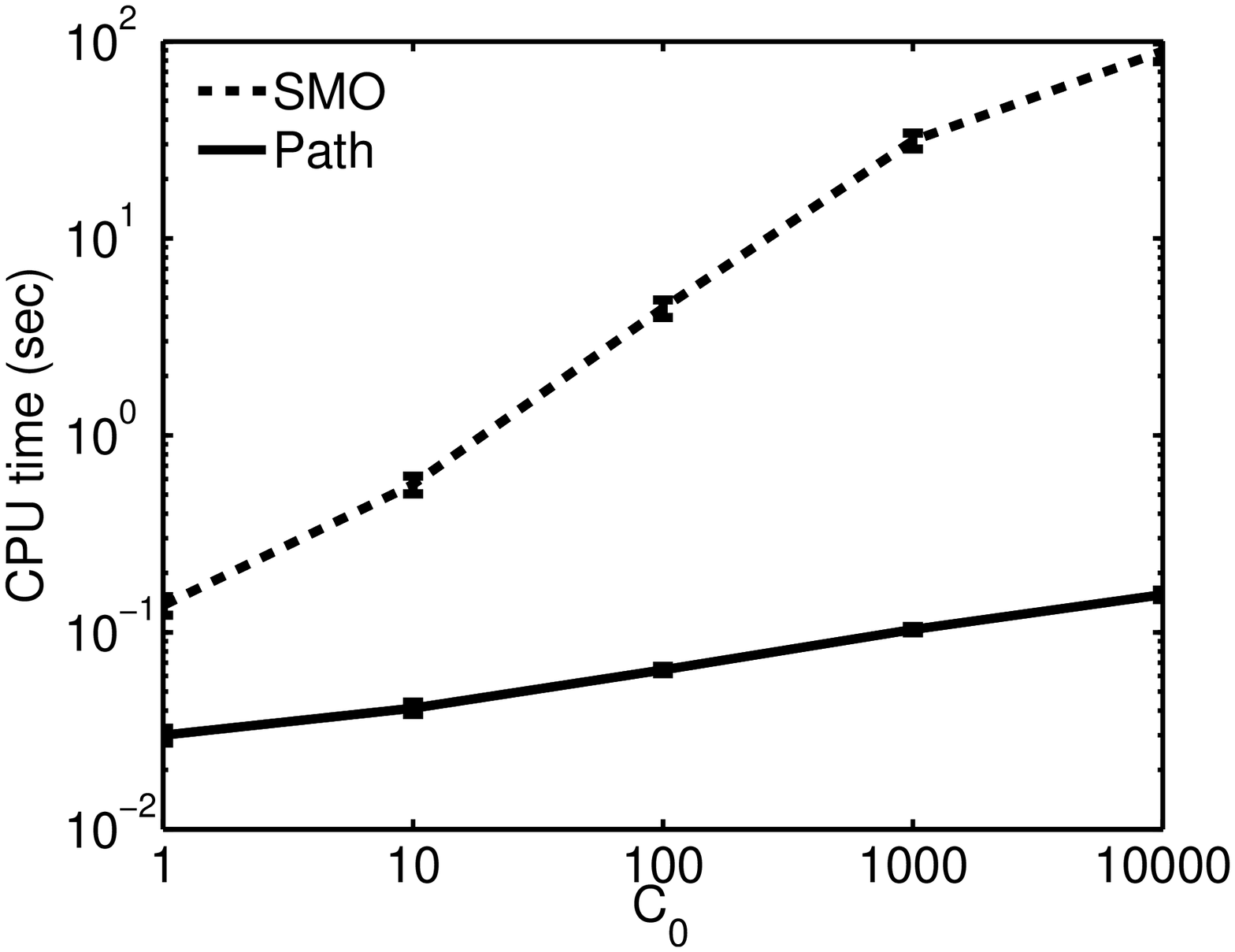}}
 \caption{CPU time comparison for heteroscedastic modeling using Boston housing
 data.}
 % \caption{CPU time comparison for heteroscedastic modeling using housing
 % data set.}
 \label{fig:hetero_cpu}
% \end{figure}
% \begin{figure}[tb]
%  %
\vspace*{5mm}
 \subfigure[$\gamma = 10$]{
 \includegraphics[width=0.3\linewidth]{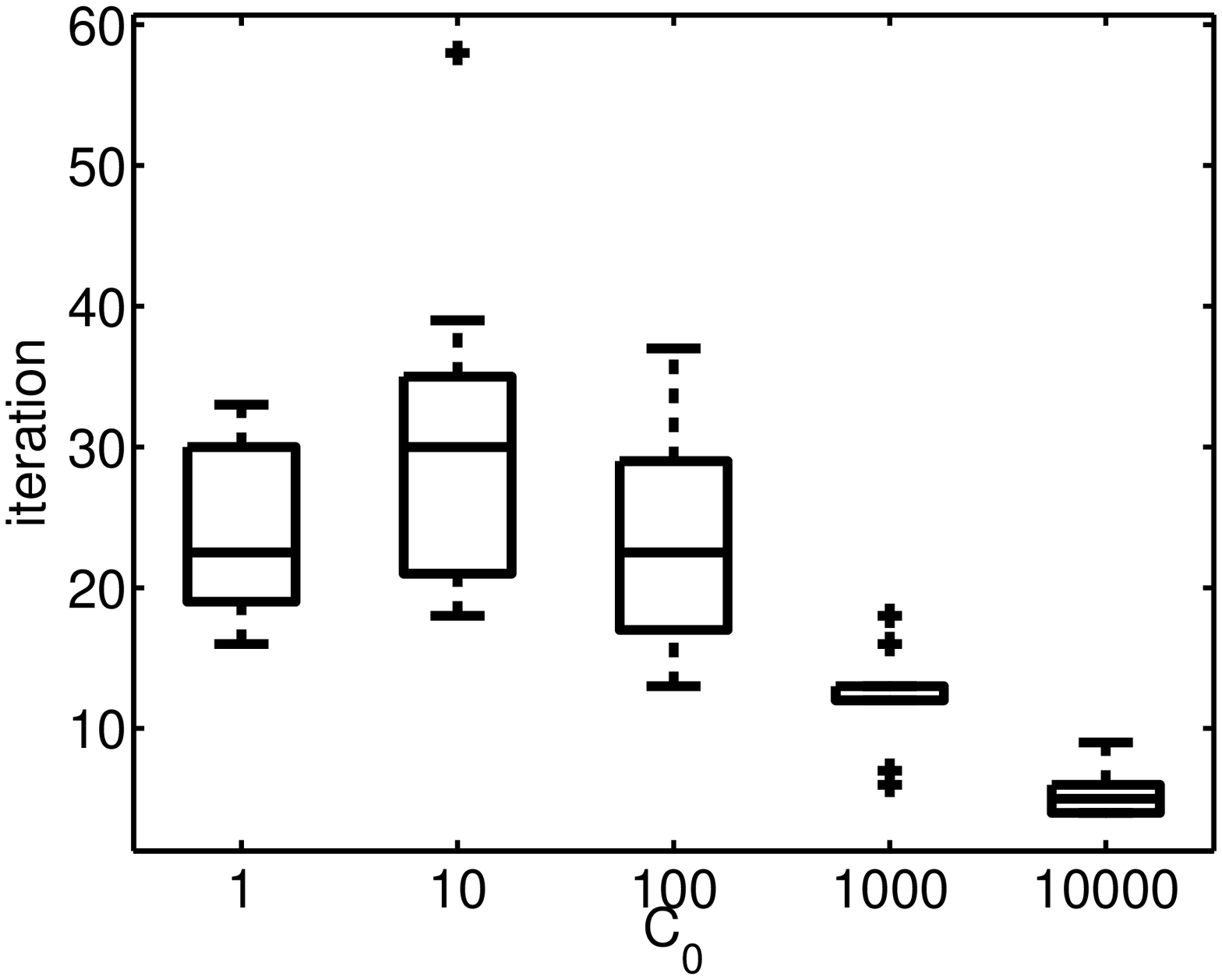}}
 \subfigure[$\gamma = 1$]{
 \includegraphics[width=0.3\linewidth]{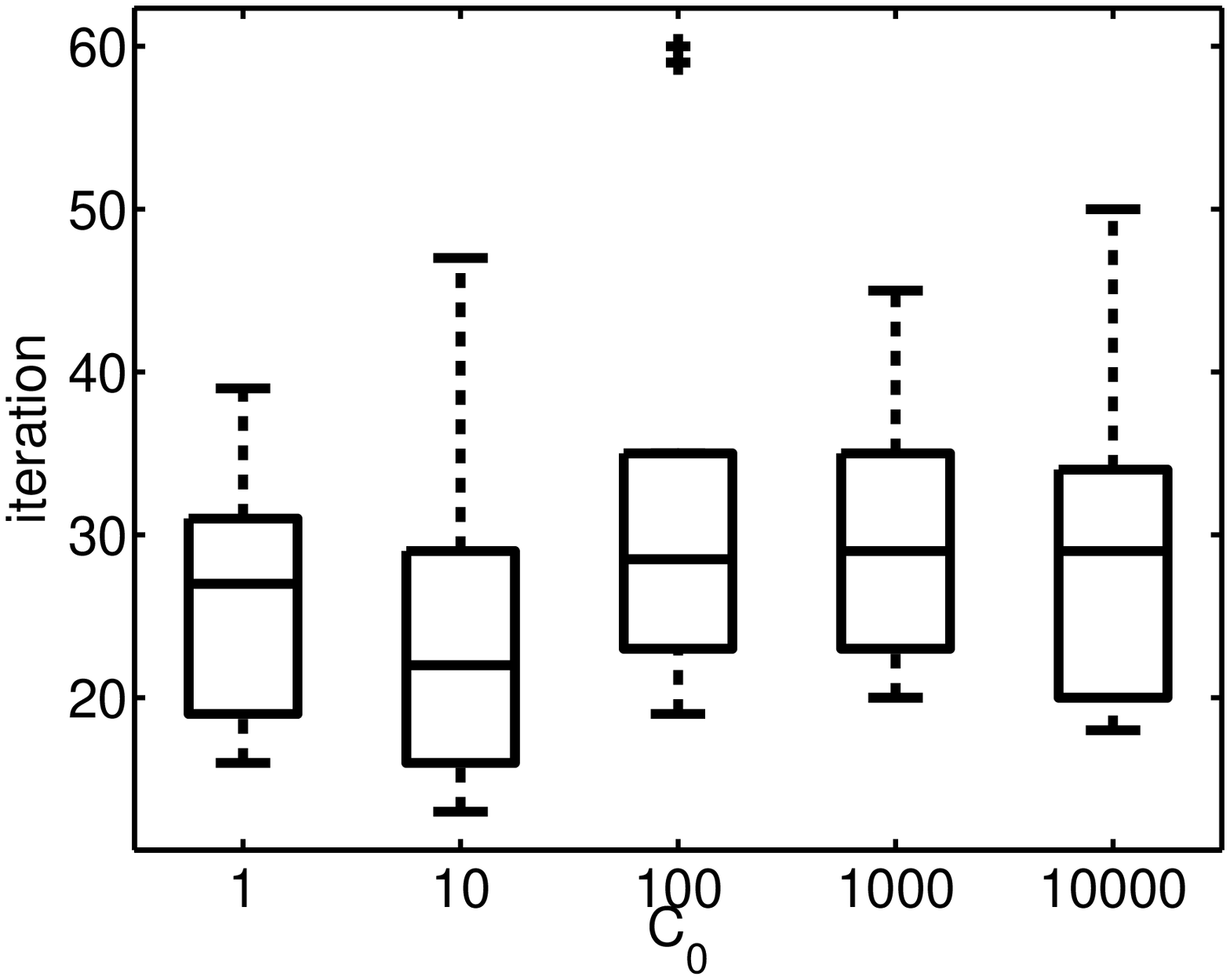}}
 \subfigure[$\gamma = 0.1$]{
 \includegraphics[width=0.3\linewidth]{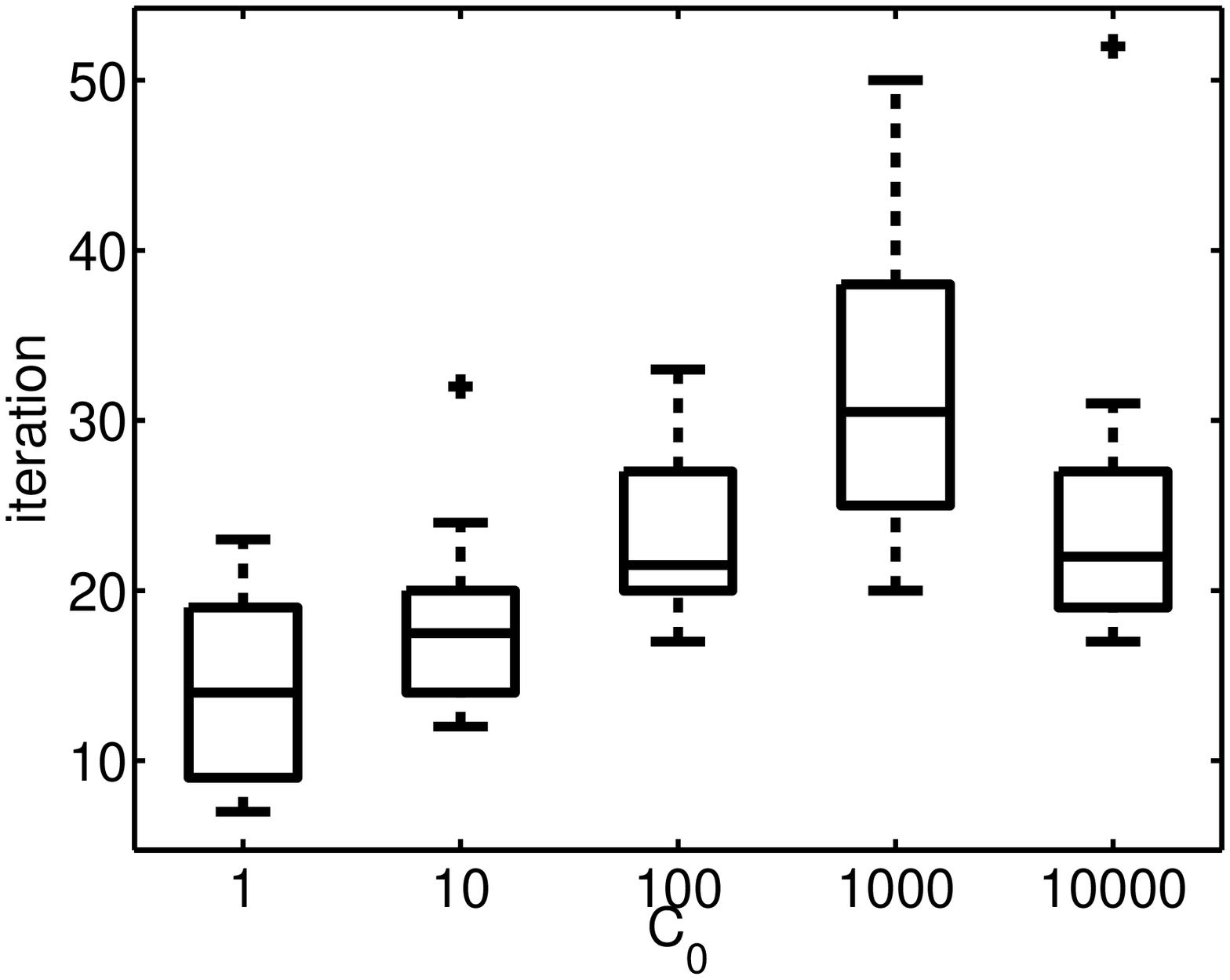}}
 \caption{The number of weight updates for Boston housing data.}
 \label{fig:hetero_iter}
\end{figure}

We investigated the computational cost of Step2.
We applied the above procedure to the well-known \emph{Boston housing} data set.
The sample size is $506$ and the number of features is $p = 13$.
The inputs were normalized in $[-1, 1]^p$.
We randomly sampled $n = 404$ instances from the original
data set, and the experiments were repeated $10$ times.
We used the Gaussian kernel \eqref{Gaussian-kernel} with 
$\gamma \in \{10, 1, 0.1\}$.
The insensitive zone thickness in WSVR
was fixed to $\varepsilon = 0.05$.

Each plot of \figurename~\ref{fig:hetero_cpu} shows the CPU time comparison 
for $C_0 \in \{1, 10, \ldots, 10^4 \}$, and
\figurename~\ref{fig:hetero_iter} shows
the number of iterations performed in Step2.
Our path-following approach is faster than the SMO algorithm
especially for large $C_0$.

% --------------------------------------------------
\subsection{Ranking}
\label{subsec:Ranking}
Recently, the problem of \emph{learning to rank} has attracted wide interest as
a challenging topic in machine learning and information retrieval 
\citep{Liu09}.
Here, we focus on a method called the \emph{ranking SVM} (RSVM) \citep{Herbrich00}.

\subsubsection{Formulation}
Assume that we have a set of $n$ triplets
$\{(\vx_i, y_i, q_i)\}_{i = 1}^n$ where 
$\vx_i \in \mathbb{R}^p$ is a feature vector of an item and 
$y_i \in \{ r_1, \ldots, r_q \}$ is a relevance of $\vx_i$ to 
a query $q_i$.
The relevance has an order of the preference 
$r_q \succ r_{q-1} \succ \cdots \succ r_1$,
where $r_q\succ r_{q-1}$  means that $r_q$ is preferred to $r_{q-1}$.
The goal is to learn a ranking function $f(\vx)$ which 
returns a larger value for a preferred item.
More precisely, for items $\vx_i$ and $\vx_j$ such that $q_i = q_j$,
we want the ranking function $f(\vx)$ to satisfy
% --------------------------------------------------
% Preference and Ranking Function
% --------------------------------------------------
\begin{eqnarray*}
 y_i \succ y_j &\Leftrightarrow& f(\vx_i) > f(\vx_j).
\end{eqnarray*}
Let us define the following set of pairs:
\begin{align*}
 \cP = \{(i, j) \mid y_i \succ y_j, q_i = q_j \}.
 % \{(k_i, \ell_i) \mid \vx_{k_i} \succ \vx_{\ell_i}, q_{k_i} = q_{\ell_i} \}_{i = 1}^m,
\end{align*}
RSVM solves the following optimization problem:
% --------------------------------------------------
% Ranking SVM
% --------------------------------------------------
\begin{eqnarray*}
 \min_{\mb{w}, \{\xi_{ij}\}_{(i,j) \in \cP}} &&
  \frac{1}{2} \|\mb{w}\|_2^2
  + C \sum_{(i,j) \in \cP} \xi_{ij} \\
 {\rm s.t.} &&
 f(\vx_{i}) - f(\vx_{j}) \geq 1 - \xi_{ij}, \ (i,j) \in \cP. 
\end{eqnarray*}

In practical ranking tasks such as information retrieval,
a pair of items 
with highly different preference levels 
should have a larger weight 
than those with similar preference levels. 
Based on this prior knowledge, 
\citet{Cao06} and \citet{Xu06} proposed to assign different weights 
$C_{ij}$
to different relevance pairs $(i, j) \in \cP$.
This is a \emph{cost-sensitive} variant of
RSVM
whose primal problem is given as
% --------------------------------------------------
% Cost-sensitive Ranking SVM
% --------------------------------------------------
\begin{eqnarray*}
 \min_{\mb{w}, \{\xi_{ij}\}_{(i,j) \in \cP}} &&
  \frac{1}{2} \|\mb{w}\|_2^2
  + \sum_{(i,j) \in \cP} C_{ij} \xi_{ij} \\
 {\rm s.t.} &&
 f(\vx_{i}) - f(\vx_{j}) \geq 1 - \xi_{ij}, \ (i,j) \in \cP.
\end{eqnarray*}
Since this formulation is interpreted as a WSVM for
 pairs of items $(i, j) \in \cP$,
we can easily apply our multi-parametric path approach.
% This leads to the following dual problem:
% % --------------------------------------------------
% % Raking SVM Dual
% % --------------------------------------------------
% \begin{eqnarray*}
%  \max_{\{\alpha_i\}_{i=1}^m} && - \frac{1}{2}
%   \sum_{i=1}^m  \sum_{j=1}^m 
%   \alpha_i \alpha_j \widetilde{Q}_{ij}
%   + \sum_{i=1}^m \alpha_i \\
%  {\rm s.t.} &&
%  0 \leq \alpha_i \leq C_i, \ i = 1, \ldots, m,
% \end{eqnarray*}
% where 
% % --------------------------------------------------
% % Q_{ij}
% % --------------------------------------------------
% \begin{eqnarray*}
% \widetilde{Q}_{ij} = K_{k_i,k_j} - K_{k_i,\ell_j}
%  - K_{\ell_i,k_j} + K_{\ell_i,\ell_j}.
% \end{eqnarray*}
% Then, the ranking function $f$ is given as
% \begin{eqnarray*}
%  f(\vx) = \sum_{i=1}^n \alpha_i (K(\vx,\vx_{k_i}) - K(\vx,\vx_{\ell_i})).
% \end{eqnarray*}
% The index partitioning of the optimal solution is given as follows:
% % ------------------------------------------------
% %  Index Sets
% % ------------------------------------------------
% \begin{subequations}
%  \begin{align*}
%   {\cal O} &= \{ i \mid  f(\vx_{k_i}) - f(\vx_{\ell_i}) \geq 1, \alpha_i = 0\}, \\
% %    \label{eq:ranking_outside} \\
%   {\cal M} &= \{ i \mid  f(\vx_{k_i}) - f(\vx_{\ell_i}) = 1, 0 < \alpha_i < C_i \}, \\
% %   \label{eq:ranking_margin} \\
%   {\cal I} &= \{ i \mid  f(\vx_{k_i}) - f(\vx_{\ell_i}) \leq 1, \alpha_i = C_i \}. 
% %   \label{eq:ranking_inside} 
%  \end{align*}
% \end{subequations}
% Using these sets, we can compute the solution path for changing
% weight vector $\mb{c}$.
Note that the solution path algorithm for
the cost-sensitive RSVM is regarded as
an extension of the previous work by \citet{Arreola08},
in which the solution path for the standard RSVM was studied.

In this paper,
we consider a model selection problem for the weighting pattern
$\{C_{ij}\}_{(i, j) \in \cP}$. 
We assume that the weighting pattern is represented as 
\begin{eqnarray}
 C_{ij} = C_{ij}^{\rm (old)} + \theta (C_{ij}^{\rm (new)} - C_{ij}^{\rm (old)}),
 \;
 (i, j) \in \cP, 
 \;
 \theta \in [0, 1],
\end{eqnarray}
where 
\begin{eqnarray}
C^{({\rm old})}_{ij} &=& C_0, \ (i,j) \in \cP,
\label{eq:rsvm_flat_weight}
\\
C^{({\rm new})}_{ij} &=& (2^{y_{i}} - 2^{y_{j}}) C_0, \ (i,j) \in \cP,
\label{eq:rsvm_rel_weight}
\end{eqnarray}
and 
$C_0$ 
is the common regularization parameter\footnote{In
\citet{Chapelle10}, ranking of each item is also incorporated to
define the weighting pattern. However, these weights
depend on the current ranking, and it might change during training.
We thus, for simplicity, introduce the weighting pattern
(\ref{eq:rsvm_rel_weight}) that depends only on the difference of the
preference levels.}. 
We follow the multi-parametric solution path from 
$\{C^{({\rm old})}_{ij}\}_{(i, j) \in \cP}$ 
to 
$\{C^{({\rm new})}_{ij}\}_{(i, j) \in \cP}$ 
and the best 
$\theta$ 
is selected based on the validation performance. 

The performance of ranking algorithms is usually evaluated 
by some information-retrieval measures such as the 
\emph{normalized discounted cumulative gain} (NDCG) \citep{Jarvelin00}.
Consider a query 
$q$ 
and define 
$q(j)$ 
as the index of the $j$-th largest item among 
$\{f(\bm{x}_i)\}_{i \in \{i | q_i = q\}}$. 
The NDCG at position 
$k$ 
for a query 
$q$ 
is defined as 
\begin{align}
 \mathrm{NDCG}@k = Z \sum_{j = 1}^k
  \left\{
   \begin{array}{cc}
    2^{y_{q(j)}} - 1, & j = 1, \\
    \frac{2^{y_{q(j)}} - 1}{\log(j)}, & j > 1,
   \end{array}
  \right. 
  \label{eq:ndcg}
\end{align}
where 
$Z$ 
is a constant to normalize the NDCG in $[0, 1]$. 
Note that 
the NDCG value in 
\eq{eq:ndcg} 
is defined using only the top $k$ items and the rest are ignored.
The NDCG for multiple queries are defined as the average of (\ref{eq:ndcg}).

\begin{figure}[t]
\centering
\includegraphics[width=4in]{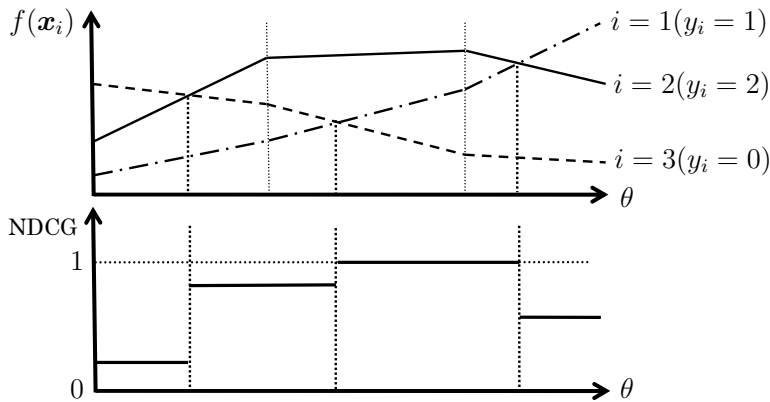}
\caption{The schematic illustration of the NDCG path.
The upper plot shows outputs for $3$ items which have
different levels of preferences $y$. The bottom plot shows the changes
of the NDCG. 
Since the NDCG depends on the sorted order of items, 
it changes only when two lines of the upper plot intersect.}
\label{fig:ndcg_path}
\end{figure}

The goal of our model selection problem is 
to choose $\theta$ with the largest NDCG value. 
As explained below,
we can identify $\theta$ that attains the exact maximum NDCG value 
for validation samples
by exploiting the piecewise linearity of the solution path.  
The NDCG value changes only when there is a change in the top $k$ ranking,
and the rank of two items 
$\bm{x}_i$ 
and 
$\bm{x}_j$ 
changes
only when 
$f(\bm{x}_i)$ 
and 
$f(\bm{x}_j)$ 
cross.
Then change points of the NDCG can be exactly identified 
because $f(\bm{x})$ 
changes in piecewise-linear form. 
\figurename~\ref{fig:ndcg_path} 
schematically illustrates piecewise-linear paths 
and the corresponding NDCG path for validation samples. 
The validation NDCG changes in piecewise-constant form, 
and change points are found when there is a crossing between two
piecewise-linear paths. 

\subsubsection{Experiments on ranking}

We used the OHSUMED data set from the LETOR package
(version 3.0) provided by 
Microsoft Research Asia \citep{Liu07}.
We used the query-level normalized version of the data set containing $106$ queries.
The total number of query-document pairs is
$16140$, and the number of features is $p = 45$.
The data set provided is originally partitioned into $5$ subsets, % of data set, 
each of which has training, validation, and
test sets for $5$-fold cross validation.
Here, we only used the training and the validation sets.

\begin{figure}[p]
 \centering
 \subfigure[$\gamma = 10$]
 {\includegraphics[width=0.3\linewidth]{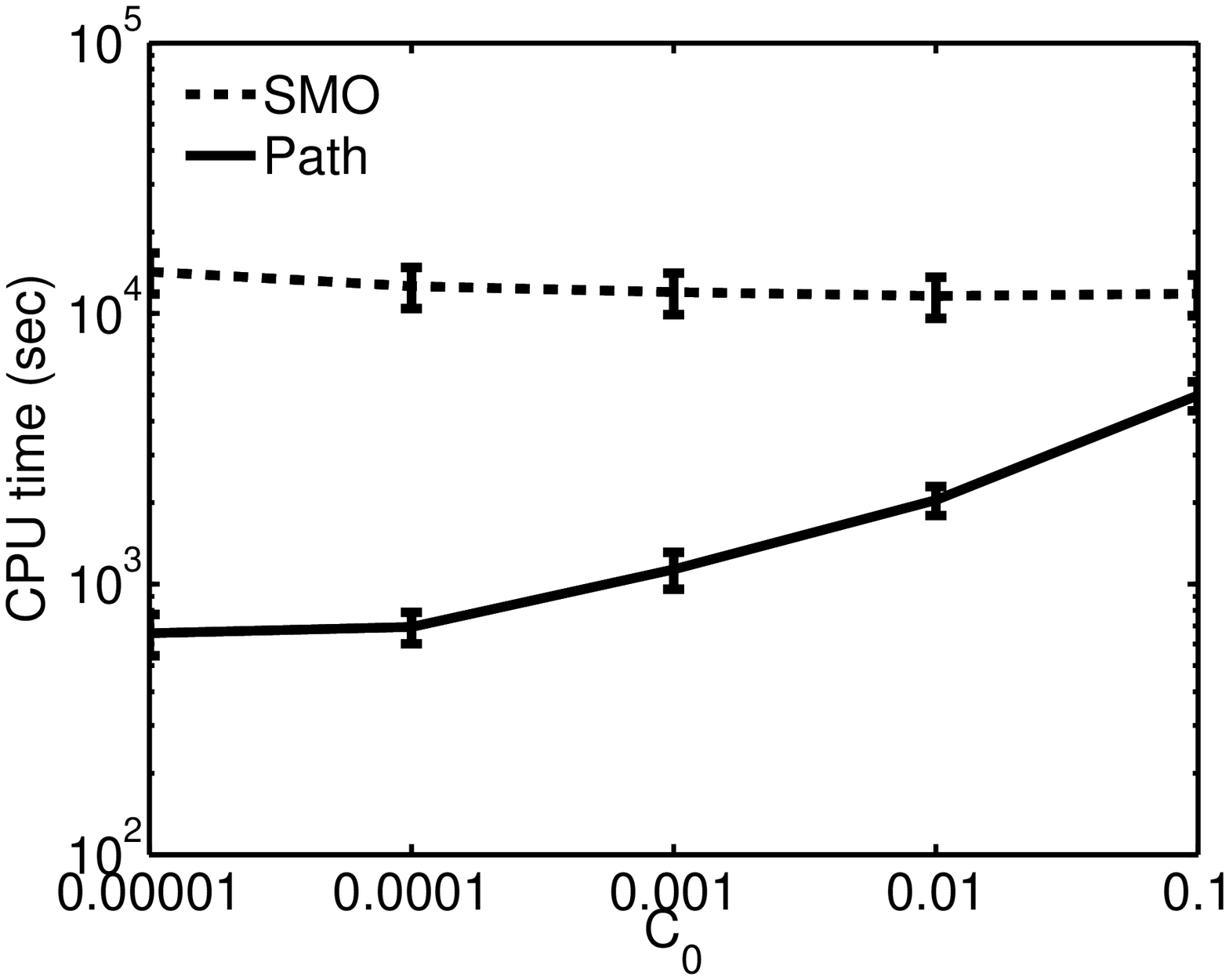}
 \label{fig:rsvm_time_a}}
 \subfigure[$\gamma = 1$]
 {\includegraphics[width=0.3\linewidth]{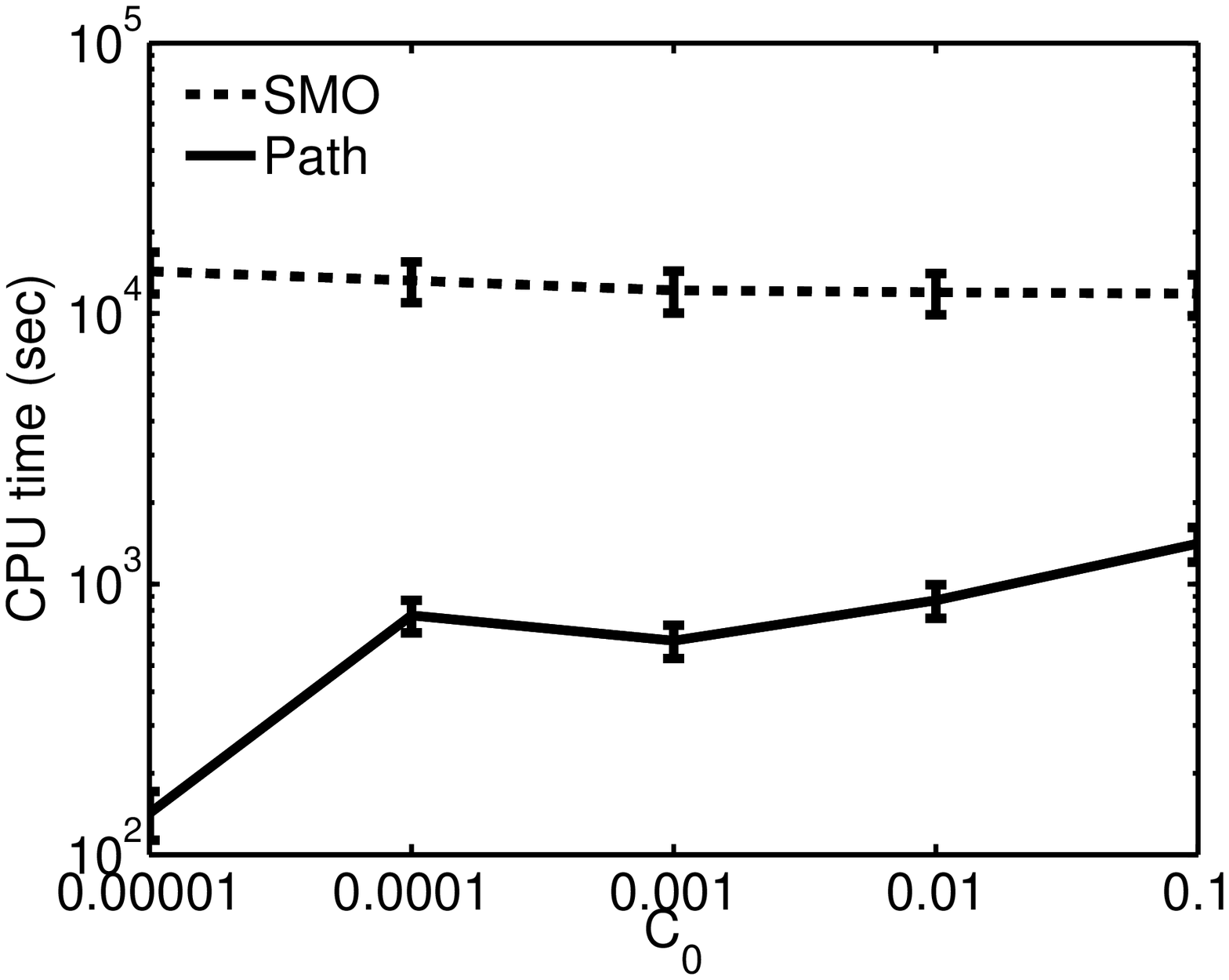}}
 \subfigure[$\gamma = 0.1$]
 {\includegraphics[width=0.3\linewidth]{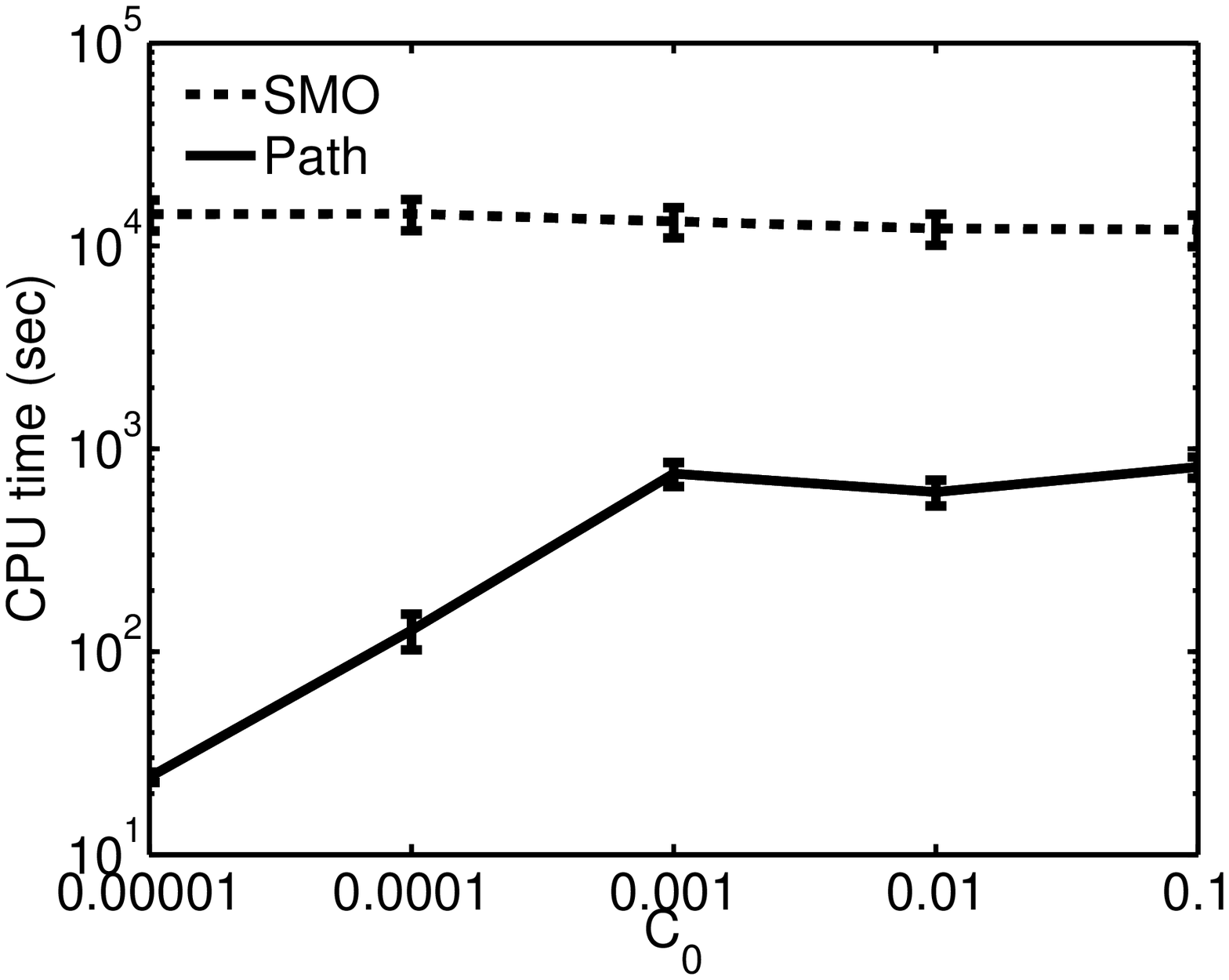}}
 \caption{CPU time comparison for RSVM.}
 \label{fig:rsvm_time}
% \end{figure}
% \begin{figure}[!t]
% \centering
\vspace*{5mm}
 \includegraphics[width=3.5in]{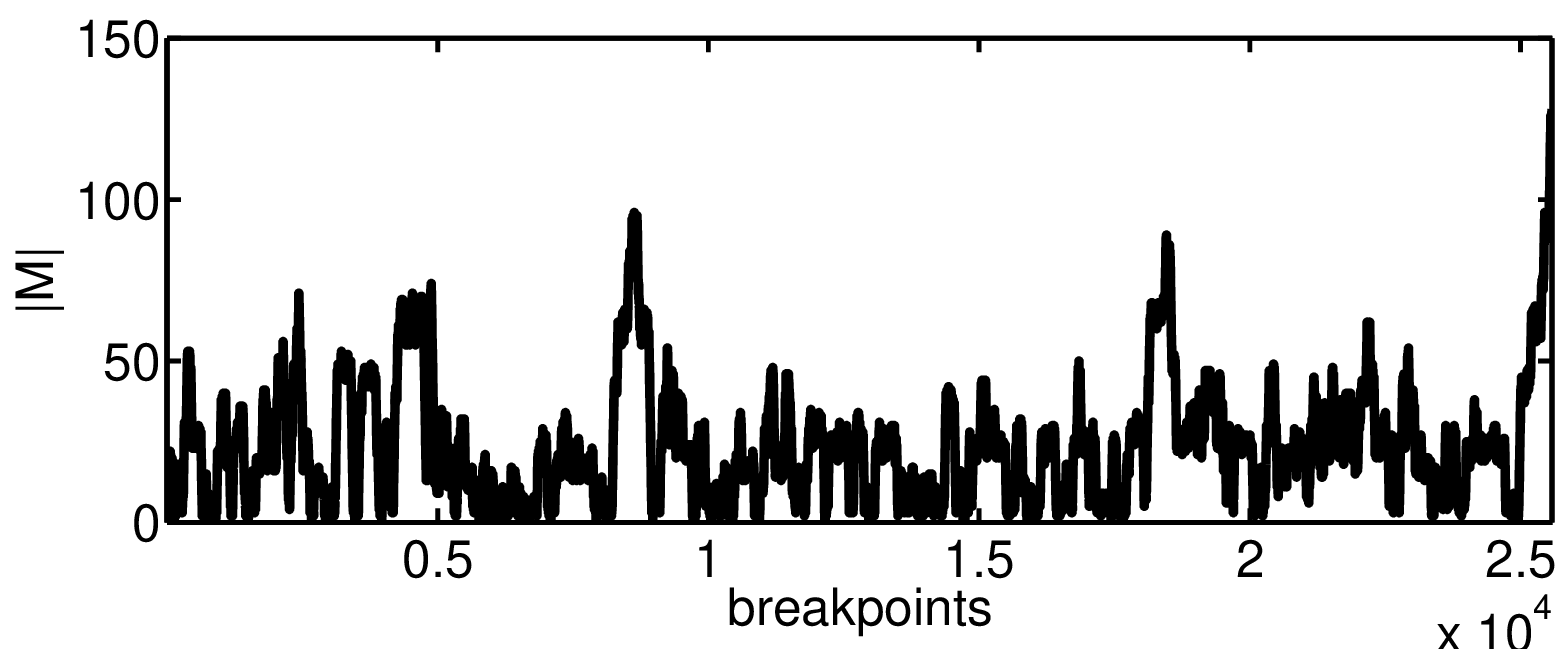}
 \caption{The number of instances on the margin $|{\cal M}|$ in ranking
 experiment.}
 \label{fig:rsvm_M}
% \end{figure}
% \begin{figure}[t]
% \centering
\vspace*{5mm}
 \includegraphics[width=3.5in]{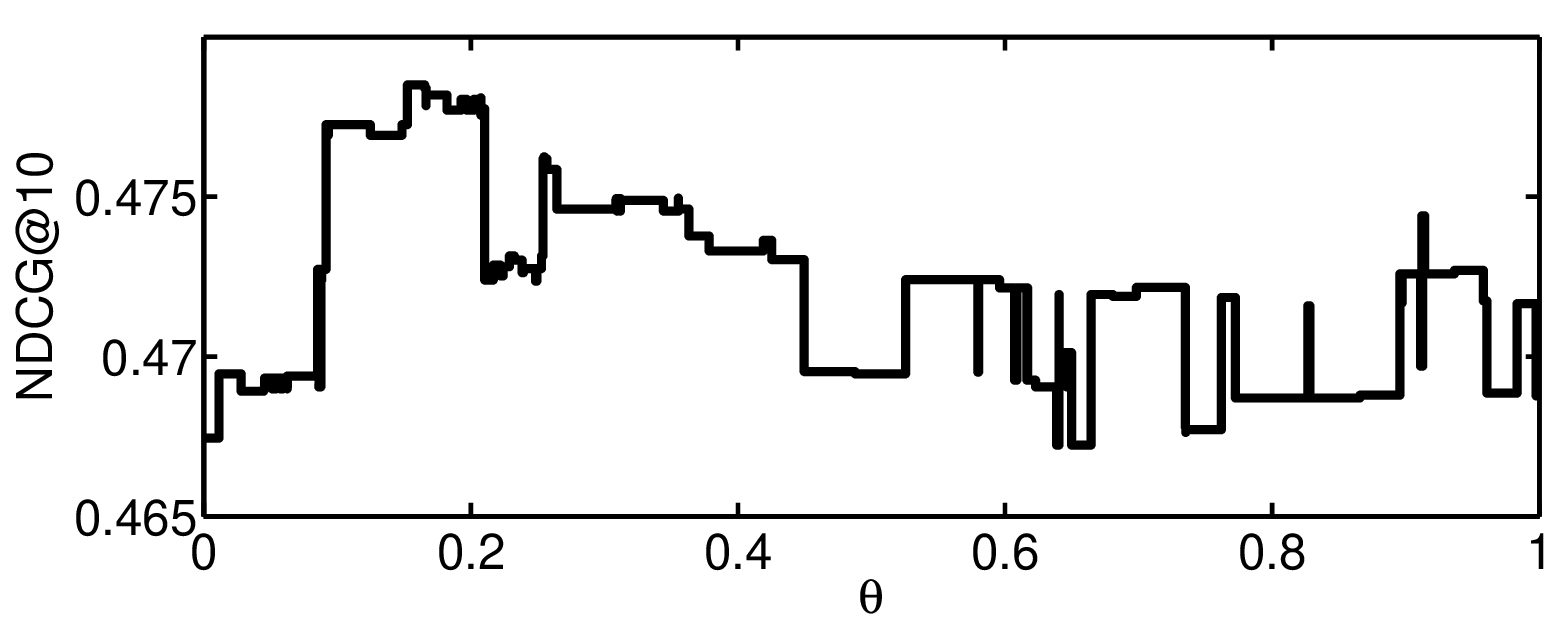}
 \caption{The change of NDCG@10 for $\gamma = 0.1$ and $C = 0.01$.
 The parameter $\theta$ in the horizontal axis is used as
 $\mb{c}^{\rm (old)} + \theta (\mb{c}^{\rm (new)} - \mb{c}^{\rm (old)})$.}
 \label{fig:rsvm_ndcg}
\end{figure}

We compared the CPU time of our path algorithm and the SMO algorithm to
change $\{C_{ij}\}_{(i,j) \in \cP}$ from flat ones (\ref{eq:rsvm_flat_weight})
 to relevance weighted ones (\ref{eq:rsvm_rel_weight}).
We need to modify the SMO algorithm to train the model without the
explicit bias
term $b$.
The usual SMO algorithm updates selected two parameters per iteration
to ensure that the solution satisfies the equality constraint derived
from the optimality condition of $b$.
Since RSVM has no bias term, the algorithm is adapted to update one
parameter per iteration \citep{Vogt02}.
We employed the update rule of \citet{Vogt02} to adapt the SMO
algorithm to RSVM and we chose the maximum violating point as an
update parameter. This strategy is analogous to the maximum violating-pair
working set selection of \citet{Keerthi01} in ordinary SVM.
Since it took relatively large computational time, we ran 
the SMO algorithm only at $10$ points uniformly 
taken in $[C_{ij}^{\rm (old)}, C_{ij}^{\rm (new)}]$.
We considered every pair of
initial weight $C_0 \in \{10^{-5}, \ldots, 10^{-1}\}$ and
Gaussian width $\gamma \in \{10, 1, 0.1\}$.
The results, given in \figurename~\ref{fig:rsvm_time},
show that the path algorithm is faster than the SMO in all of the settings.

The CPU time of the path algorithm in \figurename~\ref{fig:rsvm_time_a}
increases as $C_0$ increases
because the number of breakpoints 
and the size of the set ${\cal M}$ also increase.
Since our path algorithm solves a linear
system with size $|\cM|$ using $O(|{\cal M}|^2)$ update in each iteration,
practical computational time depends on $|{\cal M}|$
especially in large data sets.
In the case of RSVM, the maximum value of $|{\cal M}|$ is the 
number of pairs of training documents $m = |\cP|$.
For each fold of the OHSUMED data set,
$m = 367663$, $422716$, $378087$, $295814$, and $283484$.
If $|{\cal M}| \approx m$, a large computational cost may be needed for updating
the linear system.
However,
%  in many cases, $|{\cal M}|$ was much smaller than $m$.
as \figurename~\ref{fig:rsvm_M} shows,
% an example of the change of $|{\cal M}|$ in the path.
the size $|{\cal M}|$ is at most about one hundred in this
setup.

\figurename~\ref{fig:rsvm_ndcg} shows the example of
% changes in
the path of validation
NDCG@10. % for validation data set.
Since the NDCG depends on the sorted order of documents,
direct optimization is rather difficult \citep{Liu09}.
Using our path algorithm, however, 
we can detect the exact behavior of the NDCG by monitoring the change
of scores $f(\vx)$ in the validation data set.
Then we can find the best weighting pattern 
by choosing $\theta$ with the maximum NDCG for the validation set.

% --------------------------------------------------
\subsection{Transduction} \label{subsec:tsvm}
In \emph{transductive inference} \citep{book:Vapnik:1995}, 
we are given unlabeled instances
along with labeled instances.
The goal of transductive inference is not to estimate the true
decision function, but to classify the given unlabeled instances correctly.
The \emph{transductive SVM} (TSVM) \citep{Joachims99} is one of the most 
popular approaches to %the $2$-class transduction 
transductive binary classification.
The objective of the TSVM is to maximize the classification margin for
both labeled and
unlabeled instances.

\subsubsection{Formulation} 
Suppose we have $k$ unlabeled instances $\{ \vx_i^* \}_{i=1}^k$
in addition to
$n$ labeled instances $\{ (\vx_i, y_i) \}_{i=1}^n$.
The optimization problem of TSVM is formulated as %given as follows:
%
% --------------------------------------------------
% Optimization Problem of TSVM
% --------------------------------------------------
\begin{eqnarray}
 \min_{\{ y^*_i, \xi^*_i \}_{i = 1}^k, \mb{w}, b, \{ \xi_i \}_{i = 1}^n} 
  && \frac{1}{2} \| \mb{w} \|_{2}^{2} + 
  C \sum_{i = 1}^n \xi_i +
  C^* \sum_{j = 1}^k \xi_j^* \label{eq:tsvm} \\
 {\rm s.t.} &&
 \begin{array}{lc}
   y_i ( \mb{w}^\top \mb{\Phi}(\vx_i) + b ) \geq 1 - \xi_i, & i = 1,\ldots, n, \\
   y^*_j ( \mb{w}^\top \mb{\Phi}(\vx_j) + b ) \geq 1 - \xi^*_j, & j = 1, \ldots, k, \\
   \xi_i \geq 0, & i = 1,\ldots, n, \\
   \xi^*_j \geq 0, & j = 1,\ldots, k,
 \end{array}
 \nonumber
\end{eqnarray}
where $C$ and $C^*$ are the regularization parameters for
labeled and unlabeled data, respectively, and
$y^*_j \in \{-1, +1\}, j = 1, \ldots, k,$ are the labels of the
unlabeled instances
$\{ \vx_i^* \}_{i=1}^k$.
Note that (\ref{eq:tsvm}) is a combinatorial optimization problem
with respect to $\{ y_j^* \}_{j \in \{ 1, \ldots, k \}}$. The optimal
solution of (\ref{eq:tsvm}) can be found if we solve binary SVMs for all
possible combinations of $\{y_j^*\}_{j \in \{1,\ldots,k\}}$, but 
this is computationally intractable even for moderate $k$.
To cope with this problem, \citet{Joachims99} proposed an algorithm
which approximately optimizes
(\ref{eq:tsvm})
by solving a series of WSVMs.
The subproblem is formulated by assigning temporarily estimated labels 
$\widehat{y}_j^*$ to unlabeled instances:
% --------------------------------------------------
% Sub-Optimization Problem of TSVM
% --------------------------------------------------
\begin{eqnarray}
 \min_{\{ \xi^*_i \}_{i = 1}^k, \mb{w}, b, \{ \xi_i \}_{i = 1}^n} 
  && \frac{1}{2} \| \mb{w} \|_{2}^{2} + 
  C \sum_{i = 1}^n \xi_i +
  C^*_{-} \sum_{j \in \{ j | \widehat{y}_j^* = -1 \}} \xi_j^* +
  C^*_{+} \sum_{j \in \{ j | \widehat{y}_j^* = 1 \}} \xi_j^* 
  \label{eq:tsvm_sub} \\
 {\rm s.t.} &&
 \begin{array}{lc}
   y_i ( \mb{w}^\top \mb{\Phi}(\vx_i) + b ) \geq 1 - \xi_i, & i = 1,\ldots, n, \\
   \widehat{y}^*_j ( \mb{w}^\top \mb{\Phi}(\vx_j) + b ) \geq 1 - \xi^*_j, & j = 1, \ldots, k, \\
   \xi_i \geq 0, & i = 1,\ldots, n, \\
   \xi^*_j \geq 0, & j = 1,\ldots, k,
 \end{array}
 \nonumber
\end{eqnarray}
where $C_{-}^*$ and $C_{+}^*$ are the weights for 
unlabeled instances for
$\{j \mid \widehat{y}_j^* = -1 \}$ and
$\{j \mid \widehat{y}_j^* = +1 \}$, respectively.
The entire algorithm is given as follows
\citep[see][for details]{Joachims99}:
% --------------------------------------------------
% Algorithm of TSVM
% --------------------------------------------------
\begin{description}
 \item[Step1:] Set the parameters
	    $C$, $C^*$, and $k^+$, where 
	    $k^+$ is defined as 
            \begin{align*}
              k^+ = k \times \frac{|\{j \mid y_j = +1, j = 1,\ldots, n\}|}{n}.
            \end{align*}
	    $k^+$ is defined so that the balance of positive and
	    negative instances in the labeled set is equal to that in
	    the unlabeled set.
 \item[Step2:] Optimize the decision function using only the labeled instances
	    and compute the decision function values $\{f(\vx_j^*)\}_{j=1}^k$.
	    Assign positive label $y_j^* = 1$ to 
	    the top $k^+$ unlabeled instances in decreasing order of $f(\vx_j^*)$,
	    and negative label $y_j^* = -1$ to the remaining instances.
	    Set $C_-^*$ and $C_+^*$ to some small values 
	    \citep[see][for details]{Joachims99}.
 \item[Step3:] Train SVM using all the instances (i.e., solve
	    (\ref{eq:tsvm_sub})). 
	    Switch the labels of a pair of positive and negative
	    unlabeled instances if the objective value
	    (\ref{eq:tsvm}) is reduced, where the pair of instances
	    are selected based on 
	    $\{\xi_j^*\}_{j \in \{1,\ldots,k\}}$
	    \citep[see][for details]{Joachims99}.
	    Iterate this step until no data
	    pair decreases the objective value.
 \item[Step4:] Set $C_-^*=\min(2 C_-^*, C^*)$ and $C_+^*=\min(2 C_+^*, C^*)$.
	    If $C_-^* \geq C^*$ and $C_+^* \geq C^*$, terminate the
	    algorithm.
	    Otherwise return to Step3.
\end{description}

Our path-following algorithm can be applied to Step3 and Step4
for improving computational efficiency.
Step3 can be carried out via path-following as follows:
\begin{description}
\item[Step3(a)] Choose a pair of positive instance $\vx_m^*$ and
	   negative instance $\vx_{m'}^*$.
\item[Step3(b)] After removing the positive instance $\vx_m^*$ 
	   by decreasing its weight parameter $C_m$ from $C_+^*$ to $0$,
 %\item[Step3(c)]  
	   add the instance $\vx_m^*$ as a negative one
	   by increasing $C_m$ from $0$ to  $C_-^*$.
\item[Step3(c)] After removing the negative instance $\vx_{m'}^*$ 
	   by decreasing its weight parameter $C_{m'}$ from $C_-^*$ to $0$,
% \item[Step3(e)]  
	   add the instance $\vx_{m'}^*$ as a positive one
	   by increasing $C_{m'}$ from $0$ to  $C_+^*$.
\end{description}
Note that the steps 3(b) and 3(c) for switching the labels may be merged into a single step.
Step4 also can be carried out
by our path-following algorithm.

%==================================================
% Experiments on Transduction
%==================================================
\subsubsection{Experiments on Transduction}

We compare the computation time of the proposed path-following algorithm
and the SMO algorithm for Step3 and Step4 of TSVM. 
We used the \emph{spam} data set obtained from the \emph{UCI machine learning repository}
\citep{Asuncion07}.
The sample size is $4601$, and the number of features is $p = 57$.
We randomly selected $10\%$ of data set as labeled instances,
and the remaining $90\%$ were used as unlabeled instances.
The inputs were normalized in $[0, 1]^p$.

\begin{figure}[!t]
 \centering
 \subfigure[$\gamma = 10$]
 {\includegraphics[width=0.3\linewidth]{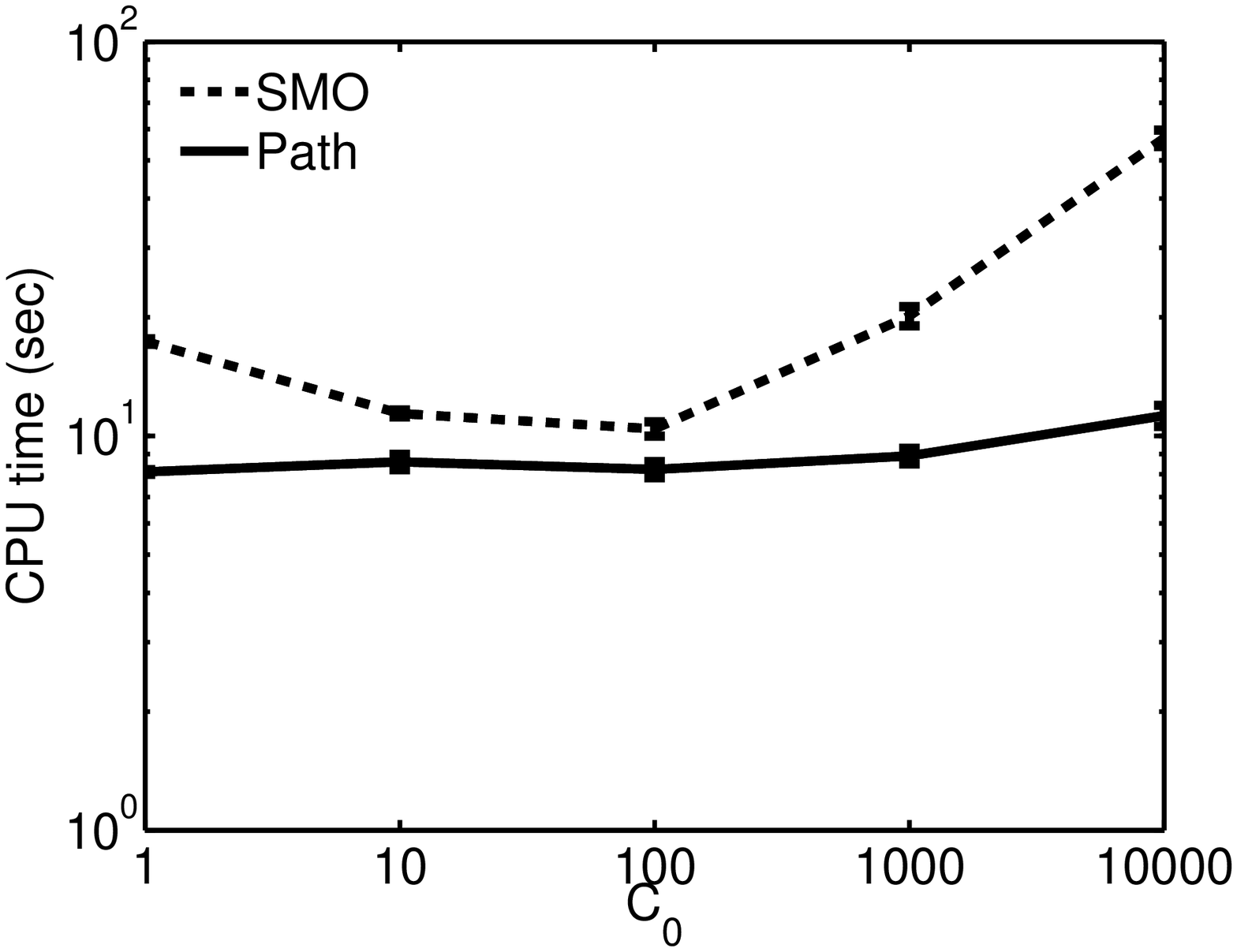}}
 \subfigure[$\gamma = 1$]
 {\includegraphics[width=0.3\linewidth]{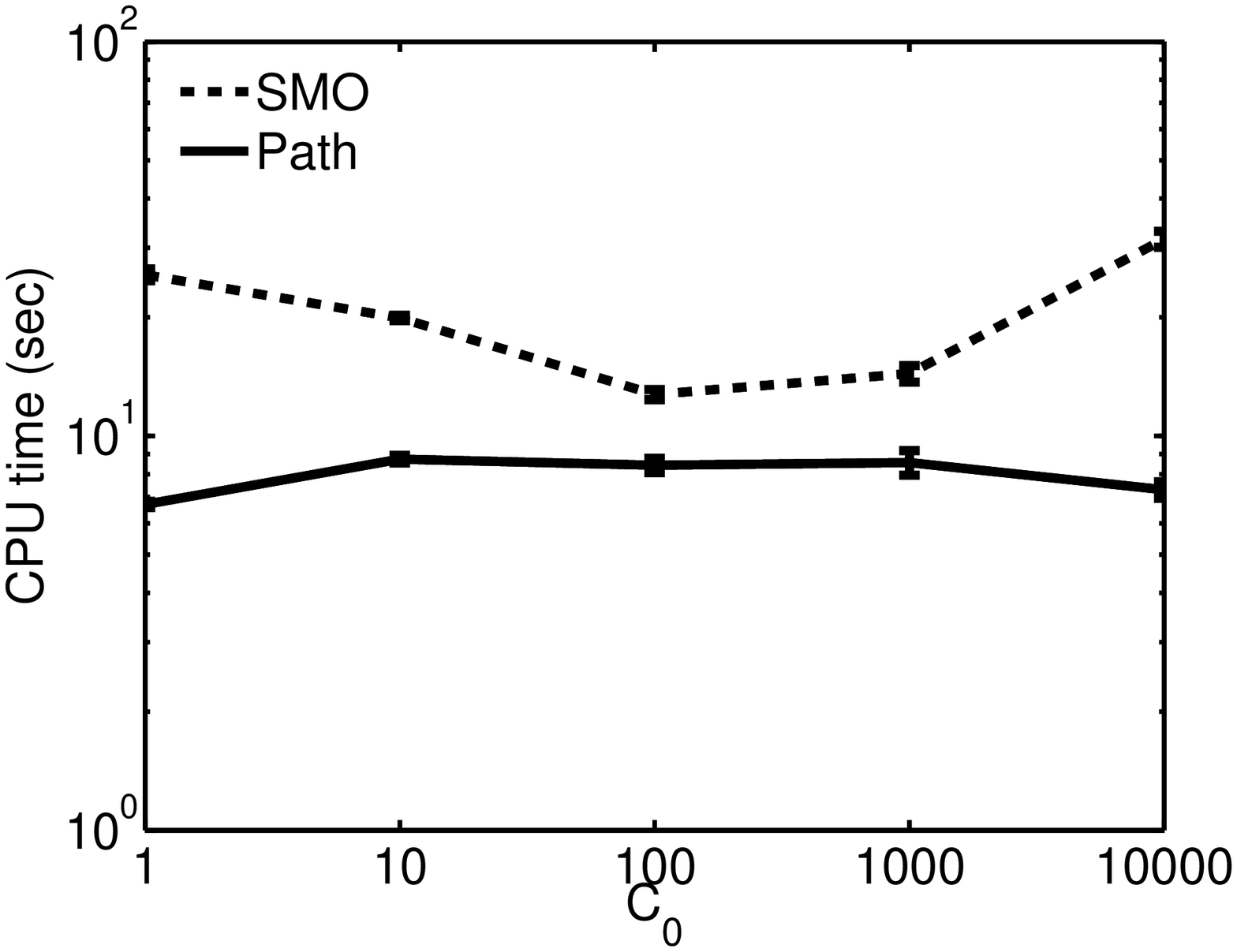}}
 \subfigure[$\gamma = 0.1$]
 {\includegraphics[width=0.3\linewidth]{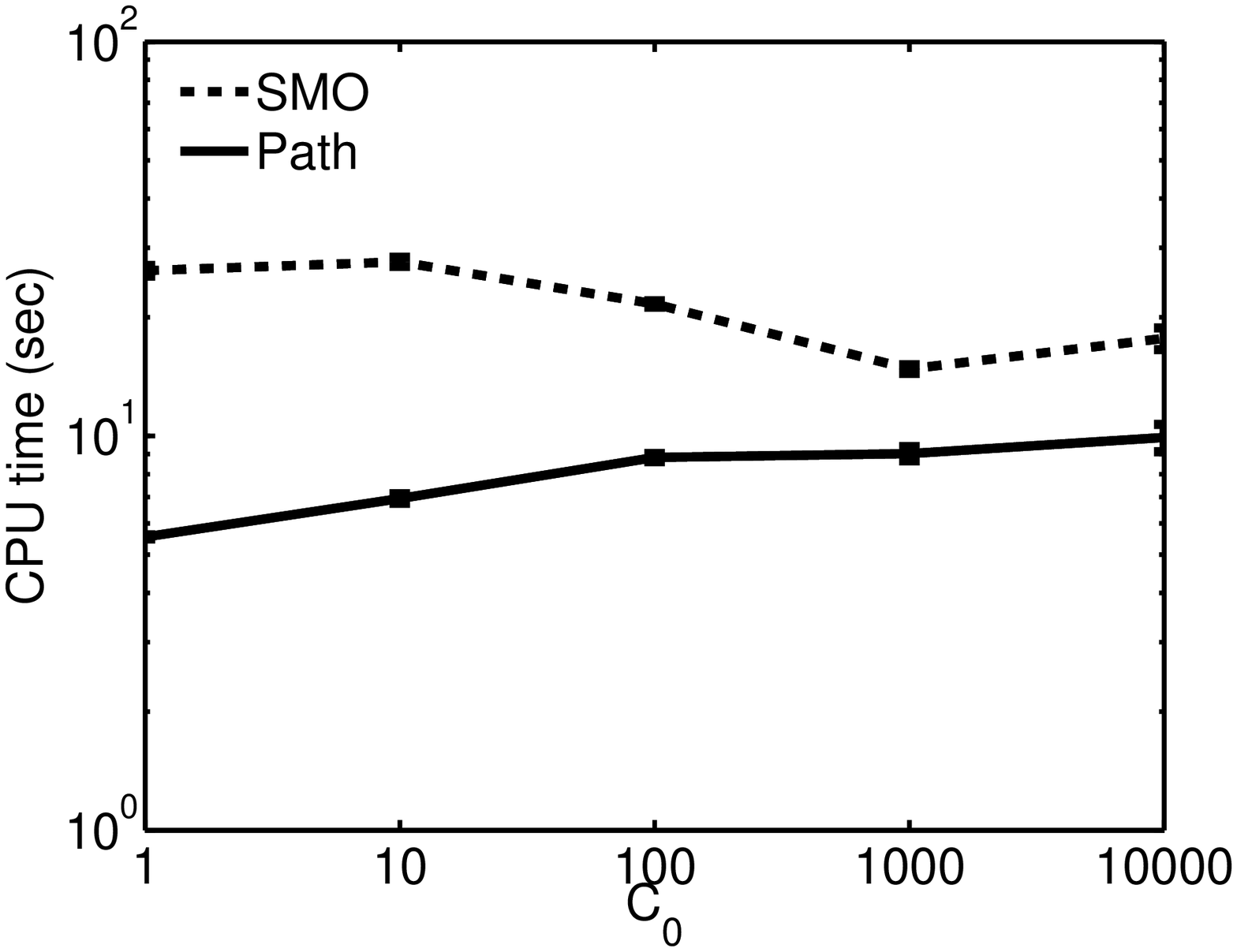}}
 \caption{CPU time comparison for the transductive SVM.}
 \label{fig:tsvm_time}
\end{figure}

\figurename~\ref{fig:tsvm_time} shows the average CPU time and its
standard deviation over $10$ runs for the Gaussian width $\gamma \in \{10, 1, 0.1\}$
and $C \in \{1, 10, 10^2, \ldots, 10^4\}$.
The figure shows that our algorithm is 
consistently faster than the SMO algorithm in all of these settings.

%\subsection{Outlier Detection}

%%% Local Variables: 
%%% mode: latex
%%% TeX-master: "paper"
%%% End: 

%==================================================
% Sec::Conclusion
%==================================================
\section{Discussion and Conclusion} \label{sec:conclusion}
In this paper, we developed an efficient algorithm for updating 
solutions of instance-weighted SVMs. 
Our algorithm was built upon multiple parametric programming techniques,
and it is an extension of existing single-parameter path-following
algorithms to multiple parameters. 
We experimentally demonstrated the computational advantage of the proposed algorithm
on a wide range of applications
including 
on-line time-series analysis, 
heteroscedastic data modeling, 
covariate shift adaptation, 
ranking, 
and 
transduction. 

Another important advantage of the proposed approach beyond computational efficiency
is that
the exact solution
path is represented in piecewise linear form.
In SVM (and its variants),
the decision function 
$f(\bm{x})$ 
also has a piecewise linear form 
because 
$f(\bm{x})$ 
is linear in 
the parameters 
$\bm{\alpha}$ 
and 
$b$.
It enables us to compute the entire path of the validation errors 
and to select the model with the minimum validation error. 
Let 
$\cV$ 
be the validation set 
and 
the validation loss is defined as
$\sum_{i \in \cV} \ell(y_i, f(\bm{x}_i))$.
Suppose that 
the current weight is expressed as 
$\bm{c} + \theta \bm{d}$
in a critical region 
$\cR$ using a parameter 
$\theta \in \mathbb{R}$, 
and the output has the form 
$f(\bm{x}_i) = a_i \theta + b_i$ 
for some scalar constants
$a_i$ 
and 
$b_i$.
Then the minimum validation error in the current critical region 
$\cR$ 
can be identified by solving the following optimization problem: 
\begin{eqnarray}
 \label{eq:validation.critical.region}
 \min_{\theta \in \mathbb{R}} \; \sum_{i \in \cV} \ell(y_i, a_i \theta + b_i)
 \;\;
 \mbox{s.t.} \; \mb{c} + \theta \mb{d} \in \cR.
\end{eqnarray}
After following the entire solution-path, 
the best model can be selected 
among the candidates in the finite number of critical regions.
In the case of 
0-1 loss, 
i.e., 
\begin{align*}
  \ell(y_i, f(\bm{x}_i)) = I\{y_i {\rm sgn}(f(\bm{x}_i)) = -1\},
\end{align*}
the problem \eq{eq:validation.critical.region}
can be solved 
by monitoring all the points at which 
$f(\bm{x}_i) = 0$ 
(see \figurename~\ref{fig:cerror_path}).
Furthermore, 
if the validation error is measured by squared loss 
\begin{align*}
  \ell(y_i, f(\bm{x}_i)) = (y_i - f(\bm{x}_i))^2, 
\end{align*}
the problem 
\eq{eq:validation.critical.region} 
can be analytically solved in each critical region.
As another interesting example, 
we described how to find the maximum validation NDCG in
ranking problem (see Section~\ref{subsec:Ranking}).
In the case of NDCG,
the problem 
\eq{eq:validation.critical.region}
can be solved 
by monitoring all the intersections of 
$f(\bm{x}_i)$ 
and 
$f(\bm{x}_j)$ 
such that 
$y_i \neq y_j$
(see \figurename~\ref{fig:ndcg_path})\footnote{
In NDCG case, 
the ``$\min$'' is replaced with ``$\max$''
in the optimization problem 
\eq{eq:validation.critical.region}.
}.

Although we focused only on quadratic programming (QP) machines in this paper,
similar algorithms can be developed for 
linear programming (LP) machines.
It is well-known in the parametric programming literature that 
the solution of LP and QP have piecewise linear form 
if a linear function of hyper-parameters 
appears in the constant term of the constraints 
and/or 
the linear part of the objective function
\citep[see][for more details]{Ritter84,UWaterloo:Best:1982,Gal95,Pistikopoulos07}.
Indeed, 
the parametric LP technique \citep{Gal95} has already been applied to
several machine learning problems \citep{Zhu03,Yao07,Li08}. 
One of our future works is to apply the multi-parametric approach to these LP machines.

In this paper, we studied the changes of instance-weights of
various types of SVMs.
%However, 
There are many other situations in which a path of multiple hyper-parameters can be
exploited.
For instance, 
the application of the multi-parametric path approach
to the following problems would be interesting future works:
\begin{itemize}
 \item Different (functional) margin SVM:
	\begin{eqnarray*}
	 \begin{split}
	  \min_{\mb{w}, b, \{\xi_i\}_{i=1}^n} ~~& \frac{1}{2} \| \mb{w} \|_2^2 
	  + C \sum_{i=1}^n \xi_i, \\
	  {\rm s.t. }        ~~& y_i f(\bm{x}_i) \geq \delta_i - \xi_i, 
	  \  \xi_i \geq 0, \ i = 1, \ldots, n,
	 \end{split}
	\end{eqnarray*}
       where $\delta_i \in \mathbb{R}$ is a margin rescaling parameter.
       \citet{Chapelle10} indicated that this type of parametrization can
       be used to give different costs to each pair of items in ranking
       SVM.
 \item SVR with different insensitive-zone thickness.
       Although usual SVR has the common insensitive-zone thickness
       $\varepsilon$ for all instances, 
       different thickness $\varepsilon_i$
       for every instance
       can be assigned:
       \begin{eqnarray*}
	\min_{\mb{w}, b, \{\xi_i, \xi_i^*\}_{i=1}^n} && \frac{1}{2} \| \mb{w} \|_2^2 
	 + C \sum_{i=1}^n (\xi_i + \xi_i^*), \\
	 {\rm s.t.}  
	 && y_i - f(\vx_i) \leq \varepsilon_i + \xi_i, \\
	&& f(\vx_i) - y_i \leq \varepsilon_i + \xi_i^*, \\
	&& \xi_i, \xi_i^* \geq 0, \ i = 1, \ldots, n.
       \end{eqnarray*}
       In the case of common thickness,
       the optimal $\varepsilon$ is known to be asymptotically
       proportional to the noise
       variance \citep{Smola98b,Kwok03}. 
       In the case of heteroscedastic noise,
       it would be reasonable to set different $\varepsilon_i$, each of which
       is proportional to the variance of each (iteratively estimated) $y_i$.
 \item The weighted lasso.
       The lasso solves the following quadratic programming problem \citep{JRSS:Tibshirani:1996}:
       \begin{eqnarray*}
	\min_{\mb{\beta}} 
	 \| \vy - \sum_{j = 1}^p \vx_j \beta_j \|^2_2 
	 + \lambda \sum_{j = 1}^p | \beta_j |,
       \end{eqnarray*}
       where $\lambda$ is the regularization parameter.
       A weighted version of the lasso has been considered in 
       \citet{Zou06}:
       \begin{eqnarray*}
	\min_{\mb{\beta}} 
	 \| \vy - \sum_{j = 1}^p \vx_j \beta_j \|^2_2 
	 + \lambda \sum_{j = 1}^p w_j | \beta_j |,	
       \end{eqnarray*}
       where $w_j$ is an individual weight parameters for each
       $\beta_j$.
       The weights are adaptively determined by
       $w_j = |\hat{\beta}_j|^{-\gamma}$, where 
       $\hat{\beta}_j$ is an initial estimation of $\beta_j$ 
       and $\gamma > 0$ is a
       parameter.
       A similar weighted parameter-minimization problem has also been
       considered in \citet{Candes08} in the context of signal reconstruction.
       They considered the following weighted $\ell_1$-minimization problem\footnote{
       Note that
       $\vx$ and $\vy$ 
       in the above equation
       have different meanings from other parts of this paper.
}:
       \begin{eqnarray*}
	\min_{\vx \in \mathbb{R}^n} && \sum_{i = 1}^p w_i |x_i| \\
	\mbox{s.t.} && \vy = \mb{\Phi} \vx.
       \end{eqnarray*}
       where $\vy \in \mathbb{R}^m$, 
       $\mb{\Phi} \in \mathbb{R}^{m \times n}$, and $m < n$.
       The goal of 
       this problem is to reconstruct a sparse signal $\vx$ from the measurement
       vector $\vy$ and sensing matrix $\mb{\Phi}$.
       The constraint linear equations have infinitely many solutions
       and the simplest explanation of $\vy$ is desirable.
       To estimate better sparse representation, they proposed 
       an iteratively re-weighting strategy for estimating $w_i$.
\end{itemize}
In order to apply the multi-parametric approach to these problems, 
we need to determine the search direction of the path in the 
multi-dimensional hyper-parameter space.
In many situations 
search directions can be estimated from data.

\emph{Incremental-decremental SVM}
\citep{Cauwenberghs01,Martin02,Ma03,Laskov06,Karasuyama09b} 
exploits 
the piecewise linearity of the solutions.
It updates SVM solutions efficiently when instances are added or
removed from the training set.
The incremental and decremental operations can be implemented using our
instance-weight path approach.
If we want to add an instance $(\vx_i, y_i)$, 
we increase $C_i$ from $0$ to some specified value.
Conversely, 
if we want to remove an instance $(\vx_j, y_j)$, 
we decrease $C_j$ to $0$.
The paths generated by these two approaches are different in general.
The instance-weight path keeps the optimality of all the instances including
currently adding and/or removing ones.
On the other hand,
the incremental-decremental algorithm does not satisfy the optimality of 
adding and/or removing ones until the algorithm terminates.
When we need to guarantee the optimality at intermediate points on the
path, our approach is more useful.

In the parametric programming approach, numerical instabilities sometimes
cause computational difficulty.
In practical implementation,
we usually update several quantities such as 
$\valpha$, $b$, $yf(\vx_i)$, and $\mb{L}$ from the previous values 
without calculating them from scratch.
However, the rounding error may be accumulated at every breakpoints.
We can avoid such accumulation using 
the \emph{refresh strategy}: re-calculating variables from scratch, for
 example, every $100$ steps
(note that, in this case, we need 
 $O(n |{\cal M}|^2)$ computations in every $100$ steps).
Fortunately, such numerical instabilities rarely occurred in our experiments, 
and 
the accuracy of the KKT conditions of the solutions were kept high enough.

Another (but related) numerical difficulty arises when the matrix $\vM$ 
in (\ref{eq:phi})
is close to singular.
\citet{Wang08} pointed out that 
if the matrix is singular, the update is no longer unique.
To the best of our knowledge, this degeneracy problem
is not fully solved in path-following literature. 
Many heuristics are proposed to circumvent the problem,
and we used one of them in the experiments: 
adding small positive constant to the diagonal elements of kernel
matrix.
Other strategies are also discussed in 
\citet{Wu08}, \citet{Gartner09}, and \citet{Ong10}.

Scalability of our algorithm depends on the size of $\cM$ 
because a linear system with $|\cM|$ unknowns must be solved at each 
breakpoint.
Although we can update the Cholesky factor by $O(|\cM|^2)$ cost from
the previous one, 
iterative methods such as conjugate-gradients 
may be more efficient than the direct matrix update
when $|\cM|$ is fairly large. 
When $|\cM|$ is small the parametric programming approach can be applied
to relatively large data sets 
such as more than tens of thousands of instances 
(see, for example, the ranking experiments in Section~\ref{subsec:Ranking}).

%%% Local Variables: 
%%% mode: latex
%%% TeX-master: "paper"
%%% End: 

% Acknowledgements should go at the end, before appendices and references

% \acks{We would like to acknowledge support for this project
% from the National Science Foundation (NSF grant IIS-9988642)
% and the Multidisciplinary Research Program of the Department
% of Defense (MURI N00014-00-1-0637). }


\vskip 0.2in
\bibliography{ref}

\end{document}